\documentclass[tocnosub,noragright,centerchapter,12pt]{uiucecethesis09}

\usepackage{url}
\usepackage{amsfonts}
\usepackage{amsmath}

\usepackage{graphicx}
\usepackage{amsmath}
\usepackage{amssymb}
\usepackage{subcaption}
\include{url}

\makeatletter
\usepackage{fixmath}
\usepackage[]{appendix}
\usepackage{setspace}
\usepackage{epsfig}  
\usepackage{graphicx}  
\usepackage{amsmath}  
\usepackage{amssymb}  
\usepackage{lscape}  
\usepackage[justification=raggedright]{caption}	

\usepackage{latexsym,amsfonts,epsf,cite,bbm,float}

\msthesis

\title{A Game Theoretical Framework for the Evaluation of Unmanned Aircraft Systems Airspace Integration Concepts}
\author{Negin Musavi}
\department{Mechanical Engineering}
\degreeyear{2017}

\advisor{Professor Yildiray Yildiz}

\begin{document}

%

%
\maketitle

\parindent 1em%

\frontmatter

%



\pagenumbering{roman}






\begin{abstract}
Predicting the outcomes of integrating Unmanned Aerial Systems (UAS) into the National Aerospace (NAS) is a complex problem which is required to be addressed by simulation studies before allowing the routine access of UAS into the NAS. This thesis focuses on providing 2D and 3D simulation frameworks using a game theoretical methodology to evaluate integration concepts in scenarios where manned and unmanned air vehicles co-exist. The fundamental gap in the literature is that the models of interaction between manned and unmanned vehicles are insufficient: a) they assume that pilot behavior is known a priori and b) they disregard decision making processes. The contribution of this work is to propose a modeling framework, in which, human pilot reactions are modeled using reinforcement learning and a game theoretical concept called level-k reasoning to fill this gap. The level-k reasoning concept is based on the assumption that humans have various levels of decision making. Reinforcement learning is a mathematical learning method that is rooted in human learning. In this work, a classical and an approximate reinforcement learning (Neural Fitted Q Iteration) methods are used to model time-extended decisions of pilots with 2D and 3D maneuvers. An analysis of UAS integration is conducted using example scenarios in the presence of manned aircraft and fully autonomous UAS equipped with sense and avoid algorithms.
\end{abstract}

\begin{acknowledgments}
This effort was sponsored by the Scientific and Technological Research Council of Turkey under grant number 3501-114E282.

I would like to express my sincere appreciation to my adviser Assistant Professor Dr. Yildiray Yildiz for his patience, support, motivation, professional academic behavior and for most for his supervision. I would like to state that I hope for future academic collaborations with him.

I would like to thank to my thesis committee, Assistant Professor Dr. Melih \c{C}akmak\c{c}{\i} and Assistant Professor Dr. Ali T\"{u}rker Kutay for allocating their time to investigate my work.

I am glad to express my pleasure for having Noushin Salek Faramarzi (Noosh) as a friend. She has always been a sincere and positive friend. I also thank to my vibrant and happy Nasima Afsharimani, dear Ehsan Yousefi and dear Arsalan Nikdoost for their kindness toward me during my graduate studies at Bilkent University. There are so nice friends at Bilkent University that made me feel like I am at home, thank to Ba\c{s}ak Avc{\i}, Elif Alt{\i}ntepe, Tuna Demirba\c{s} and Nurten Bulduk.

Last but not least, I wish to thank to my dear father for his endless support in all of the stages of my life. I thank to my mother for her patience and I thank to my dear friend Mahsa Asgarisabet for her passionate friendship.

I dedicate this thesis to my dear brother Nima. He will survive in my heart and mind forever. May he rest in peace.
\end{acknowledgments}
%


%

%
\tableofcontents

%

%
\listoffigures

%


%

\mainmatter

\chapter{Introduction} 

\section{Objective and Motivation}

Unmanned Aerial Systems (UAS) refer to aircraft without an on-board human pilot. Instead, UAS can be controlled with an off-board operator or can be programmed to fly autonomously. UAS have operational and cost advantages over manned aircraft in many areas such as surveillance, commercial, scientific and agricultural, and the interest in UAS is increasing rapidly. However, UAS still do not have routine access to National Airspace System (NAS) \cite{Dalamagkidi:08}. Since technologies, standards and procedures for a safe integration of UAS into airspace haven't matured yet, UAS can fly only in segregated airspace with restricting rules in most of the countries. For instance, Federal Aviation Administration (FAA) in USA have regulations which prevents routine UAS operations in the NAS. Aviation industry is very sensitive to risk and for new vehicles such as UAS to enter into this sector, they need to be proven to be safe and it must be shown that they will not affect the existing airspace system in any negative way \cite{US_UAV:13, European_UAV:13}. Technical issues such as communication between the UAS and air traffic control as well as a trustworthy Sense and Avoid (SAA) system are among the primary impediments that prevent the integration of the UAS into the NAS \cite{Dalamagkidi:08}. Sense and avoid capability refers to the ability of UAS to sense, detect and execute maneuvers to resolve a conflict with other aircraft and obstacles in the surrounding traffic. Until technologies, standards, requirements and procedures for a safe integration of UAS into airspace are matured, there will not be enough data accumulated about the technical issues and it will be hard to predict the effects of the technologies and concepts that are developed for the integration. Although research efforts exist to develop a safe and efficient real test environment for UAS integration \cite{NASA:12}, experimental tests are expensive and experimental failures can cause severe economic loss. Therefore, employing simulations is currently the only way to understand the effects of UAS integration on the air traffic system \cite{MITRE:14}. These simulation studies need to be conducted with hybrid airspace system (HAS) models, where manned and unmanned vehicles coexist.

Before an SAA system is approved, its potential impact on the safety of the surrounding air traffic should be analyzed. To perform an analysis and evaluation of any SAA logic, it is necessary to model the actions that a pilot would take during a conflict \cite{Maki:12}. HAS models in the literature are generally based on the assumption that the pilots of manned aircraft always behave as they should, without deviating from the ideal behavior. In their work, \cite{Maki:12}, Maki \textit{et al.} constructed an SAA logic based on the model developed by the MIT Lincoln Laboratory \cite{Kochenderfer:08} and Kuchar \textit{et al.} \cite{Kuchar:04} did a rigorous analysis of the traffic control alert system (TCAS) to implement an SAA algorithm for remotely piloted vehicles. TCAS is an on-board collision avoidance system which observes and tracks surrounding air traffic, detects conflicts and suggests avoidance maneuvers to the pilots. In both of these studies, SAA system are evaluated through simulations in a platform which uses MIT's NAS encounter model. In the evaluations, it is assumed that pilot decisions are known a priori and depend on the relative motion of the adversary during a specific conflict scenario. In their work, Perez-Batlle \textit{et al.} \cite{Perez:12} classified separation conflicts between manned and unmanned aircraft and proposed separation maneuvers for each class. These maneuvers are tested in a simulation environment, where it is assumed that the pilots will follow these maneuvers 100\% of the time. Florent \textit{et al.} \cite{Florent:10} developed a SAA algorithm and tested it via simulations and experiments. In both of these tests, it is assumed that the intruding aircraft does not change its path while the UAS is implementing the SAA algorithm. There are other simulation studies such as \cite{Billingsley:06} that test and evaluate different collision avoidance algorithms, where some predefined actions are used as pilot models. There are also substantial studies with remotely piloted aircraft where the effects of SAA systems on the workload and situational awareness of the pilots are investigated via simulations and flight tests \cite{Alfredson:15}, \cite{Alfredson:14}. These HAS models are designed to evaluate and test the performance of collision avoidance systems in single encounter scenarios in which the intruder (generally a manned aircraft) has a pre-defined behavior with no consideration of the decision making process of the pilot. These models are valuable and essential at the initial stages of evaluating a new method but it is not realistic to expect that the pilot, as a decision maker (DM), will always behave deterministically and in a pre-defined manner. It is not always predictable, for example, how pilots will respond to the TCAS \cite{Salas:10}. The collision between two aircraft (a DHL Boeing 757 and a Bashkirian Tupolev 154) over Uberlingen, Germany, near the Swiss border On 21:35 (UTC) July 1, 2002, is a good evidence that pilots may decide not to act parallel to TCAS advisory or may ignore traffic controller's commands, during high-stress situations \cite{Salas:10}. In addition, in recent studies, it was shown that only 13\% of pilot responses match the deterministic pilot model that was assumed for TCAS development \cite{Lee:12}, \cite{Kuchar:07}. Therefore, incorporating human decision-making processes in HAS models has a strong potential to improve the predictive power of these models.

\section{Approach}
In this thesis, a 2D and a 3D game theoretical Hybrid Airspace System (HAS) modeling framework are built, where pilot reactions are obtained through a  decision making process. Below, 2D and 3D frameworks are explained separately.

\textbf{2D framework:} In the 2D HAS model, the pilot behavior is not assumed to be known a priory and decisions are obtained utilizing a) the bounded rationality concept, which helps model imperfect decisions, as opposed to modeling the pilot as a perfect decision maker, b) reinforcement learning, which helps model time-extended decisions, as opposed to assuming one-shot decision making. In order to predict pilot reactions in complex scenarios where UAS and manned aircraft co-exist, in the presence of automation such as an SAA system, a game theoretical methodology is employed, which is formally known as semi network-form games (SNFGs) \cite{Lee:12}. Using this method, probable outcomes of HAS scenarios are obtained that contains interacting humans (pilots) who also interact with UAS equipped with an SAA algorithm. In these scenarios, close encounters are simulated where TCAS and air traffic management instructions can also be incorporated. To obtain realistic pilot reactions, bounded rationality is imposed by utilizing the level-k approach \cite{Costa:09, Stahl:95}, a concept in game theory which models human behavior assuming that humans think in different levels of reasoning. In the proposed framework, pilots optimize their trajectories based on a goal function representing their preferences for system states. During the simulations, UAS fly autonomously based on a pre-programmed flight plan. In these simulations, the effect of certain system variables, such as \textit{horizontal separation requirement} and \textit{required time to conflict} for UAS and the effect of \textit{responsibility assignment for conflict resolutions} on the safety and performance of the HAS are analyzed (see \cite{NASA:12} for the importance of these variables and responsibility assignment for UAS integration.). To enable UAS to perform autonomously in the simulations, it is assumed that they employ an SAA algorithm. The simulation results are provided for 2 different SAA methods and, in addition, these 2 methods are compared quantitatively in terms of safety and performance, using the proposed modeling framework.

In prior works, the method exploited in this paper was used to investigate small scale scenarios (in terms of number of agents): In \cite{Backhaus:13}, the dynamics between a smart grid operator and a cyber attacker, and in \cite{Lee:13, Yildiz:12, Lee:12}, the dynamics between two interacting pilots are modeled. More recently, in \cite{Yildiz:14}, a medium scale scenario with 50 interacting pilots is analyzed. In the study with 50 pilots, the simulation environment utilized a gridded airspace where aircraft moved from a grid intersection to another to represent movement. In addition, the pilots could only observe grid intersections to see whether or not another aircraft was nearby. All these simplifying assumptions decreased the computational cost but also decreased the fidelity of the simulation. In this thesis, a) a dramatically more complex scenario in the presence of manned and unmanned aircraft is investigated, b) the simulation environment is not discretized and the aircraft movements are simulated in continuous time, c) realistic aircraft and UAS physical models are used and d) initial states of the aircraft are obtained from real flight data. Hence, a much more representative simulation environment with the inclusion of UAS equipped with a SAA algorithm is used to obtain probabilistic outcomes of HAS scenarios.

\textbf{3D framework:} The 2D game theoretical approach has two limitations: First, HAS models are developed for a 2D airspace. Second, the policies, i.e. maps from observation spaces to action spaces, obtained for the decision makers remain unchanged during their interaction. To remove these limitations the 2D framework is extended and a 3D HAS model is introduced where the strategic decision makers can modify their policies during interactions between each other. Therefore, compared to the 2D HAS model a much larger class of interactions are modeled. 

It is shown in the literature that 1) in repeated strategic interactions, where agents consider other agents' possible actions before determining their own, agents with different cognitive abilities change their behavior during the interaction \cite{Gill:16}; and 2) there is a positive relationship between cognitive ability and reasoning levels \cite{Gill:12} and \cite{Gill:16}. These observations lead to agents with different levels of reasoning who can observe their opponents behavior during repeated interactions, update their beliefs on their opponents reasoning level and change their own level-k rule against them. In their works in \cite{Gill:12} and \cite{Gill:16}, authors introduce a systematic level-k structure where players can update their beliefs about their opponents, and switch their own level rule up one level during their interactions. There are also other level-k rule learning models in the literature such as the one presented in \cite{Chong:16} and \cite{Ho:13}. Based on the rule learning methods introduced in these works the agent levels can reach up to infinity. This is not a problem for the applications investigated in these studies since they are executed on games in the field of economics in which obtaining level-k rules (k=0,1,2,...,infinity) are straight forward due to the number of agents in the games. Since 1) it is computationally expensive to obtain higher levels, and 2) in certain experimental studies it is shown that humans in general have a maximum reasoning level of 2 \cite{Costa:09}, the existing level-k rule learning methods may not be suitable for the application considered in this work where more than 180 decision makers are modeled simultaneously in a time extended manner. In this study, we propose a simpler method for modeling level rule update during interactions by a) limiting the levels up to 2 and b) allowing rule update only if a trajectory conflict is detected. 

Different from the 2D HAS model developed in \cite{Negin:16}, \cite{Musavi:16}, in this study, the game theoretical modeling framework is developed for a 3D HAS model which allows to cover a much larger class of integration scenarios. The reinforcement learning algorithm used in the authors' earlier works \cite{Negin:16}, \cite{Musavi:16} employs a table to store the Q values of all state (location of the intruder in a girded observation space, approach angle of the intruder, best trajectory action, best destination action and previous action)-action (turn left, turn right, go straight) pairs, which define how preferable to take a certain action given the observations/states. This poses a challenge for the application of the method to systems with a large numbers of state-action pairs such as the proposed 3D HAS model in this study. To circumvent this issue, Neural Fitted Q Iteration (NFQ) method \cite{Rie:05}, \cite{Rie:07} and \cite{Rie:11}, an approximate reinforcement learning algorithm is utilized. Approximate reinforcement learning methods use function approximators to represent the Q value function \cite{Volo:15}. In other words, instead of saving Q values for each state-action pair, Q value function is approximated by a function approximator. In the case of NFQ, a neural network is used as the function approximator. NFQ approach also allowed using a continuous observation space, which also contributed to obtain a more precise definition of the agents observations, compared to [our earlier papers], where a discretized observation space was used. 

In the simulations, pilot models, that are obtained using the proposed game theoretical modeling framework, are used in complex scenarios, where UAS and manned aircraft co-exist, to analyze the probable outcomes of HAS scenarios. The HAS scenarios contain interacting humans (pilots) who also interact with multiple UAS with their own sense and avoid (SAA) systems. It is noted that automation other than SAA systems, such as TCAS, and possible air traffic management instructions can also be incorporated into the proposed framework. During the simulations, UAS fly autonomously based on a pre-programmed flight plan. In these simulations, the effect of responsibility assignment for conflict resolutions on the safety and performance of the HAS are analyzed.

\section{Organization}
The organization of the thesis in the following chapters is is described as below:
Chapter II is devoted to the 2D game theoretical framework. The modeling methodology as well as details of the scenario consisting of multiple manned aircraft and a UAS is described completely. The obtained 2D HAS model is validated and the results of simulation for the UAS integration is shown at the end of chapter.
In Chapter III, the 3D game theoretical pilot model and the modifications that are done upon the 2D HAS model are explained in detail. The rest of chapter is devoted to the results of simulation in single encounter scenarios and UAS integrated scenario.
Finally, concluding remarks are provided in Chapter IV.

\chapter{2D Framework}

\section{Modeling Methodology} \label{Pilot Behavior Modeling Methodology}
The most challenging problem in the prediction of the outcomes of complex scenarios where manned and unmanned aircraft co-exist is obtaining realistic pilot models. A pilot model in this paper refers to a mapping from observations of the pilot to his/her actions. To achieve a realistic human reaction model, certain requirements need to be met. First, the model should not be deterministic since it is known from everyday experience that humans do not always react exactly the same when they are in a given ``state''. Here, ``state'' refers to the observations and the memory of the pilot. For instance, observing that an aircraft is approaching from a certain distance is an observation and remembering one's own previous action is memory. Second, pilots should show the characteristics of a strategic Decision Maker (DM) meaning that the decisions must be influenced by the expected moves of other ``agents''. Agents can be either the other pilots or the automation logic of UAS. Third, the decisions emanating from the model should not always be the best (or mathematically optimal) decisions since it is known that human actions are less than optimal in many situations. Finally, it should be considered that a human DM's predictions about other human DMs are not \textit{always} correct. To accomplish all of these requirements, level-k reasoning and reinforcement learning are used together, forming a non-equilibrium game theoretical solution concept. It is noted that in this study, the UAS are assumed to be fully autonomous.

\subsection{Game Theoretical Modeling of Interactive Decision Making} \label{Level-k Reasoning}
Level-k reasoning is a game theoretical solution concept whose main idea is that humans have various levels of reasoning in their decision making process. Level-0 represents a ``non-strategic'' DM who does not take into account other DMs' possible moves when choosing his/her own actions. This behavior can also be named as reflexive since it only reacts to the immediate observations. In this study, given a state, a level-0 pilot flies an aircraft with constant speed and heading starting from its initial position toward its destination. A level-1 DM assumes that the other agents in the scenario are level-0 and takes actions accordingly to maximize his/her rewards. A level-2 DM takes actions as though the other DMs are level-1. In a hierarchical manner, a level-k DM takes actions assuming that the other DMs behave as level-(k-1) DMs. 

\subsection{Reinforcement Learning for the Partially Observable Markov Decision Process} \label{Reinforcement Learning}
Reinforcement learning is a mathematical learning mechanism which mimics the human learning process. An agent receives an observable message of an environment's state, then chooses an action, which changes the environment's state, and the environment in return encourages or punishes the agent with a scalar reinforcement signal known as reward. Given a state, when an action increases (decreases) the value of an objective function (reward), which defines the goals of the agent, the probability of taking that action increases (decreases). Reinforcement learning algorithms mostly involve estimating two value functions: State value function $V$ and state-action value function $Q$ \cite{Sutton:98}. $V(s)$, the value of the state ``s'', is the total amount of reward an agent can expect to gather in the future, starting from that state. It is an estimate of how good it is for an agent to be in a particular state. $Q(s,a)$, the value of taking the action ``a'' in the state ``s'', under a particular policy, is the total amount of reward an agent can expect to gather in the future starting from that state and taking that action by following the given policy. It is an estimate of how good it is for an agent to perform a given action in a given state.

Since the human DM as the agent in the reinforcement learning process is not able to observe the whole environment state (i.e. the positions of all of the aircraft in the scenario), the agent receives only partial state information. By assuming that the environment is Markov, and based on the partial observability of the state information, this problem can be generalized to a Partially Observable Markov Decision Process (POMDP). In this study, the reinforcement learning algorithm developed by Jaakkola \textit{et al.} \cite{Jaakkola:95} is used to solve this POMDP. In this approach, different from conventional reinforcement learning algorithms, the agent does not need to observe all of the states, for the algorithm to converge. A POMDP value function and a Q-value function, in which $m$ and $a$ refer to the observable message of the state, $s$, and action, respectively, are given by:
\begin{equation}
V(m) = \sum\nolimits_{s \in m} P(s|m)V(s)
\end{equation}
\begin{equation}
Q(m,a) = \sum\nolimits_{s \in m} P(s|m)Q(s,a).
\end{equation}
A recursive Monte-Carlo strategy is used to compute the value function and the Q-value function. It is noted that the pilot model is given in terms of a policy, $\pi$, which is a mapping from observations or messages, $m$, to actions, $a$. During reinforcement learning, the policy update is calculated as below, where the policy is updated toward $\pi^{1}$ with $\varepsilon$ learning rate, after each iteration:
\begin{equation}
\pi(a|m)\rightarrow(1-\varepsilon)\pi(a|m)+\varepsilon\pi^{1}(a|m)
\end{equation}
where, $\pi^{1}$ is chosen such that, $J^{\pi^{1}}=\max\limits_{a}(Q^{\pi}(m,a)-V^{\pi}(m))$. For any policy $\pi^{1}(a|m)$, $J^{\pi^{1}}$ is defined as:
\begin{equation}
J^{\pi^{1}}= \sum\nolimits_{a} \pi^{1}(a|m)(Q^{\pi}(m,a)-V^{\pi}(m)).
\end{equation}
It is noted that the pilot model, or the policy, is obtained once the policy converges during this iterative process.

\subsection{Combining Game Theory With Reinforcement Learning} \label{sec:Combination}
The method employed in this study carefully combines the two concepts explained in the above sections: game theory (GT) and reinforcement learning (RL). The method consists of two stages: 1) obtaining pilot reaction models with various levels (level-k) and 2) simulating a given scenario using these models. In the first stage, which can also be considered as the ``training'' stage, a level-1 type model is trained by assigning level-0 behavior to all of the agents but the one that is being trained. The trainee learns to react as best as he/she can in this environment using RL. Thus, the resulting behavior becomes a level-1 type. Similarly, a level-2 behavior is trained by assigning level-1 behavior to all of the agents but the trainee. This process continues until the highest desired level is reached. Once all of the desired levels are obtained, the first stage ends and in the second stage a given scenario is simulated by assigning certain proportions of these levels to the agents in the scenario. It is noted that in the scenario investigated in this paper, the pilots of the manned aircraft have level-0, level-1 and level-2 behavior types whereas the movements of the UAS are commanded via sense and avoid algorithms. 

\section{Components of the Hybrid Airspace Scenario} \label{Hybrid Airspace Scenario}
\begin{figure}[htp]
	\centering	
	\includegraphics[width=13.5cm]{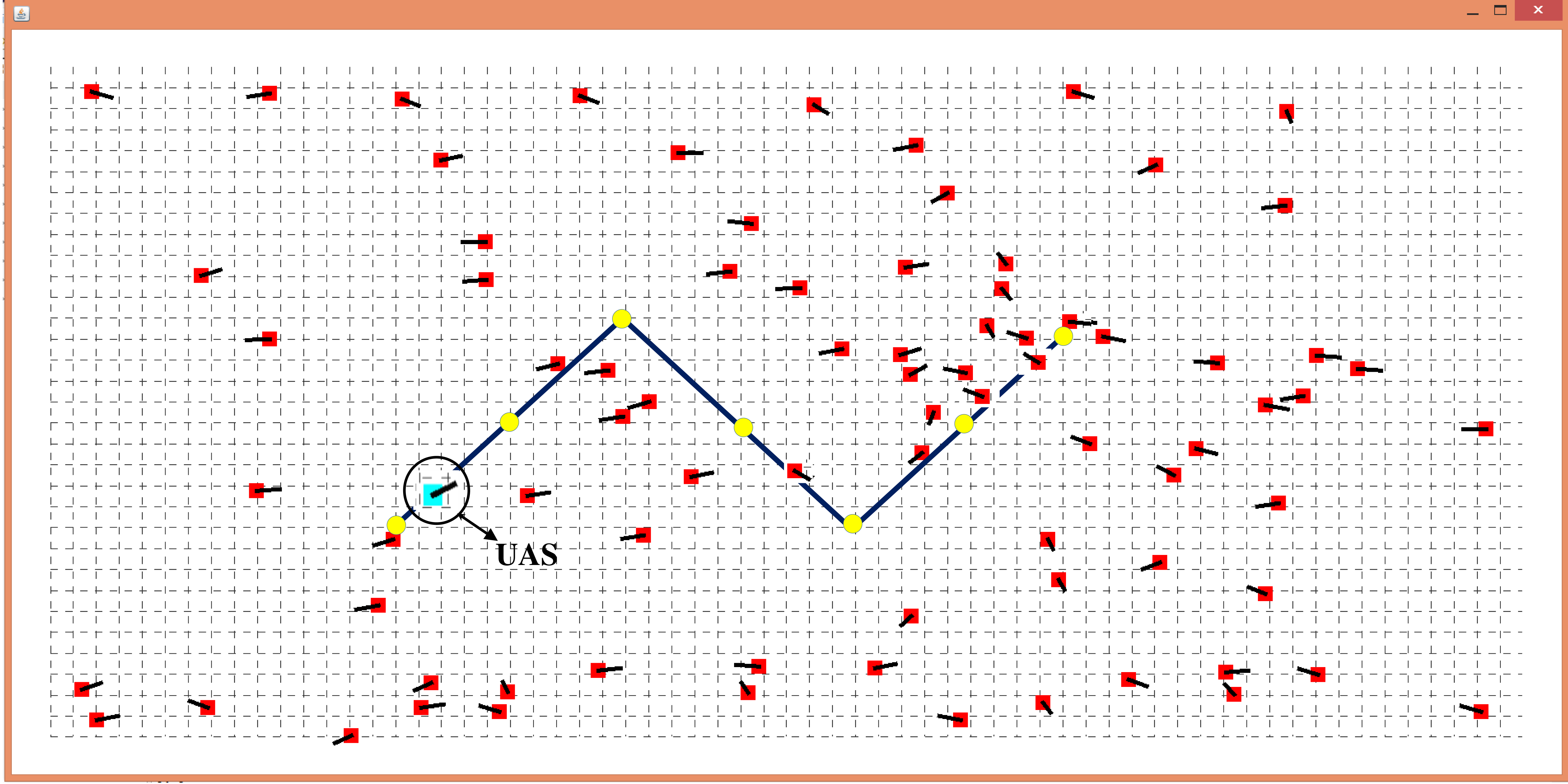}
	\caption{Snapshot of the hybrid airspace scenario in the simulation platform. Each square stands for a $5nm\times5nm$ area.}
	\label{f:Crowded_Scenario}
\end{figure}

The investigated scenario consists of 180 manned aircraft with predefined desired trajectories and an UAS which moves based on its pre-programmed flight plan from one waypoint to another. Fig.~\ref{f:Crowded_Scenario} shows a snapshot of this scenario where the red squares correspond to manned aircraft and the cyan square corresponds to the UAS. The size of the considered airspace is $600km\times300km$. The airspace is gridded just to make it easier to visually grasp the dimensions (two neighboring grid points are $5nm$ away), nevertheless all aircraft, manned or unmanned, move in the airspace continuously. Yellow circles show the predetermined waypoints that the UAS is required to pass. The blue lines passing through the waypoints show the predetermined path of the UAS. It is noted that the UAS does not follow this path exactly since it needs to deviate from its original trajectory to avoid possible conflicts using an on-board SAA algorithm.
The initial positions, speeds and headings of the aircraft are obtained from Flightradar24 website which provides live air traffic data (\url{http://www.flightradar24.com}). The data is collected from air traffic volume on Colorado state, USA airspace on 11 March, 2015. It is noted that in the Next Generation Airspace System (NextGen) air travel demand is expected to increase dramatically, thus traffic density is expected to be much more than it is today. To represent this situation, the number of aircraft in the scenario is increased by projecting various aircraft at different altitudes to a given altitude. To handle the increase in aircraft volume in NextGen, it is expected that new technologies and automation will be introduced such as the automatic dependent surveillance-broadcast (ADS-B), which is a technology that enables an aircraft to receive other aircraft's identification, position and velocity information and to send its information to others. In the investigated scenario, it is assumed that each aircraft is equipped with the ADS-B. It is noted that the ADS-B can also provide information about the flight path of an aircraft, which is highly relevant to collision avoidance. In our simulations, we provided this crucial information by answering the question ``in a given time window, where in my observation space do I expect an intruding aircraft?'', for each agent (see Section~\ref{Pilot Observations and Memory} for details).

\subsection{Pilot Observations and Memory} \label{Pilot Observations and Memory}
\begin{figure}[!htb]
	\centering	
	\includegraphics[width=13.5cm]{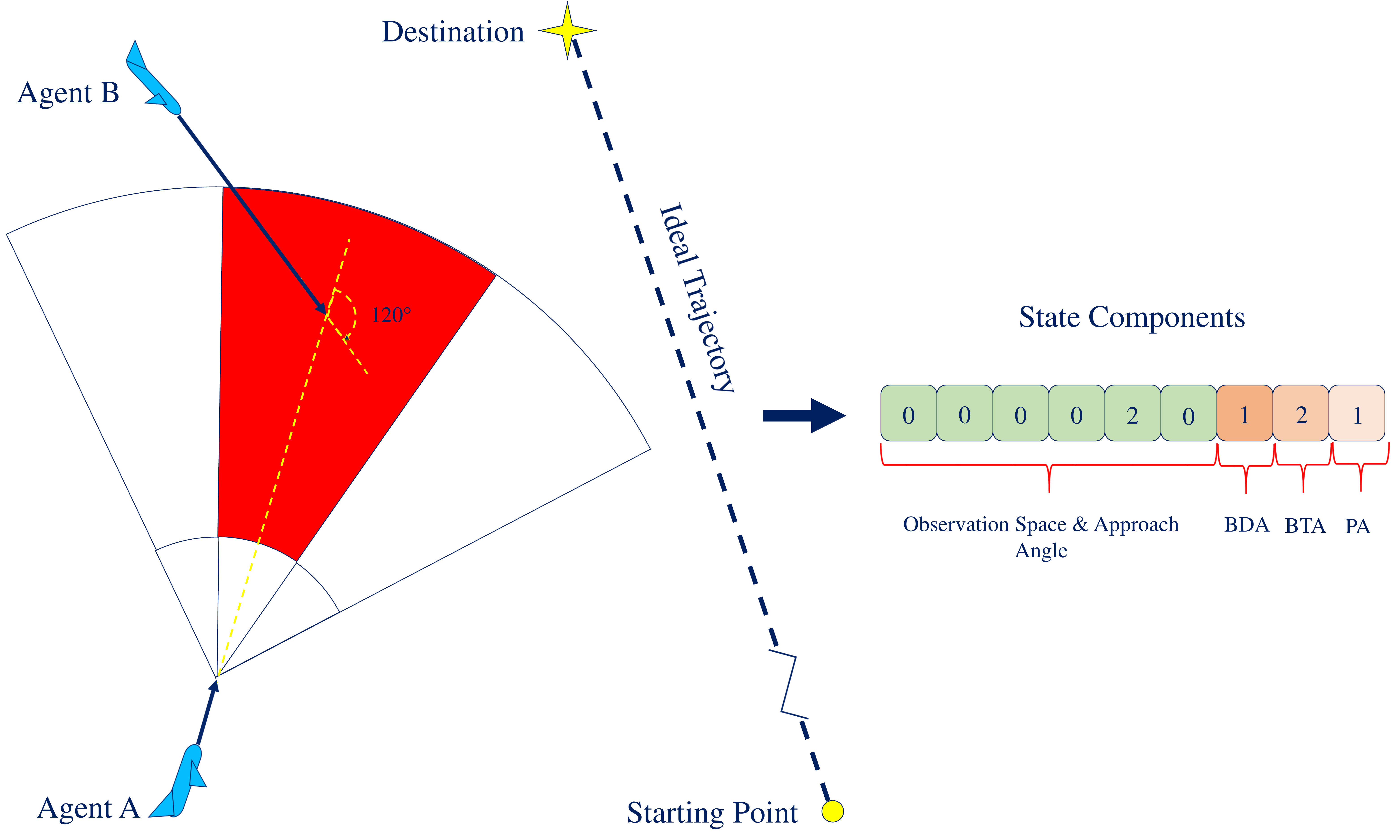}
	\caption{Pilot observation space.}
	\label{f:figure_POM}
\end{figure}

Although ADS-B provides the positions and the velocities of other aircraft, with his/her limited cognitive capabilities a pilot can not possibly process all this information during his/her decision making process. In this study, in order to model pilot limitations, including the limitations at visual acuity and perception depth, as well as the limited viewing range of an aircraft, it is assumed that the pilots can observe (or process) the information from a limited portion of the nearby airspace. This limited portion is simulated as equal angular portions of two co-centered circles called the ``observation space'' which is schematically depicted in Fig.~\ref{f:figure_POM}. The radius of the inner circle represents pilot vision range, which is taken as $1nm$ based on a survey executed in \cite{Wolfe:02}. The radius of the outer circle is a variable that depends on the separation requirements. Since standard separation for manned aviation is $3-5nm$ \cite{Perez:12}, this radius is taken as $5nm$. Whenever an intruder aircraft moves toward one of the 6 regions of the observation space (see Fig.~\ref{f:figure_POM}), the pilot perceives that region as ``full''. The pilot, in addition, can roughly distinguish the approach angle of the approaching intruder. A ``full'' region is categorized into four cases; with a) $0^{\circ}<$ approach angle $<90^{\circ}$, b) $90^{\circ}<$ approach angle $<180^{\circ}$, c) $180^{\circ}<$ approach angle $<270^{\circ}$ and d) $270^{\circ}<$ approach angle $<360^{\circ}$. Fig.~\ref{f:figure_POM} depicts a typical example, where pilot A observes that aircraft B is moving toward one of the 6 regions that is colored. In this particular example, pilot A perceives the colored region as ``full'' with approach angle in the interval $[90^{\circ} , 180^{\circ}]$ and the rest of the regions as ``empty''. The information about emptiness, fullness of a region and approach angle is fed to the reinforcement learning algorithm simply by assigning 0 to empty regions and 1, 2, 3, 4 to full regions, based on the approach angle classifications explained above. Pilots also know the best action that would move the aircraft closest to its trajectory (BTA: Best Trajectory Action) and the best action that would move the aircraft closest to its final destination (BDA: Best Destination Action). Moreover, pilots have a memory of what their actions were at the previous time step. Given an observation, the pilots can choose between three actions: $45^{\circ}$ left, straight, or $45^{\circ}$ right, which are coded with numbers 0, 1 and 2.
Six ADS-B observations, one BTA, one BDA, and one previous move make up nine total inputs for the reinforcement learning algorithm. Observations get 5 values, 0, 1, 2, 3 or 4. The previous move, BTA and BDA have three dimensions each: $45^{\circ}$ left, $45^{\circ}$ right, or straight. Therefore, the number of states for which the reinforcement learning algorithm needs to assign appropriate actions is $5^{6}\times3^{3}=421,875$.
\setlength{\parskip}{0in} 
\subsection{Pilot Objective Function} \label{Pilot Objective Function}
The goal of the reinforcement learning algorithm is to find the optimum probability distribution among possible action choices for each state. As explained above, reinforcement learning achieves this goal by evaluating actions based on their return which is calculated via a reward/objective function. A reward function can be considered as a happiness function, goal function or utility function which represents, mathematically, the preferences of the pilot among different states. In this paper, the pilot reward function is defined as
\begin{equation} \label{eq:re}
reward = w1*(-C)+w2*(-S)+w3*(-CA)+w4*(D)+w5*(-P)+w6*(-E).
\end{equation}
In (2.5), $``C"$ is the number of aircraft within the collision region. Based on the definition provided by the Federal Aviation Administration (FAA), the radius of collision is taken as $500ft$ \cite{NextGen:07}. $``S"$ is the number of air vehicles within the separation region. The radius of the separation region is $5nm$ \cite{Perez:12}. $``CA"$ represents whether the aircraft is getting closer to the intruder or going away from the intruder and takes values 1, for getting closer, or 0, for going away. $``D"$ represents how much the aircraft gets closer to or goes away from its destination normalized by the maximum distance it can fly in a time step. The time step is determined based on the frequency of pilot decisions. The average time step during reinforcement learning is determined to be 20 seconds. $``P"$ represents how much the aircraft gets closer to or goes away from its ideal trajectory normalized by the maximum distance it can fly in a time step and $``E"$ represents whether or not the pilot makes an effort (move). $``E"$ gets a value of 1 if the pilot makes a new move and 0 otherwise.

\subsection{Manned Aircraft Model} \label{Manned Aircraft Model}
The initial positions, speeds and heading angles of the manned aircraft are obtained from Flightradar24 website (http://www.flightradar24.com). It is assumed that all aircraft are in their en-route phase of travel with constant speed, $\|\vec{v}\|$, in the range of $[150-550]knots$. Aircraft are controlled by their pilots who may decide to change the heading angle for $45^{\circ}$, $-45^{\circ}$ or may decide to keep it unchanged. Once the pilot gives a heading command, the aircraft moves to the desired heading, $\psi_{d}$, in the constant speed mode. The heading change is modeled by a first order dynamics with the standard rate turn: a turn in which an aircraft changes its heading at a rate of $3^{\circ}$ per second ($360^{\circ}$ in 2 minutes) \cite{pilot-handbook:08}. This is modeled as a first order dynamics with a time constant of $10s$ ($45 \times (1-1/e)/3 \approx 10$). Therefore, the aircraft heading dynamics can be given as
\begin{equation}
\dot{\psi}= -\frac{1}{10}\times(\psi-\psi_{d})
\end{equation}
and the velocity, $\vec{v}=(v_{x},v_{y})$, is then obtained as:
\begin{equation}
\\v_{x}= \|\vec{v}\|\sin\psi
\end{equation}
\begin{equation}
\\v_{y}= \|\vec{v}\|\cos\psi.
\end{equation}

\subsection{UAS Model} \label{UAS Model}
The UAS is assumed to have the dynamics of a RQ-4 Global Hawk with operation speed of $340knots$ \cite{Dalamagkidis:09}. It moves according to its pre-programmed flight plan and is also equipped with a SAA system. The SAA system can initiate a maneuver to keep the UAS away from other traffic, if necessary, by commanding a velocity vector change. Otherwise, the UAS will continue moving based on its mission plan. Therefore, the UAS always receives a velocity command either to satisfy its mission plan or to protect its safety. Since the UAS has a finite settling time for velocity vector changes, the desired velocity, $\vec{V_{d}}$ cannot be reached instantaneously. Therefore, the velocity vector variation dynamics of the UAS is modeled by a first order dynamics with a time constant of $1s$ \cite{Mujumdar:11} which is represented as:
\begin{equation}
\dot{\vec{v}}= -(\vec{v}-\vec{v}_{d}).
\end{equation}
\subsection{Sense And Avoid Algorithms} \label{Sense and Avoid Algorithm}
In order to assure that the UAS can detect probable conflicts and can autonomously perform evasive maneuvers, it should be equipped with a SAA system. In this paper, two different SAA algorithms are investigated. These SAA algorithms are developed by Fasano \textit{et al.} \cite{Fasano:08} referred as SAA1 and Mujumdar \textit{et al.} \cite{Mujumdar:11} referred as SAA2. The algorithms consist of two phases; \textit{conflict detection} phase and \textit{conflict resolution} phase. In the detection phase, the SAA algorithms project the trajectories of the UAS and the intruder aircraft in time, using a predefined time interval, and if the minimum distance between the aircraft during this time is calculated to be less than a minimum required distance, $R$, it is determined that there will be a conflict. The same conflict detection logic is used for both of the two SAA algorithms. In order to prevent the conflict, the UAS starts an evasive maneuver in the conflict resolution phase, which is handled differently for SAA1 and SAA2. In SAA1 resolution phase algorithm, a velocity adjustment is suggested that guarantees minimum deviation from the trajectory. The velocity adjustment command, $\vec{v_{A}^{d}}$, for the UAS is given in the equation below
\begin{equation}
\label{e:function}
\vec{v}_{A}^{d} =
\left[ \frac{v_{AB}\cos(\eta-\zeta)}{\sin(\zeta)}[\sin(\eta)\frac{\vec{v}_{AB}}{v_{AB}}-\sin(\eta-\zeta)\frac{\vec{r}}{\|\vec{r}\|}] \right]+\vec{v}_{B}
\end{equation}
where, $\vec{v_{A}}$ and $\vec{v_{B}}$ refer to the velocity vectors of the UAS and the intruder. $\vec{r}$ and $\vec{v_{AB}}$ denote the relative position and velocity between the UAS and the intruder, respectively. $\zeta$ is the angle between $\vec{r}$ and $\vec{v_{AB}}$ and $\eta$ is calculated as $\eta=\sin^{-1}\frac{R}{\|\vec{r}\|}$. In the case of multiple conflict detection, the UAS will start an evasive maneuver to resolve the conflict that is predicted to happen earliest. In the SAA2 algorithm, the velocity adjustment vector is determined as given in the below equation in which $\vec{r}_{m}$ stands for the minimum relative position vector between the UAS and the intruder during the conflict:
\begin{equation}
\label{e:function}
\vec{v}_{A}^{d} =
\frac{-\vec{v}_{A}(\frac{\vec{r}_{0}.\vec{v}_{AB}}{\|\vec{v}_{AB}\|})-(R-\|\vec{r}_{m}\|)\frac{\vec{r}_{m}}{\|\vec{r}_{m}\|}}{\|-\vec{v}_{A}(\frac{\vec{r}_{0}.\vec{v}_{AB}}{\|\vec{v}_{AB}\|})-(R-\|\vec{r}_{m}\|)\frac{\vec{r}_{m}}{\|\vec{r}_{m}\|}\|}
\end{equation}
where, $\vec{r}_{0}$ refers to the initial relative position vector between the UAS and the intruder. In this solution strategy, UAS moves to resolve the conflict until it retains the minimum safe distance with the intruder. Similar to the SAA1 algorithm, in the case of multiple intruders, the UAS will start an evasive maneuver to resolve the conflict that is predicted to happen earliest.

\subsection{Model Validation}
As noted earlier, since the routine access of UAS into NAS is not a reality yet and thus there is not enough experience accumulated about the issue, it is extremely hard to predict the effects of the technologies and concepts that are developed for the integration. Therefore, employing simulation is currently the only way to understand the effects of UAS integration on the air traffic system \cite{MITRE:14}. However, regardless of whether the modeled system exists or whether it is expected to be a reality in the future, the representative model should be validated \cite{Law:08}. 

Below, we break down the validation task into two steps. In the first step, we explain that the underlying hierarchical game theoretical modeling approach is a useful and valid approach to model complex human interactions, based on earlier experimental studies. In the second step, we investigate the validity of the proposed approach for the UAS integration implementations. Since UAS integration data is not available yet, we take a different approach in this step: We first provide a validation methodology that can be used to validate the proposed approach when the data for UAS integration becomes available. We also explain that the proposed model has enough degrees of freedom that can be used to obtain a predictive model using this data. Then, we proceed to show that the trajectories created by the proposed model is similar to that of a validated encounter model for manned aircraft created using real radar data. Finally, we explain a validation method, called ``face validation'', which is commonly used  when the modeled system is expected to be a reality in the future, and we argue that our simulation results in the next section can be used to apply this method. 

\subsubsection{Validation of the game theoretical modeling approach}
In this study, we utilized a well known game theoretical modeling approach called ``level-k'' reasoning. The advantage of this approach is its computational simplicity, where the intelligent agent makes behavioral assumptions about others and then produces the best response accordingly based on a reward function. Because of this simplicity, in multi-move scenarios, such as the ones treated in this research, level-k reasoning provides computationally tractable solutions. This approach not only provides a computationally efficient solution but also is shown to be able to model complex human interactions in experimental settings: In \cite{Costa:09}, several experimental results are conducted using various sizes of subject pools that are made to play different games. Using the data from these games, models of strategic thinking are evaluated and compared and although some perform better than others depending on the game type, ``level-k'' approach is found to be ``behaviorally more plausible''. It is noted that these experimental results are cited here not to prove that the proposed approach is better than other game theoretical approaches, but to provide real world data showing that the ``level-~k'' modeling approach can represent real world behavior in complex decision making scenarios. In the scenarios studied in this work, the intelligent agents (pilots) are also strategic decision makers as explained in earlier sections and they need to make decisions in a complex environment to maximize their rewards. Therefore, the underlying game theoretical approach can also be considered as a good fit to the problem studied in this work. 

\subsubsection{Validation of the proposed modeling approach for UAS integration concepts}
As presented in the previous section, the game theoretical modeling approach is shown to model real world behavior in earlier experimental studies. In this section, we address the question ``can this approach be reliably used to model UAS integration scenarios?''.

\textbf{Validation methodology:} There exist several model validation methods such as ``face validity'' , ``historical data validation'', ``parameter sensitivity analysis'' and ``predictive validation'' \cite{MITRE_VV:16}. Among these methods, the most definitive technique is predictive validation, where the model is used to predict the system's behavior and then the outputs of the model and the real system are compared. Here, we explain two main aspects of predictive validation \cite{Kochenderfer:08} that can be used to validate the proposed model in this study, when the UAS integration data becomes available: First, relevant statistics between the model and the real data should have reasonable agreement. For example, for UAS integration, the average deviation of the UAS from their intended trajectory against the type of the SAA algorithm should be similar between the model and the data. Similarly, the average number of separation violations between manned and unmanned aircraft for different kinds of SAA algorithms should match. Secondly, individual encounters should show similar characteristics. For example, the minimum separation distance between UAS and manned aircraft, and pilot decisions during encounters with similar geometry (approach angle, heading etc.) should be able to be predicted with reasonable accuracy by the model.

textbf{Comparison with a validated model:} Since UAS integration data is not available yet, we compared the results of the proposed model with a manned-aircraft-only encounter model created and validated by the Lincoln Laboratory using real radar data \cite{Kochenderfer:08}. Sample trajectories are given in two text files that are open to the public: \texttt{cor\_ac1.txt} and \texttt{cor\_ac2.txt}. Among the encounters provided, 5 of them (listed as 3rd, 16th, 23rd, 34th and 45th encounters) do not employ altitude or speed variations for conflict resolution and thus can be used for our purposes. The objective of these comparisons are to show that the actions taken by the pilots in the proposed game theoretical model and the validated model are similar. In addition, minimum separation distances experienced between the aircraft and the times that the minimum separation occurs are shown to be reasonably close to each other for the compared models such that the status of separation violation remains the same.

\begin{figure} [t!]	
	\centering
	\begin{subfigure}[h]{10cm}
		\centering
		\includegraphics[width=10cm,height=7.5cm,keepaspectratio]{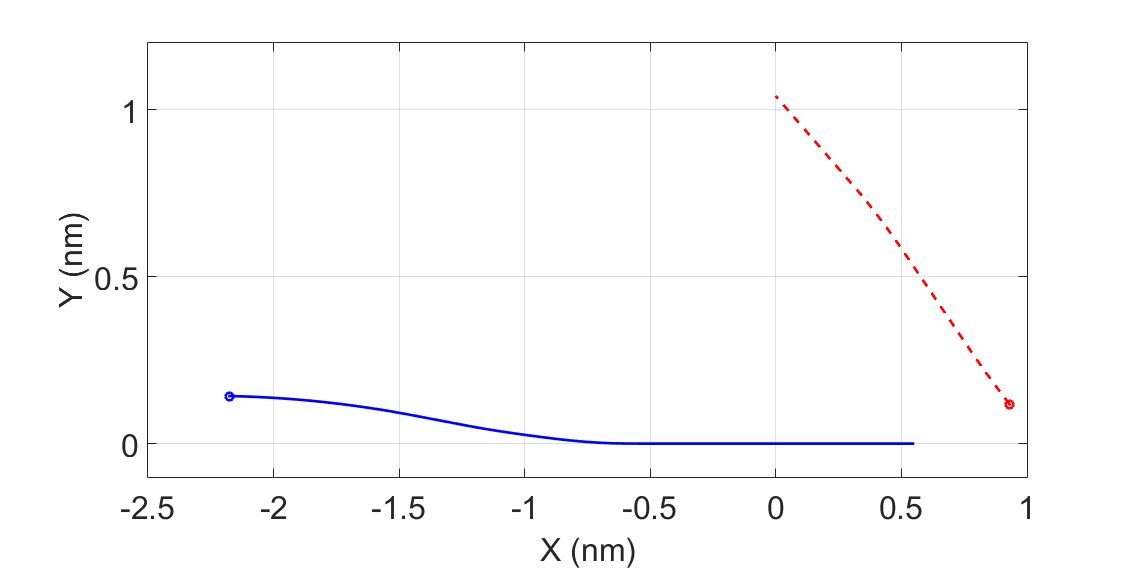}
		\caption{Trajectories created by the validated model.}\label{f:enc3_2}		
	\end{subfigure}
	\quad
	\begin{subfigure}[h]{10cm}
		\centering
		\includegraphics[width=10cm,height=7.5cm,keepaspectratio]{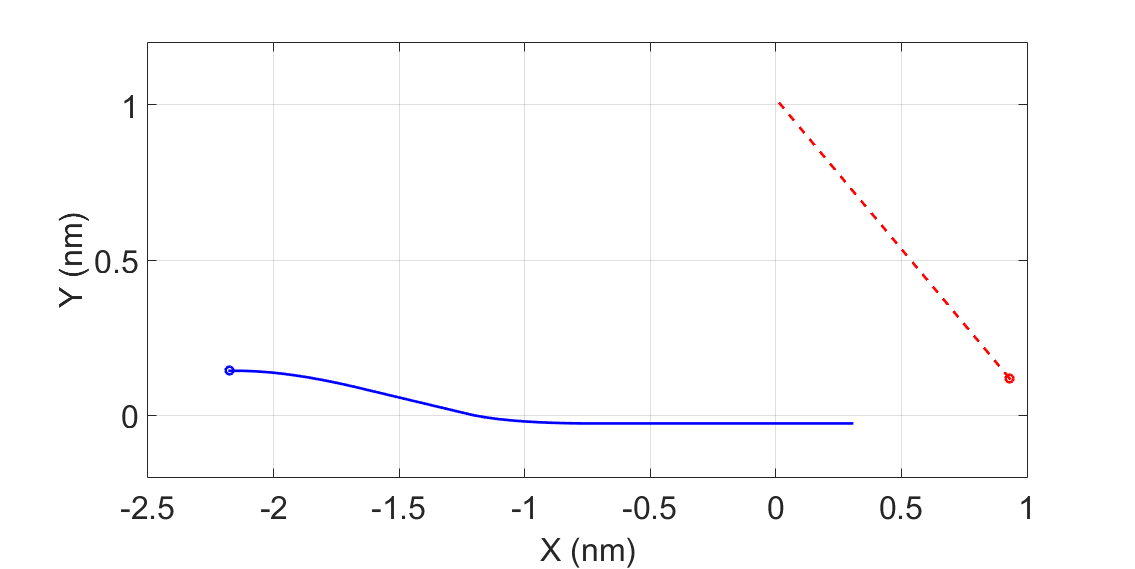}
		\caption{Trajectories created by the proposed model.}\label{f:enc3_1}
	\end{subfigure}
	
	\begin{subfigure}[h]{10cm}
			\centering
			\includegraphics[width=10cm,height=7.5cm,keepaspectratio]{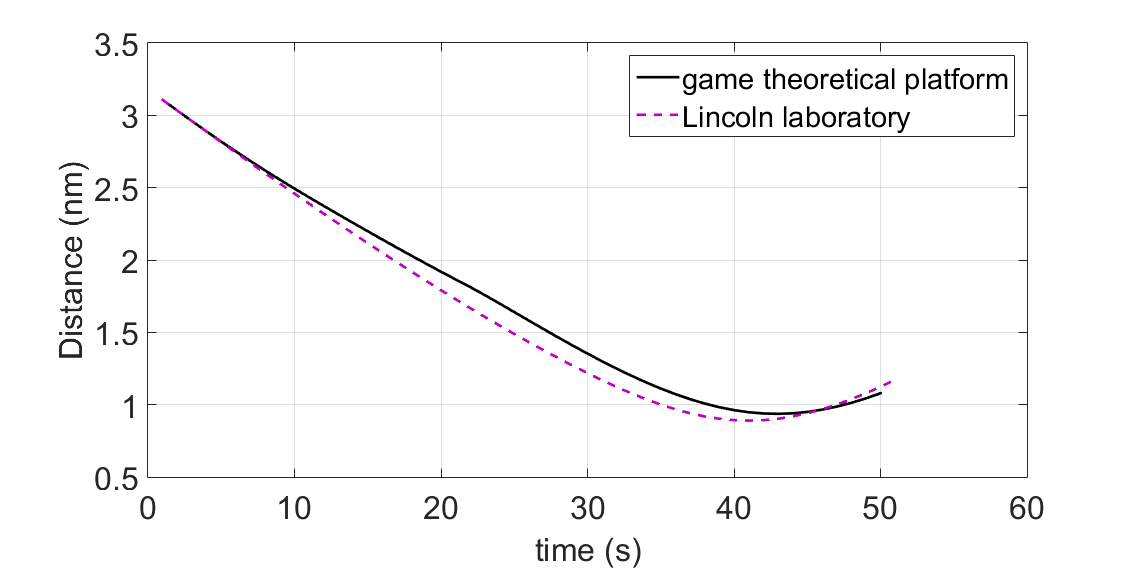}
			\caption{Separation distances for each model.}\label{f:enc3_3}
	\end{subfigure}
	\caption{Comparison of the trajectories created by the validated model and the game theoretical modeling approach for sample encounter number 3.}\label{f:enc3}
	\vspace{1.1cm}
\end{figure}

\begin{figure} [t!]	
	\centering
	\begin{subfigure}[htb]{10cm}
		\centering
		\includegraphics[width=10cm,height=7.5cm,keepaspectratio]{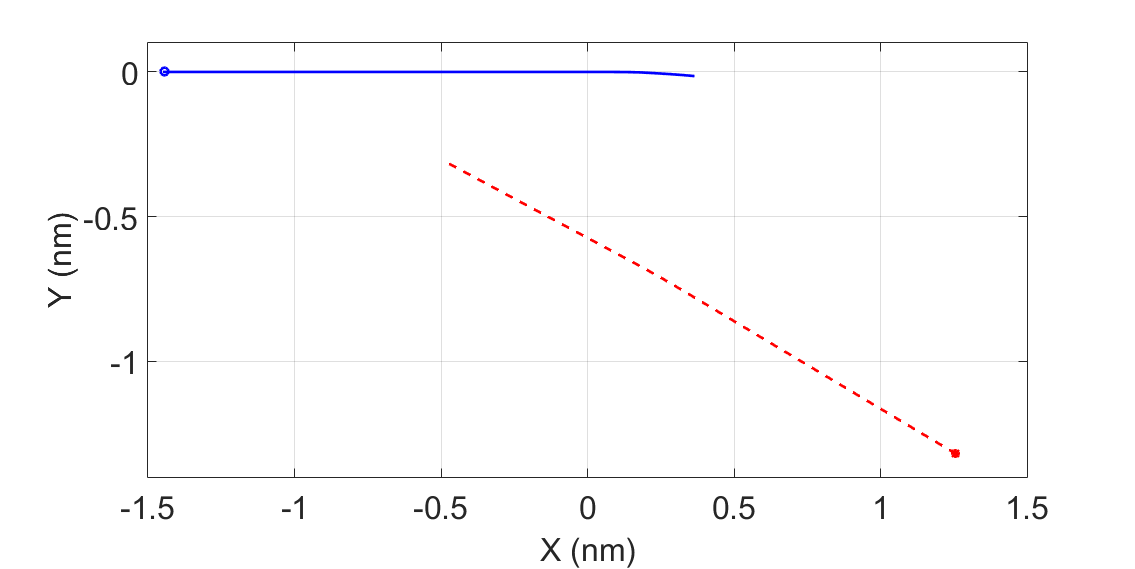}
		\caption{Trajectories created by the validated model.}\label{f:enc16_2}		
	\end{subfigure}
	\quad
	\begin{subfigure}[htb]{10cm}
		\centering
		\includegraphics[width=10cm,height=7.5cm,keepaspectratio]{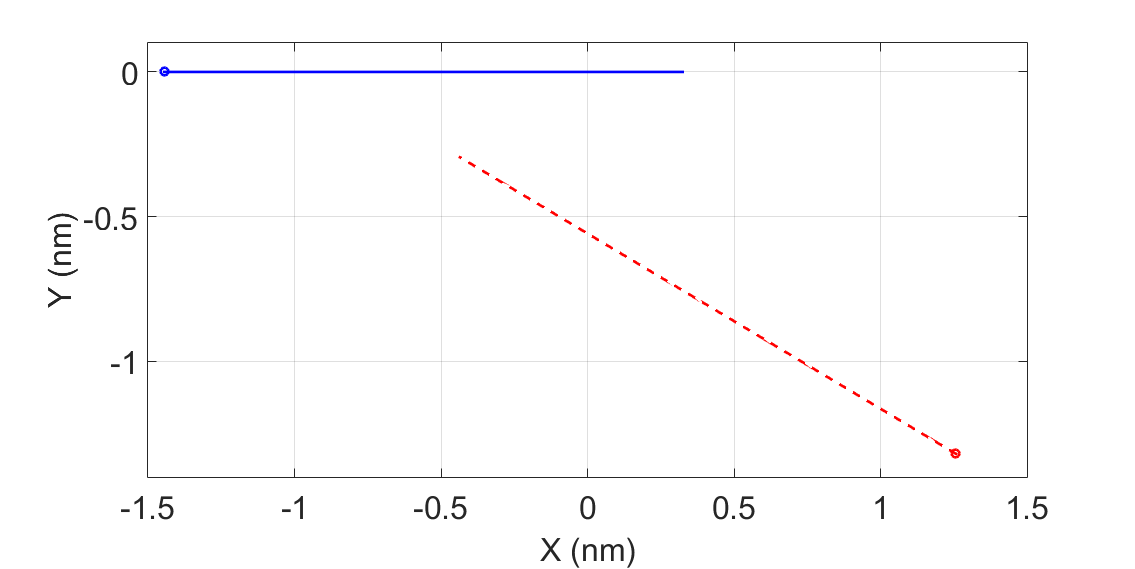}
		\caption{Trajectories created by the proposed model.}\label{f:enc16_1}
	\end{subfigure}
	\quad
	\begin{subfigure}[htb]{10cm}
			\centering
			\includegraphics[width=10cm,height=7.5cm,keepaspectratio]{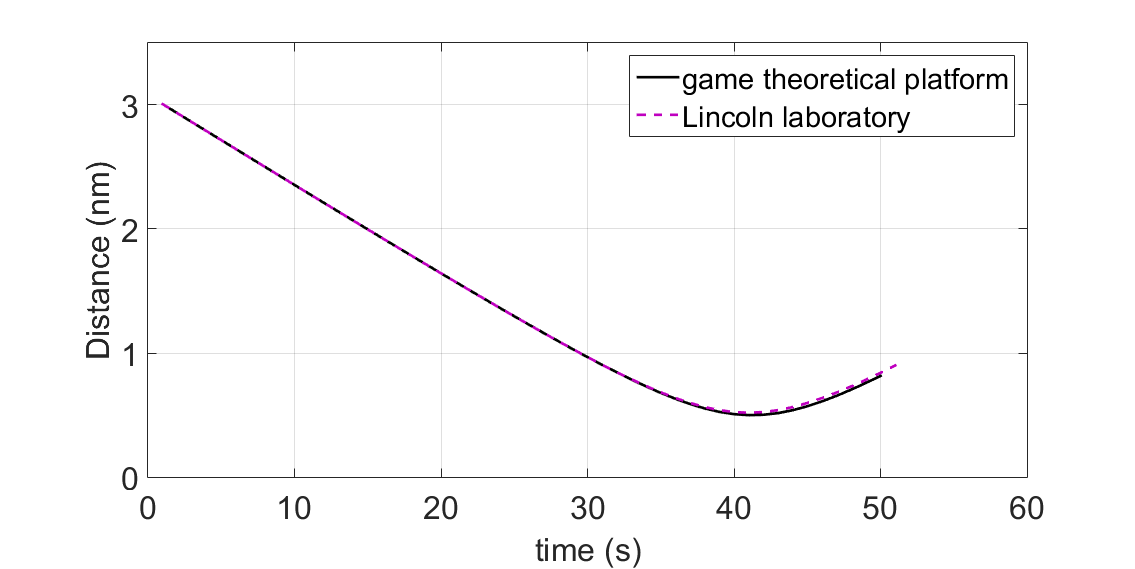}
			\caption{Separation distances for each model.}\label{f:enc16_3}
	\end{subfigure}
	\caption{Comparison of the trajectories created by the validated model and the game theoretical modeling approach for sample encounter number 16.}\label{f:enc16}
\end{figure}

\begin{figure} [t!]	
	\centering
	\begin{subfigure}[htb]{10cm}
		\centering
		\includegraphics[width=10cm,height=7.5cm,keepaspectratio]{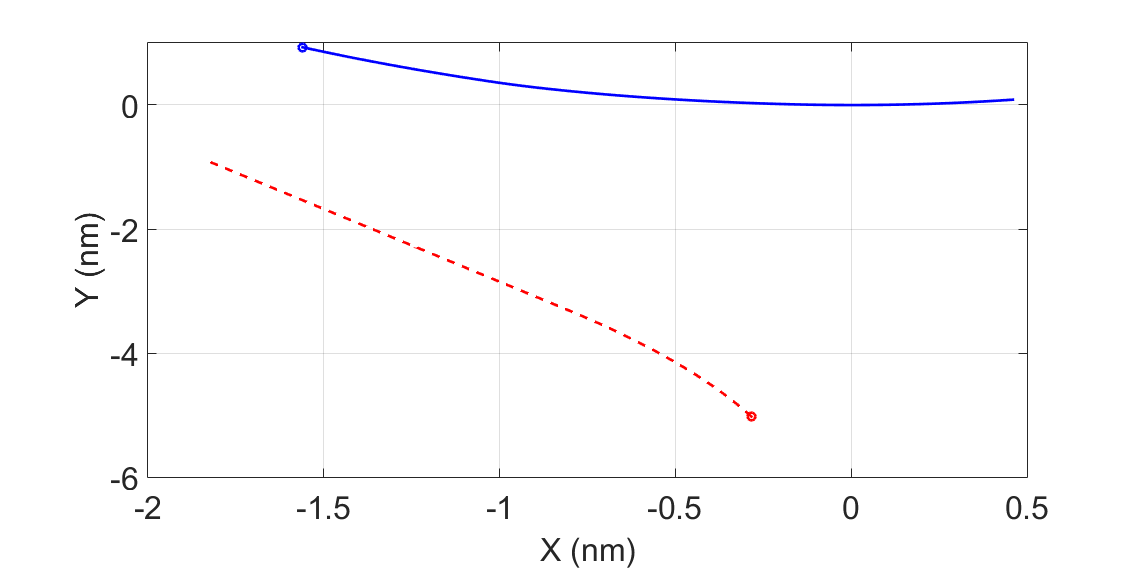}
		\caption{Trajectories created by the validated model.}\label{f:enc23_2}		
	\end{subfigure}
	\quad
	\begin{subfigure}[htb]{10cm}
		\centering
		\includegraphics[width=10cm,height=7.5cm,keepaspectratio]{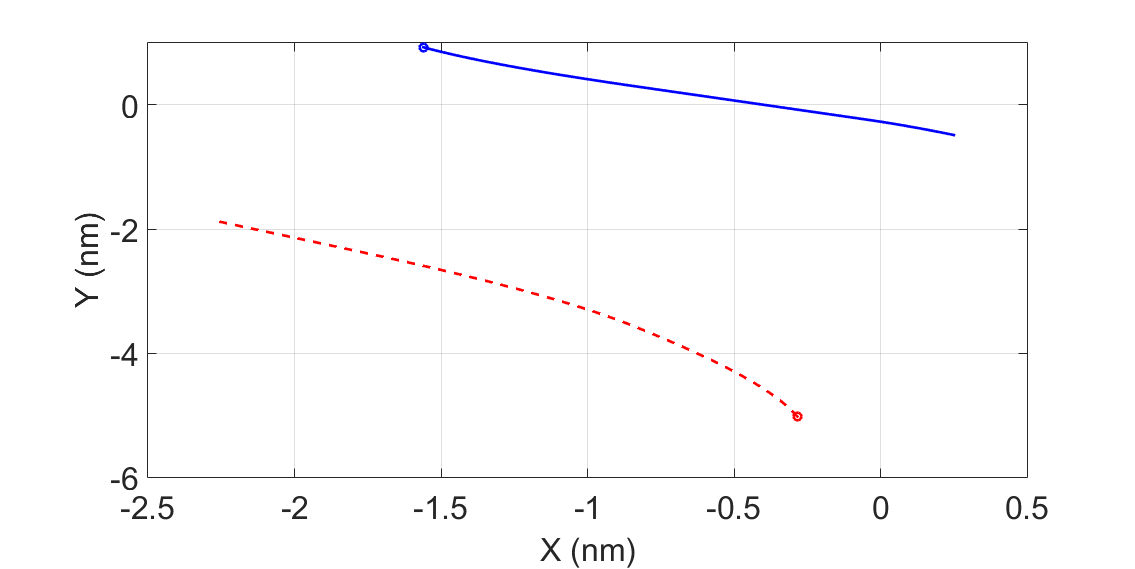}
		\caption{Trajectories created by the proposed model.}\label{f:enc23_1}
	\end{subfigure}
	\quad
	\begin{subfigure}[htb]{10cm}
			\centering
			\includegraphics[width=10cm,height=7.5cm,keepaspectratio]{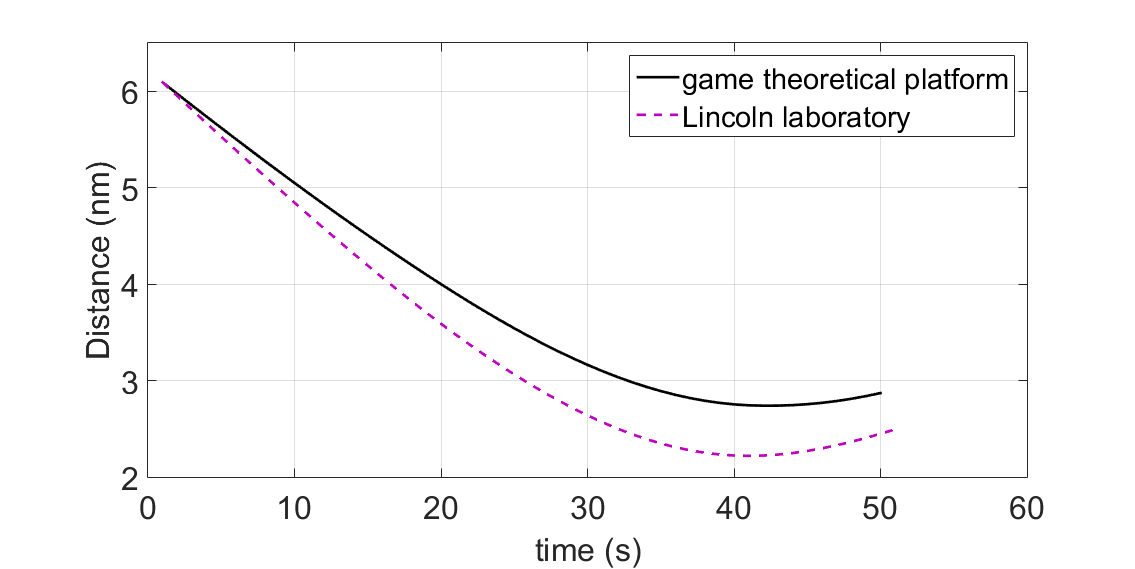}
			\caption{Separation distances for each model.}\label{f:enc23_3}
	\end{subfigure}
	\caption{Comparison of the trajectories created by the validated model and the game theoretical modeling approach for sample encounter number 23.}\label{f:enc23}
	\vspace{1.1cm}
\end{figure}

\begin{figure} [t!]	
	\centering
	\begin{subfigure}[htb]{10cm}
		\centering
		\includegraphics[width=10cm,height=7.5cm,keepaspectratio]{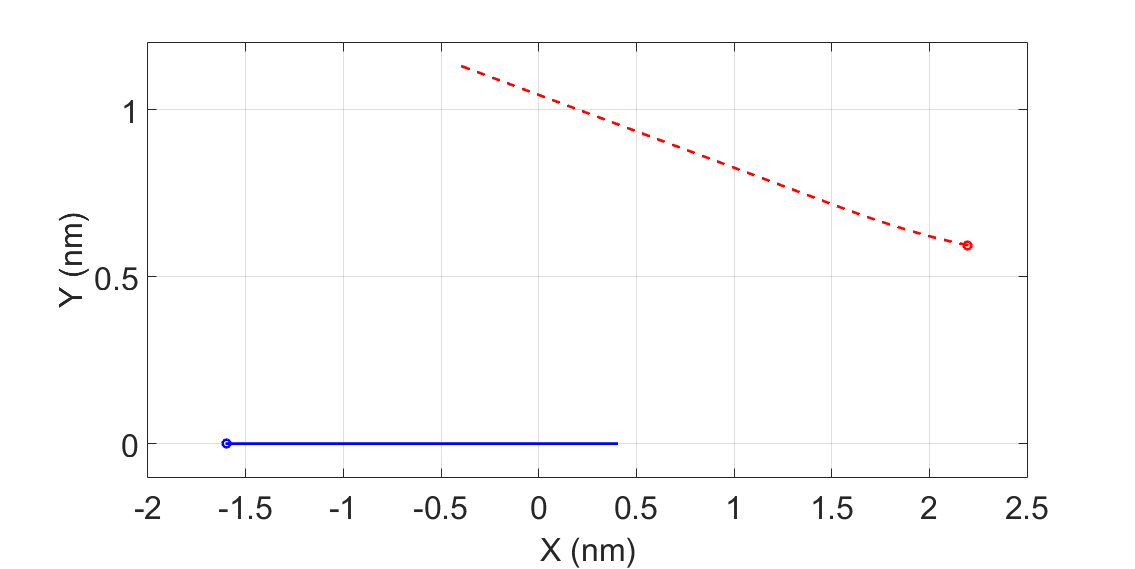}
		\caption{Trajectories created by the validated model.}\label{f:enc34_2}		
	\end{subfigure}
	\quad
	\begin{subfigure}[htb]{10cm}
		\centering
		\includegraphics[width=10cm,height=7.5cm,keepaspectratio]{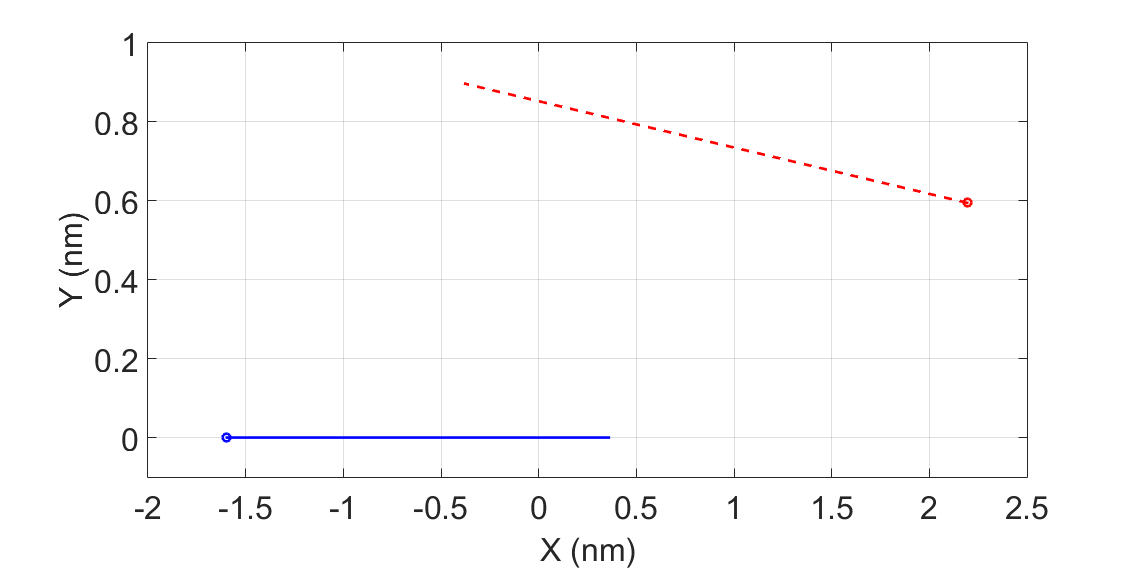}
		\caption{Trajectories created by the proposed model.}\label{f:enc34_1}
	\end{subfigure}
	\quad
	\begin{subfigure}[htb]{10cm}
			\centering
			\includegraphics[width=10cm,height=7.5cm,keepaspectratio]{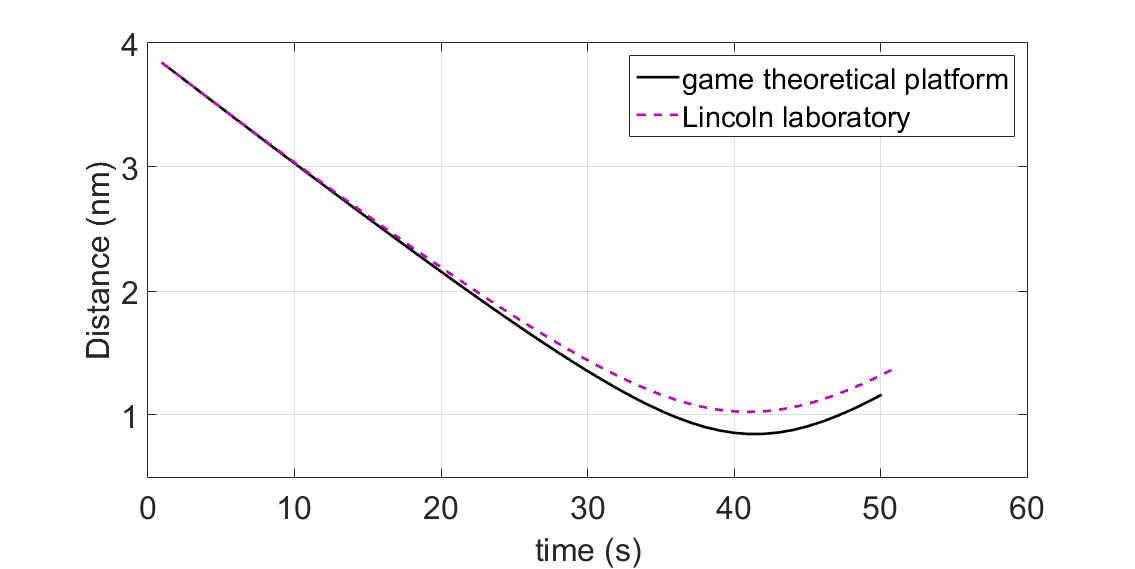}
			\caption{Separation distances for each model.}\label{f:enc34_3}
	\end{subfigure}
	\caption{Comparison of the trajectories created by the validated model and the game theoretical modeling approach for sample encounter number 34.}\label{f:enc34}
\end{figure}

\begin{figure} [t!]	
	\centering
	\begin{subfigure}[htb]{10cm}
		\centering
		\includegraphics[width=10cm,height=7.5cm,keepaspectratio]{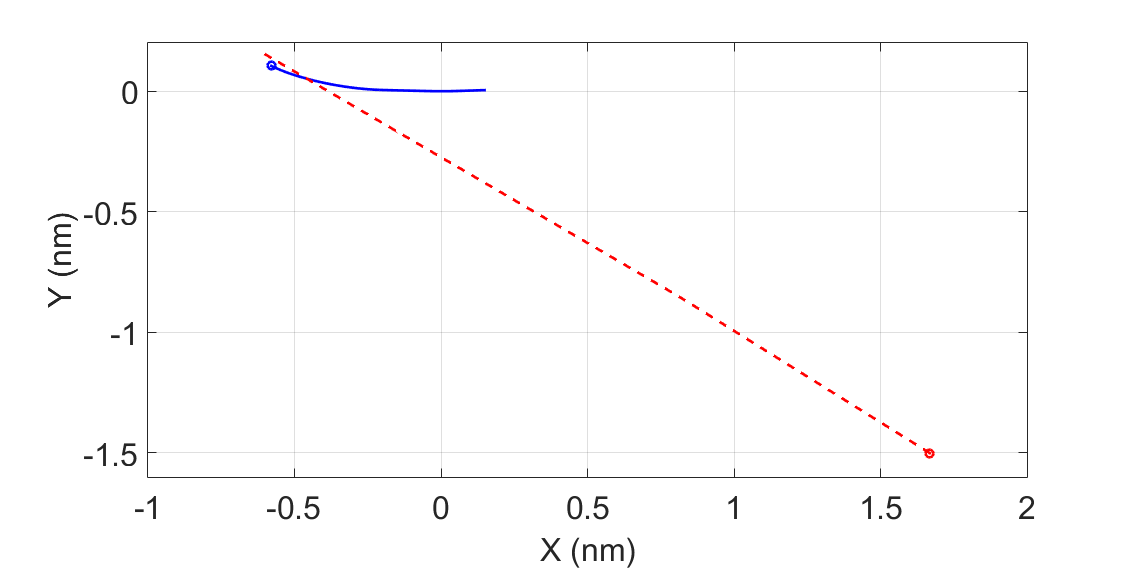}
		\caption{Trajectories created by the validated model.}\label{f:enc45_2}		
	\end{subfigure}
	\quad
	\begin{subfigure}[htb]{10cm}
		\centering
		\includegraphics[width=10cm,height=7.5cm,keepaspectratio]{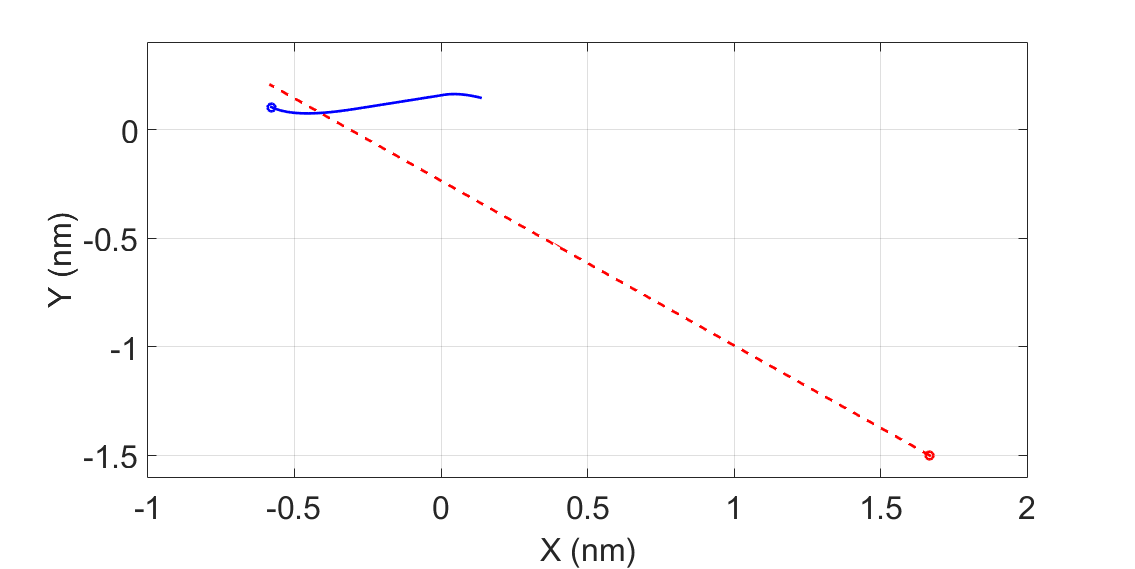}
		\caption{Trajectories created by the proposed model.}\label{f:enc45_1}
	\end{subfigure}
	\quad
	\begin{subfigure}[htb]{10cm}
			\centering
			\includegraphics[width=10cm,height=7.5cm,keepaspectratio]{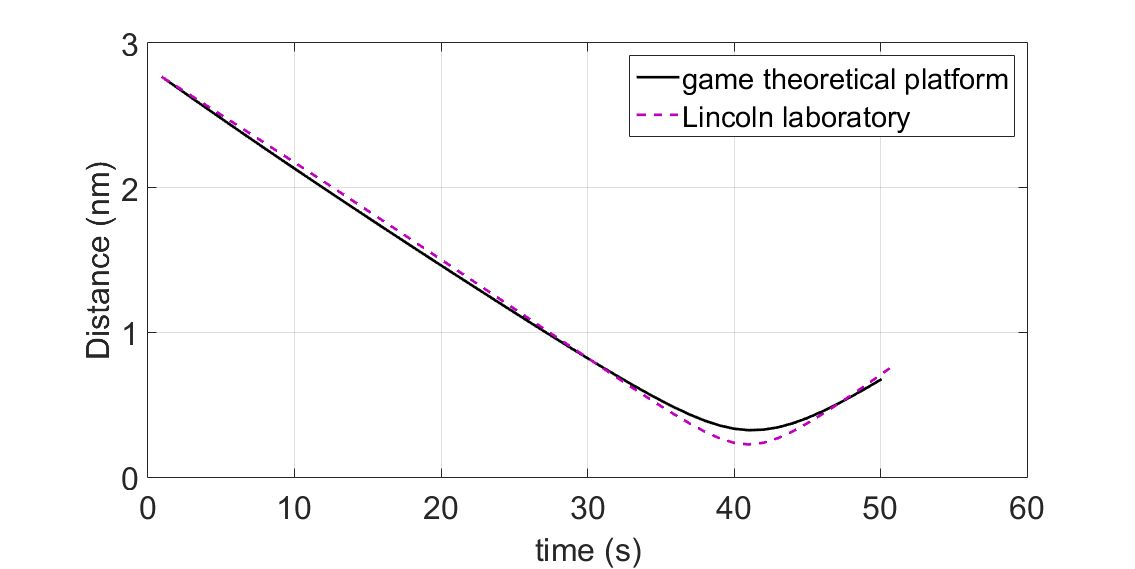}
			\caption{Separation distances for each model.}\label{f:enc45_3}
	\end{subfigure}
	\caption{Comparison of the trajectories created by the validated model and the game theoretical modeling approach for sample encounter number 45.}\label{f:enc45}
\vspace{0.5cm}
\end{figure}

Fig.~\ref{f:enc3_2} demonstrates aircraft trajectories during encounter number 3, listed in \texttt{cor\_ac1.txt}. In Fig.~\ref{f:enc3_1}, the same encounter is regenerated using the proposed game theoretical model by assigning the same initial positions, initial heading angles and initial speeds to two aircraft. It is seen that although the trajectories are not exactly the same, pilot decisions determined by the game theoretical model are similar to the decisions provided by the validated model. In addition, according to Fig.~\ref{f:enc3_3}, the minimum separation distance is experienced after about $40s$ and $42s$ for the validated model and the game theoretical model, respectively, and the difference between the minimum distances is about $0.05nm$. In this example, the pilots represented by the solid blue and dashed red curves are modeled as level-1 and level-0 pilots, respectively. Figs.~\ref{f:enc16}-\ref{f:enc45} also show similar characteristics where the encounter trajectories and pilot decisions are similar for the validated and proposed models.  

\textbf{Face validation:} ``Face validation'' is a validation method used for models that are developed for systems that are expected to be a reality in the future, such as UAS integration models \cite{MITRE_VV:16}. In this method, two aspects of the model are evaluated: 1) Is the logic in the conceptual model correct? and 2) Are the input-output relationships of the model reasonable? The core ideas of the proposed framework, such as level-k game theoretical concept, reinforcement learning and bounded rationality, are supported by several references earlier in this study. In addition, the logic of the objective function utilized during the modeling process is detailed where it is seen that the choice of the terms is logical. Finally, in the simulation results section of this work, input-output relationships of the model are discussed in length to show that they represent reasonable system behavior. Therefore, the steps needed for the face validation of the proposed method are completed.  

\textbf{Remark:} It is noted that without collecting, processing and analyzing real HAS data, which may be available in the near future, and carefully comparing the outputs with the model, using available statistical validation tools (see \cite{Law:08}), the validation of the model can not be accepted as completed. It is important for a model to have enough degrees of freedom so that when discrepancies with the real data are detected the model can be modified accordingly to obtain a match with the data with reasonable accuracy \cite{Law:08}. In this regard, the proposed game theoretical framework is a strong candidate for a successful UAS integration model since it contains several degrees of freedom such as the objective function weights representing the importance of each term. In addition, the modular structure of the objective function allows the designers to add/subtract terms to achieve an agreement with the data.

\section{Simulation Results and Discussion} \label{Simulation Results and Discussion}  
In this section, the results of a quantitative analysis of a simulation for UAS integration scenarios are presented. Before showing the results for the scenario explained in the previous section, single encounter scenarios, where a single UAS and a single manned aircraft are in a collision path, are investigated. Later, the results for the scenario with multiple encounters in a crowded airspace are shown. A quantitative comparison between the two SAA algorithms in terms of their performance and safety is also presented.

\subsection{Hybrid Airspace Scenarios with a Single Encounter} \label{Single Encounter Scenario}
\begin{figure}[t!]	
	\centering
	\begin{subfigure}[htb]{6cm}
		\centering
		\includegraphics[width=6cm]{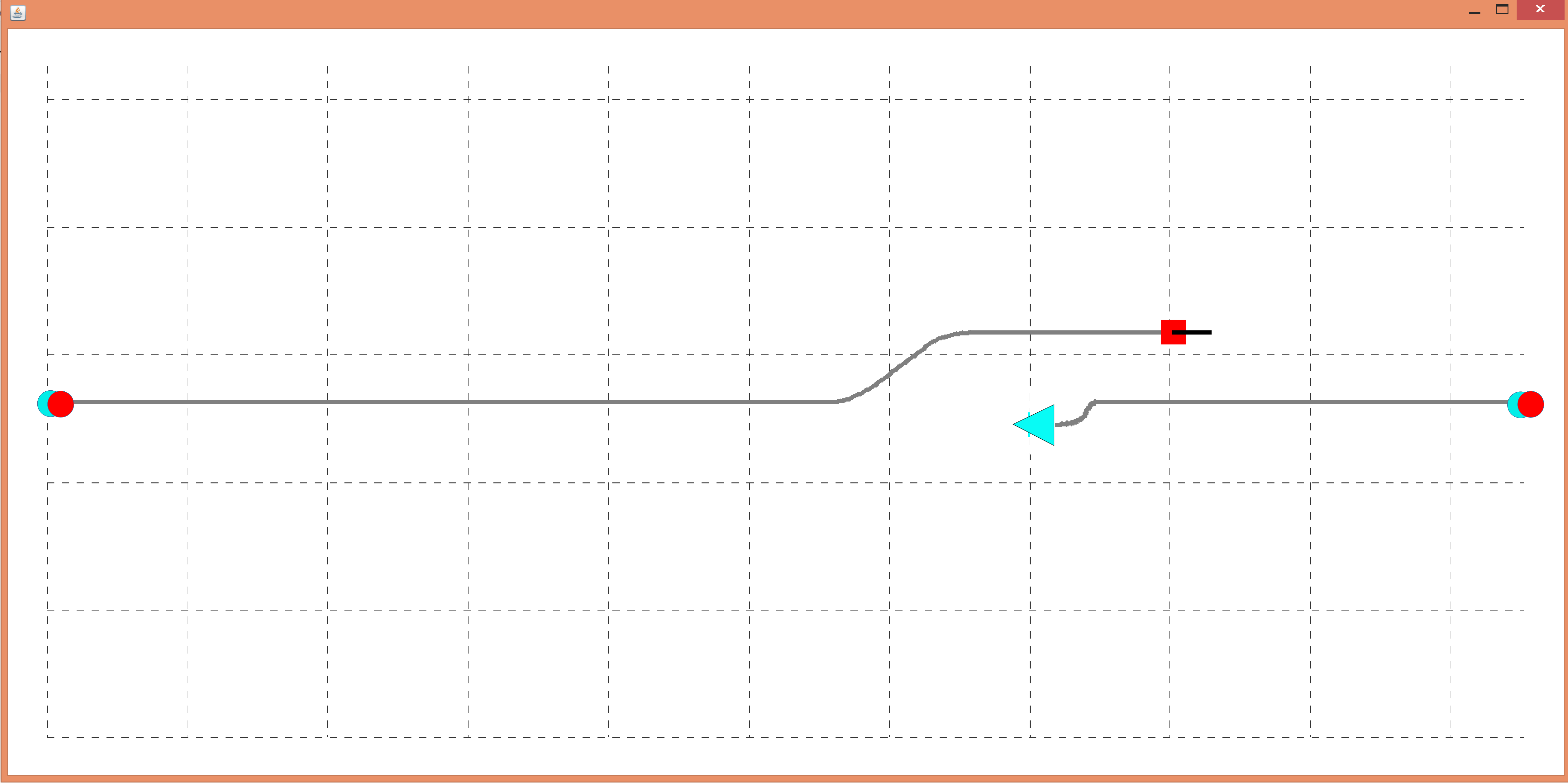}
		\caption{level-1 pilot, approach angle $45^{\circ}$}\label{f:ses4}		
	\end{subfigure}
	\quad
	\begin{subfigure}[htb]{6cm}
		\centering
		\includegraphics[width=6cm]{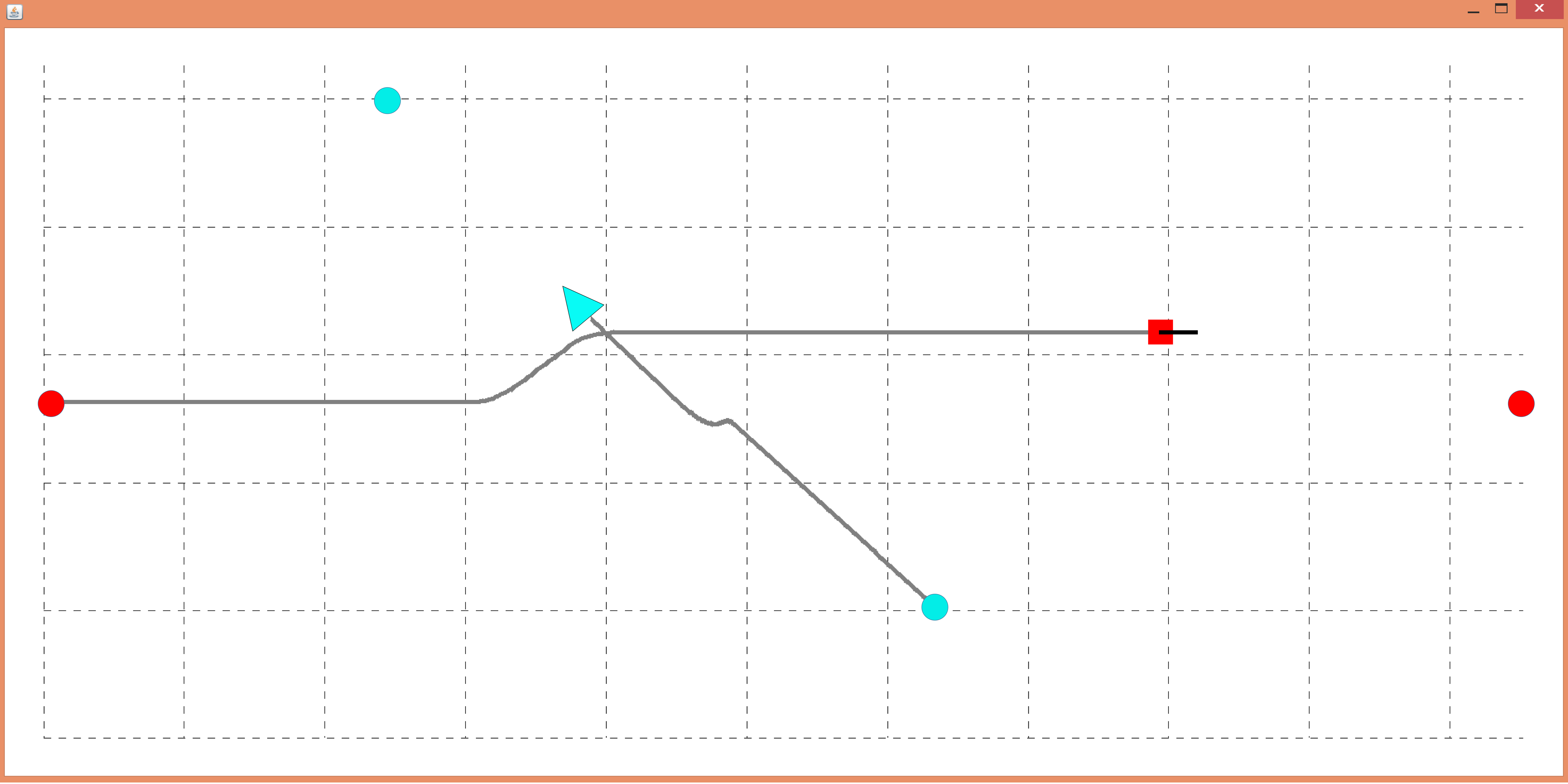}
		\caption{level-1 pilot, approach angle $90^{\circ}$}\label{f:ses3}
	\end{subfigure}
	\quad
	\begin{subfigure}[htb]{6cm}
			\centering
			\includegraphics[width=6cm]{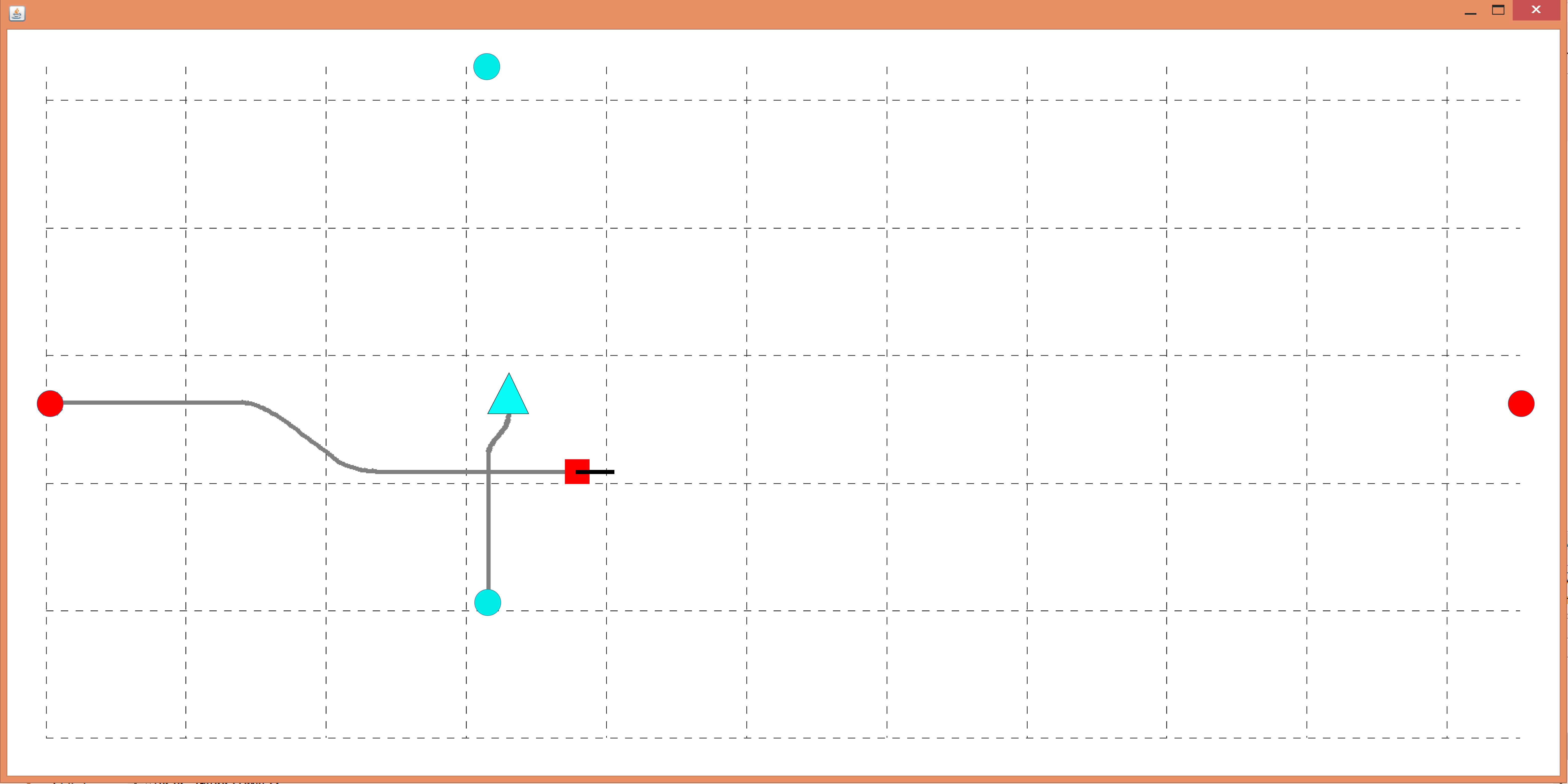}
			\caption{level-1 pilot, approach angle $135^{\circ}$}\label{f:ses2}
	\end{subfigure}
	\quad
	\begin{subfigure}[htb]{6cm}
			\centering
			\includegraphics[width=6cm]{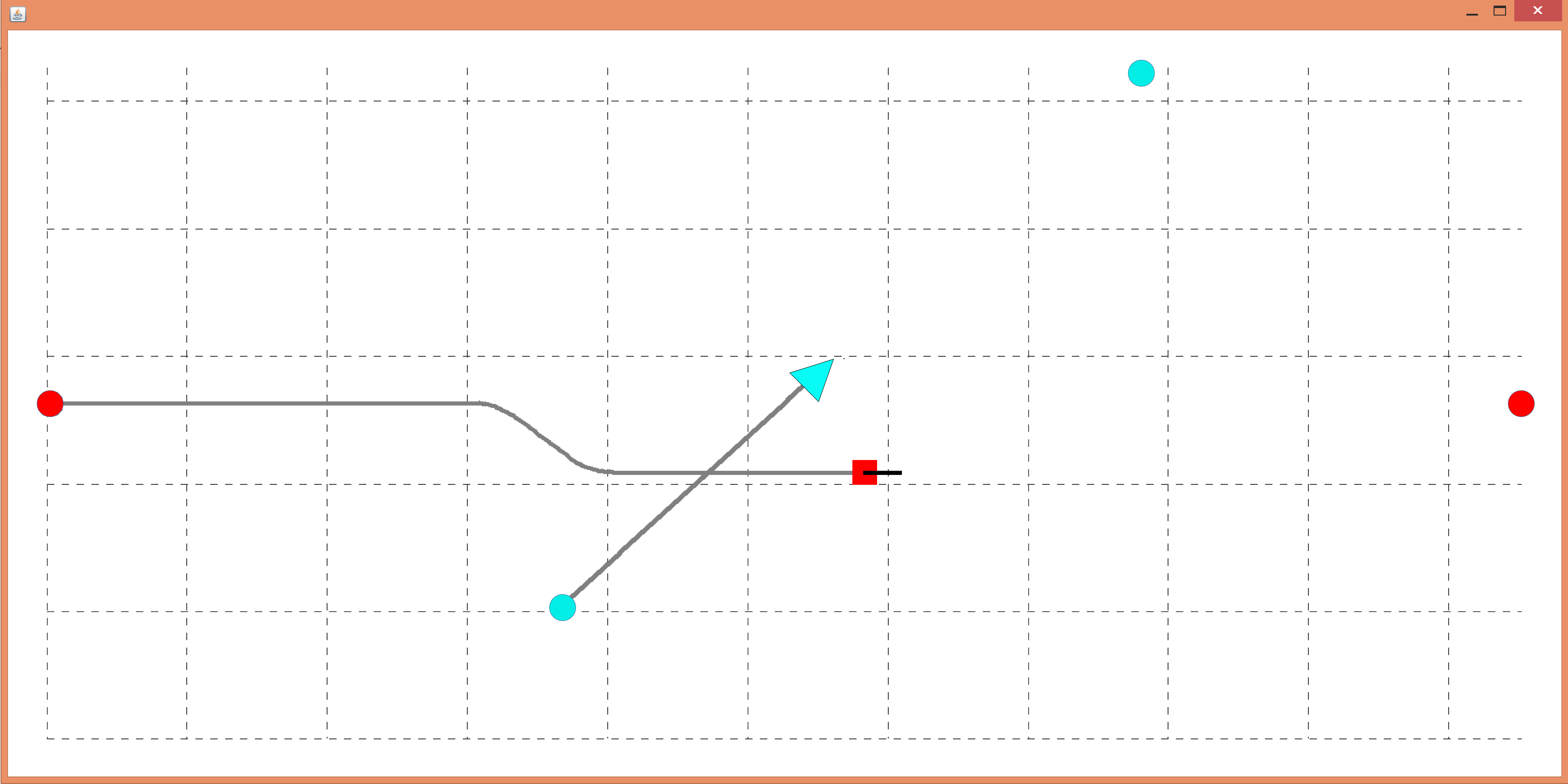}
			\caption{level-1 pilot, approach angle $180^{\circ}$}\label{f:ses1}
	\end{subfigure}
	\quad
	\begin{subfigure}[htb]{6cm}
		\centering
		\includegraphics[width=6cm]{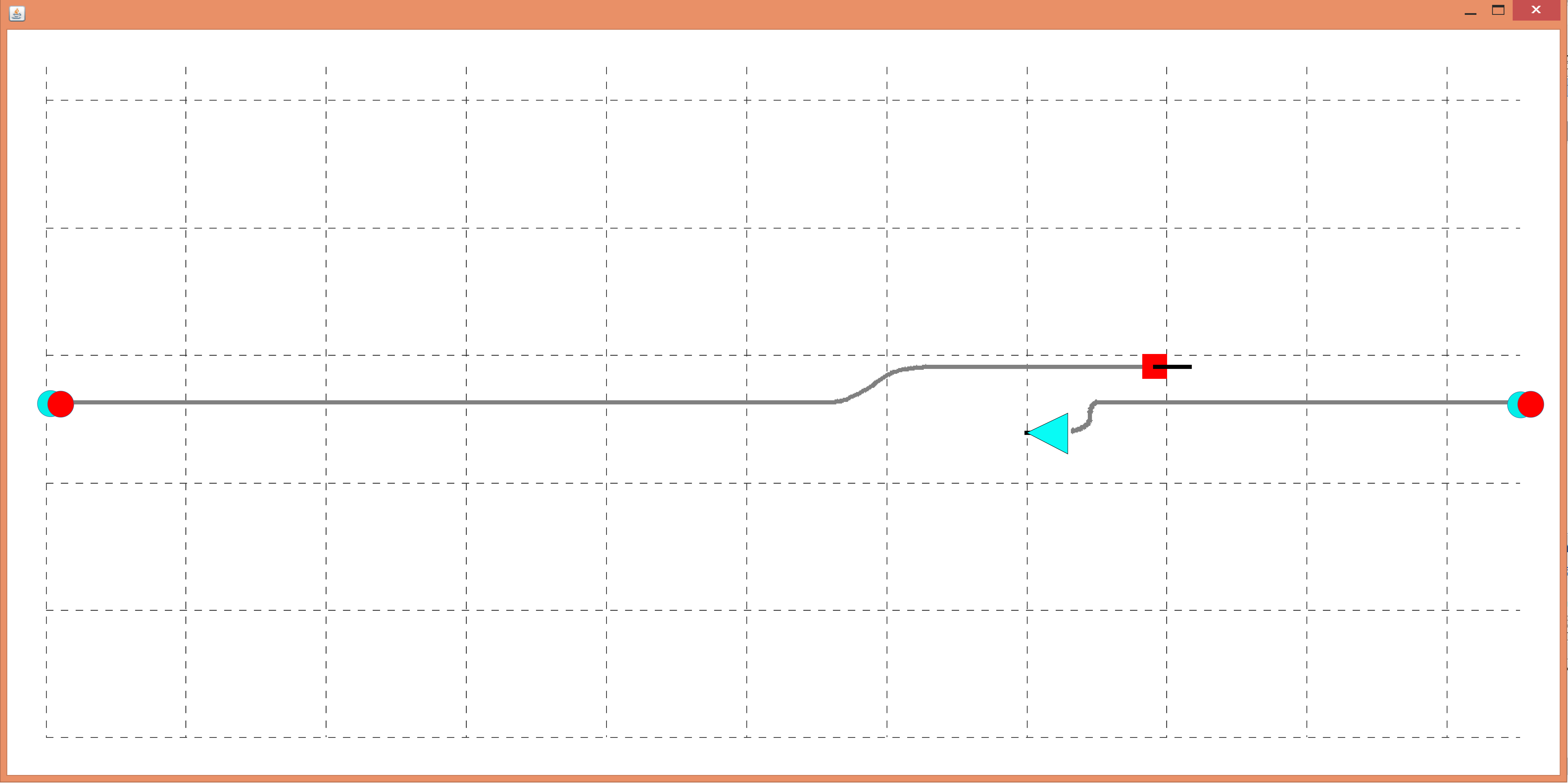}
		\caption{level-2 pilot, approach angle $45^{\circ}$}\label{f:ses8}		
	\end{subfigure}
	\quad
	\begin{subfigure}[htb]{6cm}
		\centering
		\includegraphics[width=6cm]{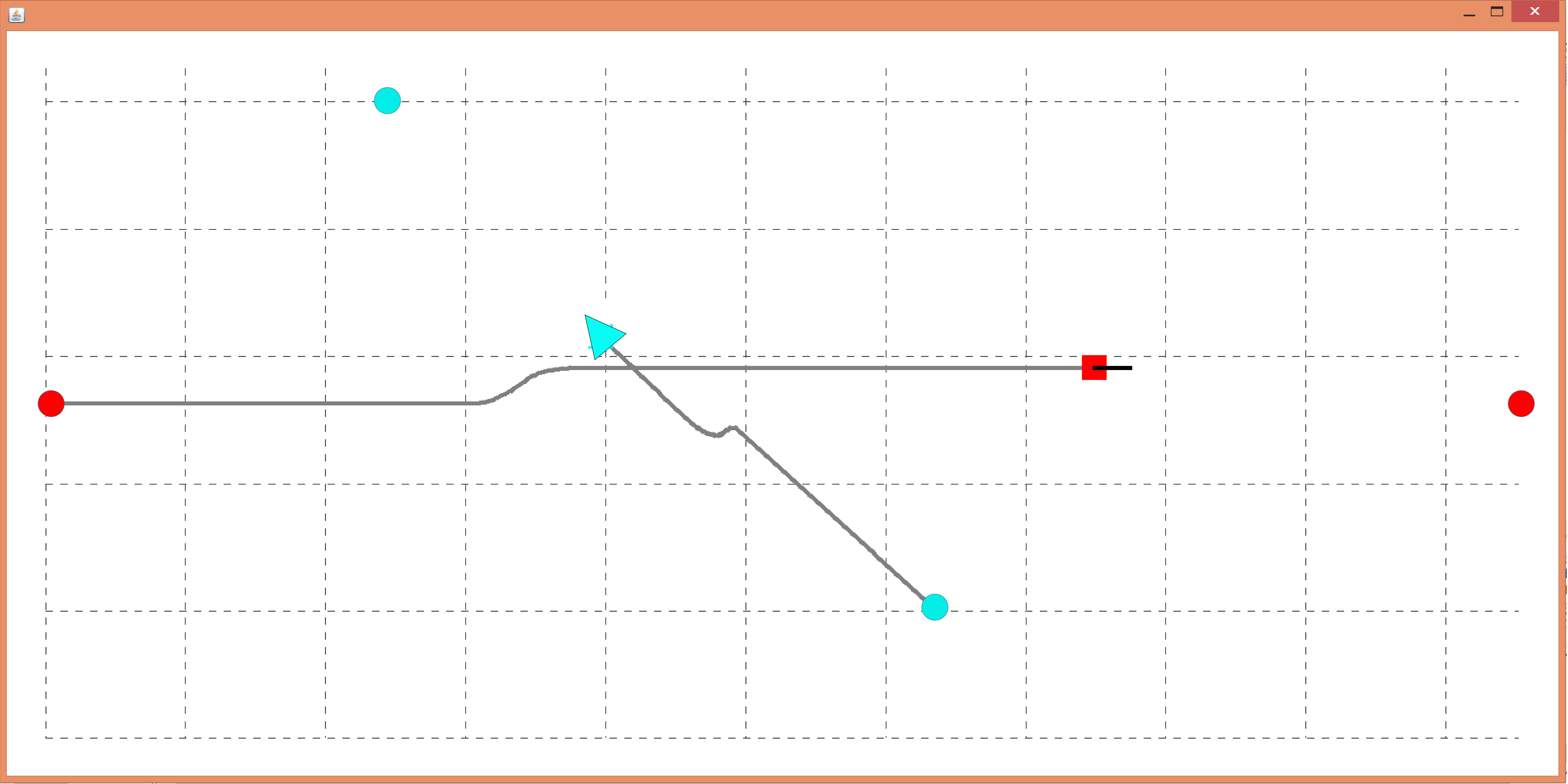}
		\caption{level-2 pilot, approach angle $90^{\circ}$}\label{f:ses7}
	\end{subfigure}
	\quad
	\begin{subfigure}[htb]{6cm}
			\centering
			\includegraphics[width=6cm]{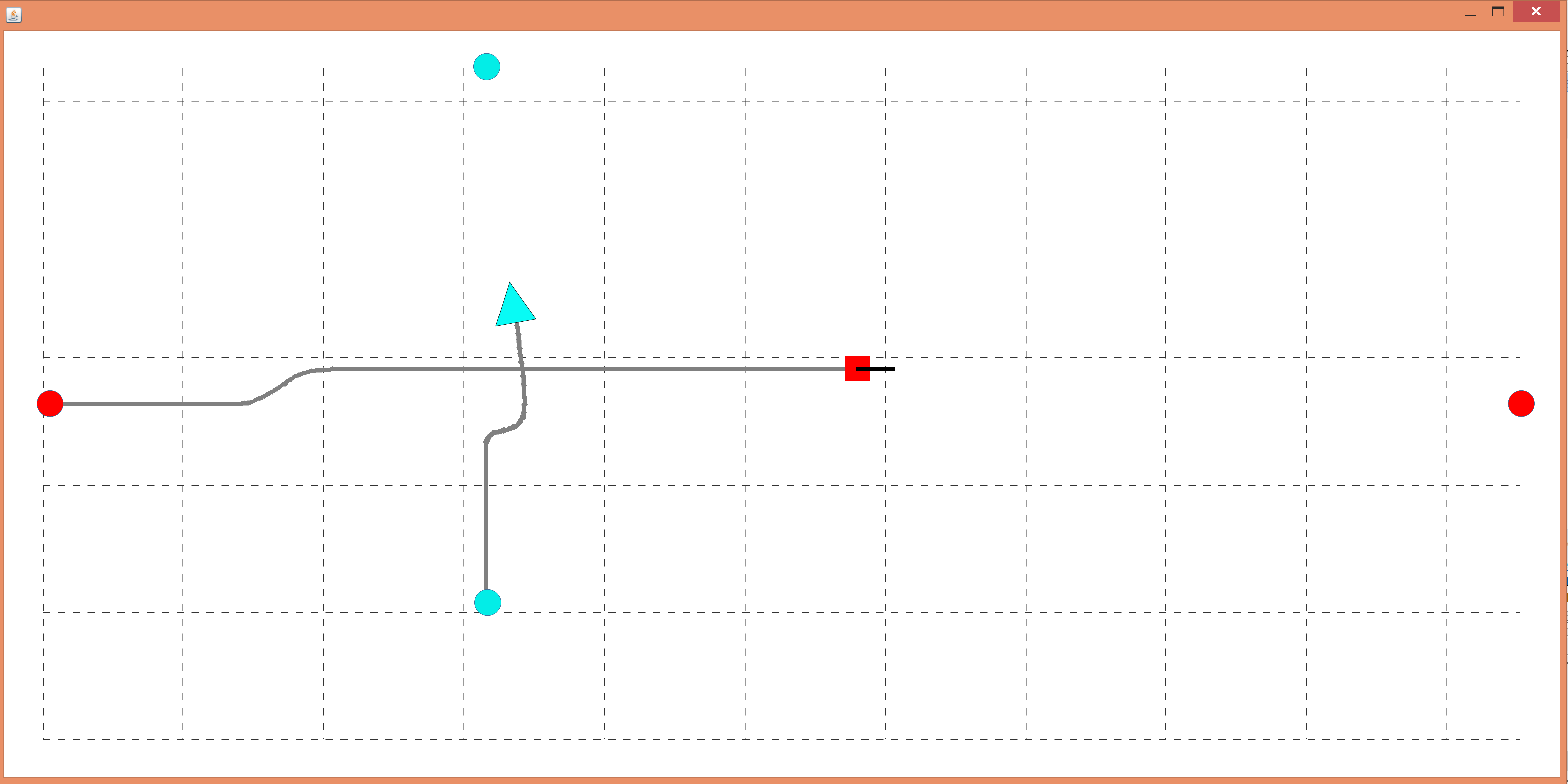}
			\caption{level-2 pilot, approach angle $135^{\circ}$}\label{f:ses6}
	\end{subfigure}
	\quad
	\begin{subfigure}[htb]{6cm}
			\centering
			\includegraphics[width=6cm]{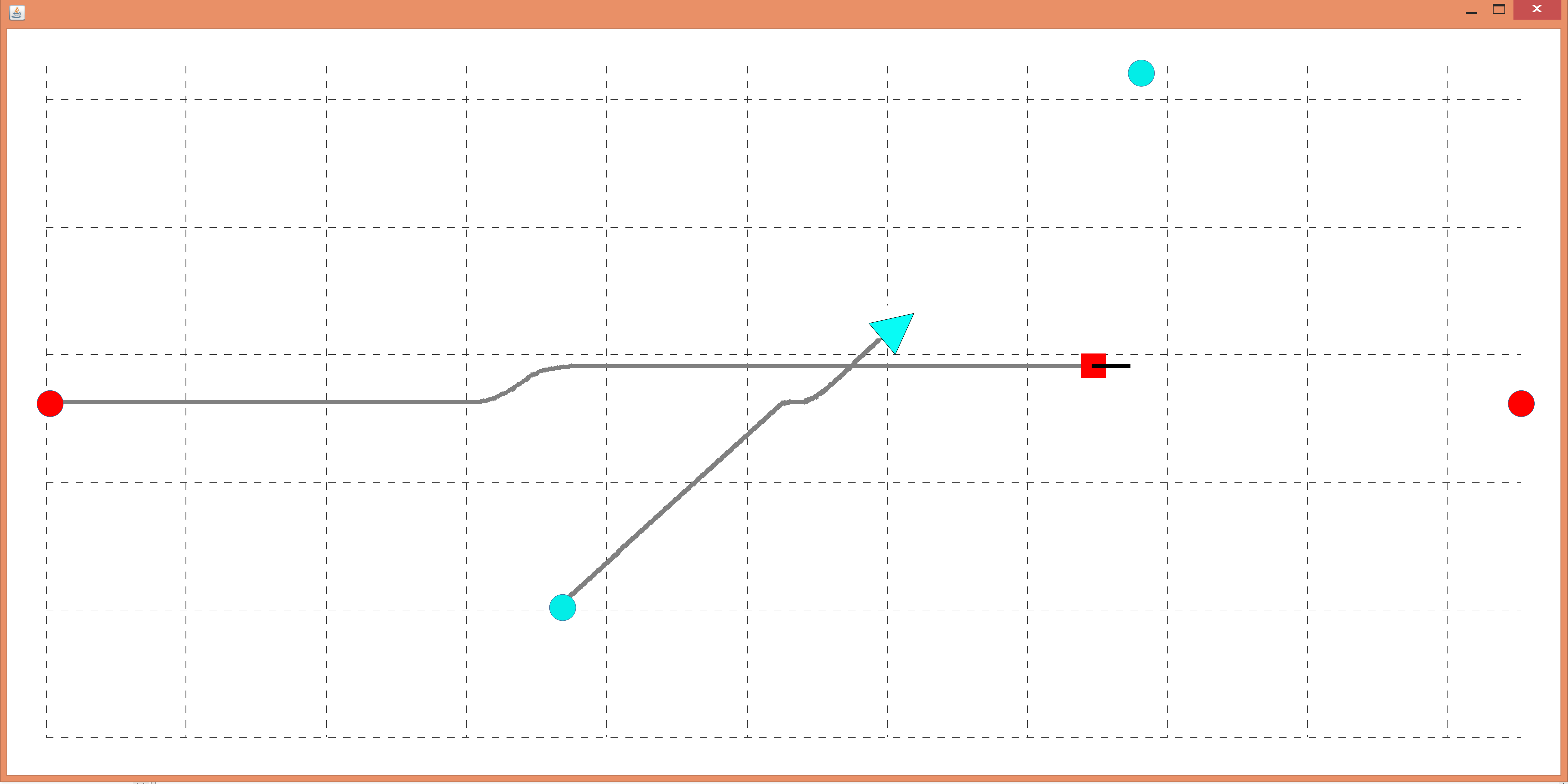}
			\caption{level-2 pilot, approach angle $180^{\circ}$}\label{f:ses5}
	\end{subfigure}
	\caption{Level-1 and level-2 pilots interacting with UAS.}\label{f:ses}
\end{figure}

\begin{figure}[t!]	
	\centering
	\begin{subfigure}[htb]{10cm}
			\centering
			\includegraphics[width=10cm,height=5cm,keepaspectratio]{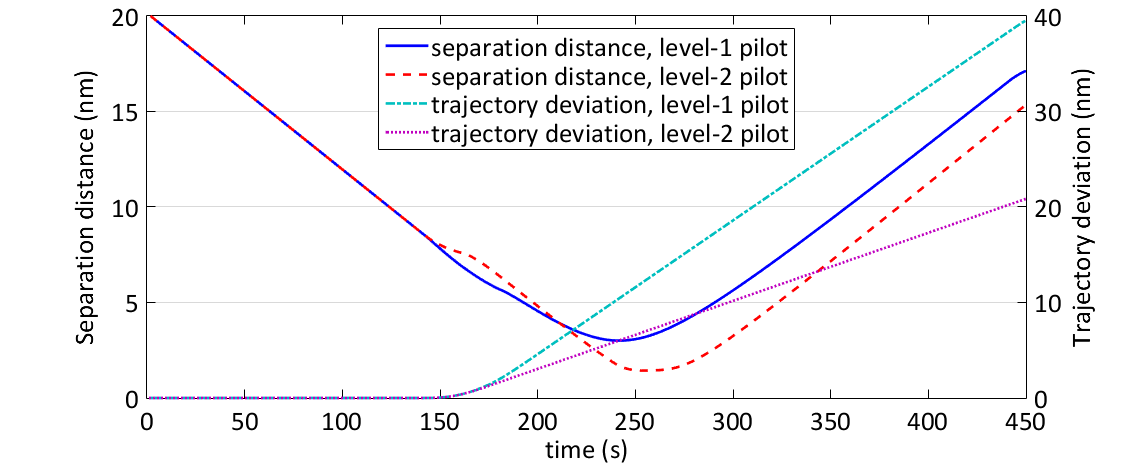}
		\caption{approach angle $45^{\circ}$}\label{f:se1}		
	\end{subfigure}
	\quad
	\begin{subfigure}[htb]{10cm}
			\centering
			\includegraphics[width=10cm,height=5cm,keepaspectratio]{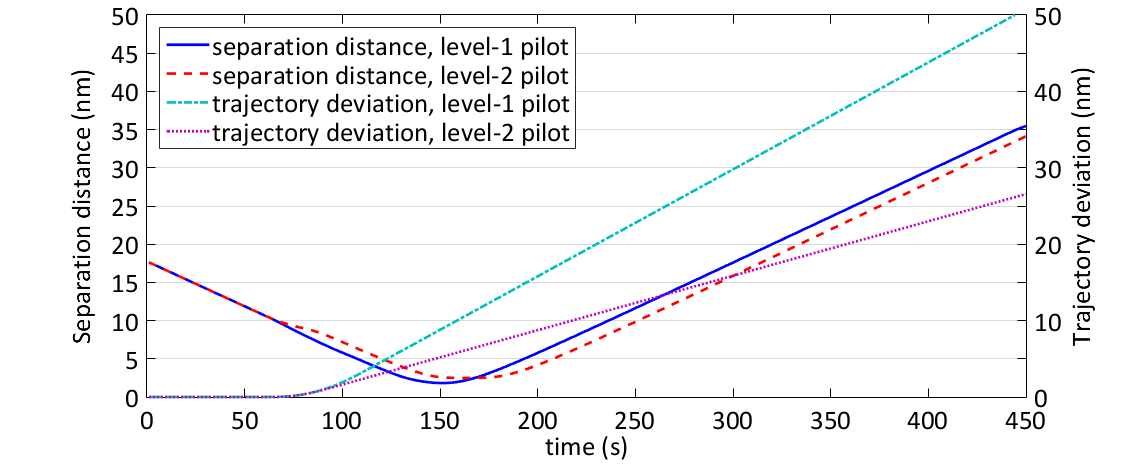}
		\caption{approach angle $90^{\circ}$}\label{f:se2}
	\end{subfigure}
	\quad 
	\begin{subfigure}[htb]{10cm}
			\centering
			\includegraphics[width=10cm,height=5cm,keepaspectratio]{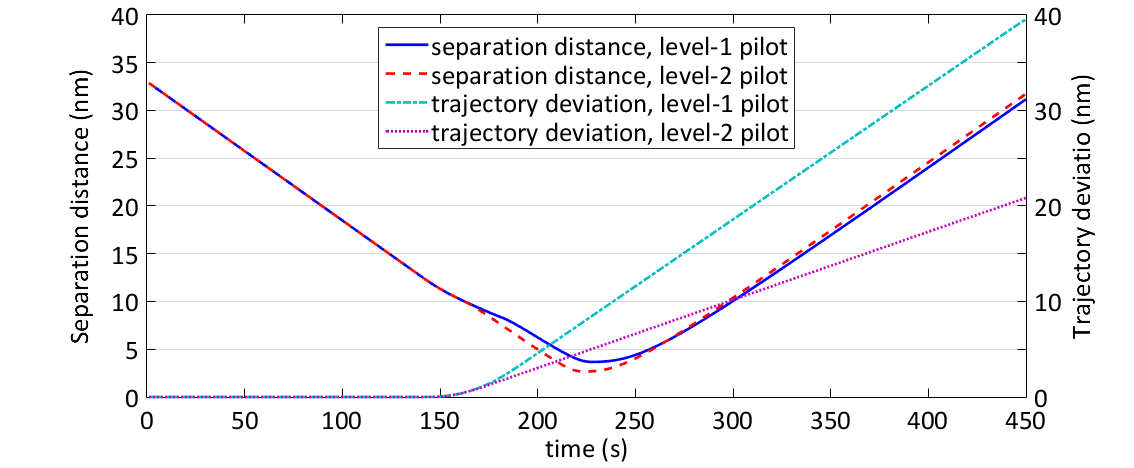}
			\caption{approach angle $135^{\circ}$}\label{f:se3}
	\end{subfigure}
	\quad
	\begin{subfigure}[htb]{10cm}
			\centering
			\includegraphics[width=10cm,height=5cm,keepaspectratio]{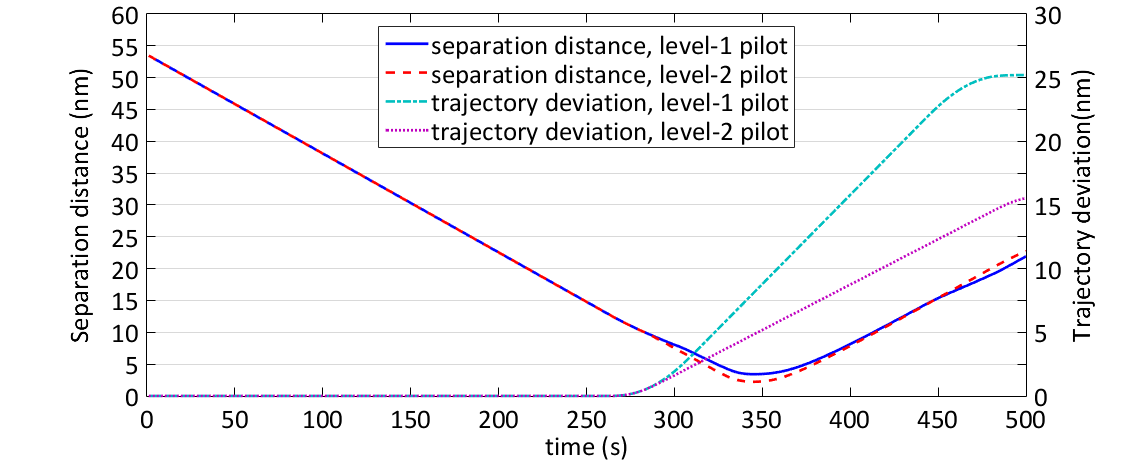}
			\caption{approach angle $180^{\circ}$}\label{f:se4}
	\end{subfigure}
	\caption{Safety vs. performance of Level-k pilot interacting with UAS}\label{f:se}
\end{figure}

In order to investigate the reactions of a level-k pilot during a conflict with an UAS, 4 single encounter scenarios are designed. In these 4 scenarios, level-1 and level-2 policies are used for the manned aircraft pilots and the UAS follows the guidelines of the SAA1 algorithm which may command velocity adjustments in order for the UAS to avoid the conflict. Apart from pilot levels, the effect of different approach angles, which take the values of $45^{\circ}$, $90^{\circ}$, $135^{\circ}$ and $180^{\circ}$, are also investigated. Figure~\ref{f:ses} depicts the snapshots of four cases, where the red square corresponds to the manned aircraft and cyan triangle corresponds to the UAS. The gray track line right behind the manned aircraft and the UAS represent their traveled path from their initial positions to where they stand in the snapshot. Circles show the initial positions and destinations. The geometric size of the scenarios is $100km\times50km$. In all cases, the manned aircraft and the UAS are heading toward a conflict which is detected by both the UAS, via its SAA system, and the pilot $20s$ prior to a probable miss separation. A miss separation is declared when the relative distance becomes less than $5nm$. The pilot then starts an evasive maneuver based on the level-1 and the level-2 reasoning policies, and the UAS implements its own evasive maneuver based on the SAA1 system, whose working principles are explained in Section~\ref{Sense and Avoid Algorithm}. Figure~\ref{f:se} depicts the separation distance and trajectory deviations during these single encounter scenarios. Comparing the performances of the level-1 and the level-2 pilots, it can be seen that the level-1 pilot maneuvers in a way that he/she provides more separation distance than a level-2 pilot does, except for the case with $90^{\circ}$ approach angle. This, in general, is expected since the level-1 pilot assumes that the intruder is a level-0 Decision Maker (DM) who will continue his/her given path without changing his/her direction, and therefore the level-1 DM takes the responsibility of the conflict resolution himself/herself. On the other hand, the level-2 pilot considers the intruder as a level-1 DM who will make maneuvers to avoid the conflict and therefore the conflict resolution responsibility will be shared. That is why the level-2 pilot, in comparison with the level-1 pilot, avoids the UAS with less separation distance. It is noted, however, that the level-1 pilot deviates from its ideal trajectory significantly more than the level-2 pilot does. Even in the case of $90^{\circ}$ approach angle, where the minimum separation distance between the level-1 pilot and the UAS is slightly less than the case with the level-2 pilot, the trajectory deviation of the level-2 pilot is less than that of the level-1 pilot. These analyses show that the type of the pilot reactions during a conflict scenario makes a significant impact on the results  when evaluating the performances of the SAA algorithms. The same conclusion are derived when the UAS maneuvers based on the SAA2 logic, however these results are omitted to save space. 

\subsection{Hybrid Airspace Scenario with Multiple Encounters} \label{Hybrid Airspace Scenarioo}
The details of this scenario was explained in Section \ref{Hybrid Airspace Scenario}. In this section, the scenario is simulated to investigate a) the effect of the variations in the objective function parameters, b) the effect of the distance and the time horizons and c) the effect of responsibility assignment for conflict resolution, on safety and performance. Since the loss of separation is the most serious issue, the safety metric is taken as the number of separation violations between the UAS and the manned aircraft. Performance metrics, on the other hand, include a) averaged manned aircraft trajectory deviations, b) UAS trajectory deviation and c) total flight time of the UAS. In all of the simulations, level-0, level-1 and level-2 pilot policies are randomly distributed over the manned aircraft in such a way that 10\% of the pilots fly based on level-0 policies, 60\% of the pilots act based on level-1 policies and 30\% use level-2 policies. This distribution is based on experimental results discussed in \cite{Costa:09}. It is noted that although the given distribution is obtained from human experimental studies, they did not necessarily include pilots and therefore may not be fully representative but can easily be adapted to other distributional data for this framework.
\subsubsection{Sensitivity analyses of the weighting parameters in the objective function} \label{Sensitivity}
\begin{figure}[t!]	
	\centering
	\begin{subfigure}[htb]{10cm}
			\centering
			\includegraphics[width=10cm,height=5cm,keepaspectratio]{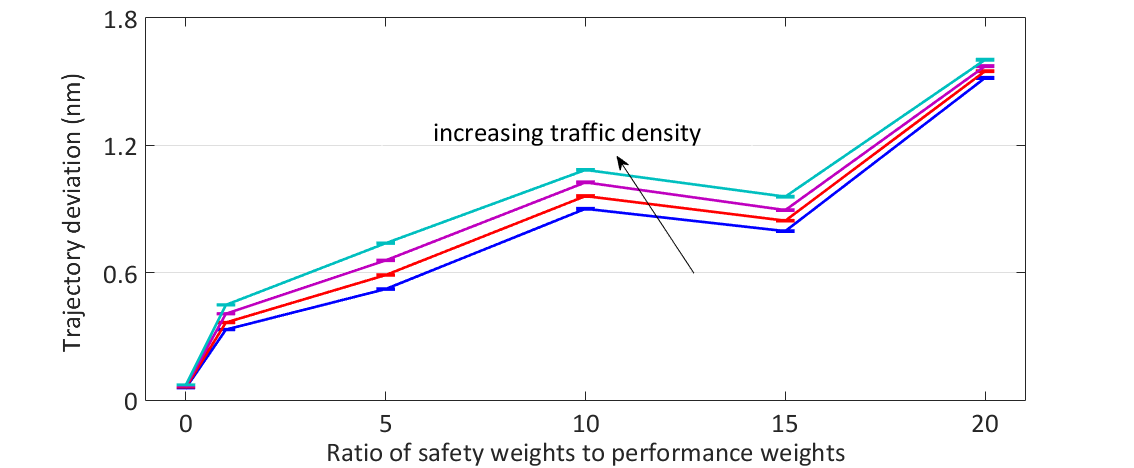}
		\caption{manned aircraft trajectory deviation}\label{f:w1}	
	\end{subfigure}
	\quad 
	\begin{subfigure}[htb]{10cm}
			\centering
			\includegraphics[width=10cm,height=5cm,keepaspectratio]{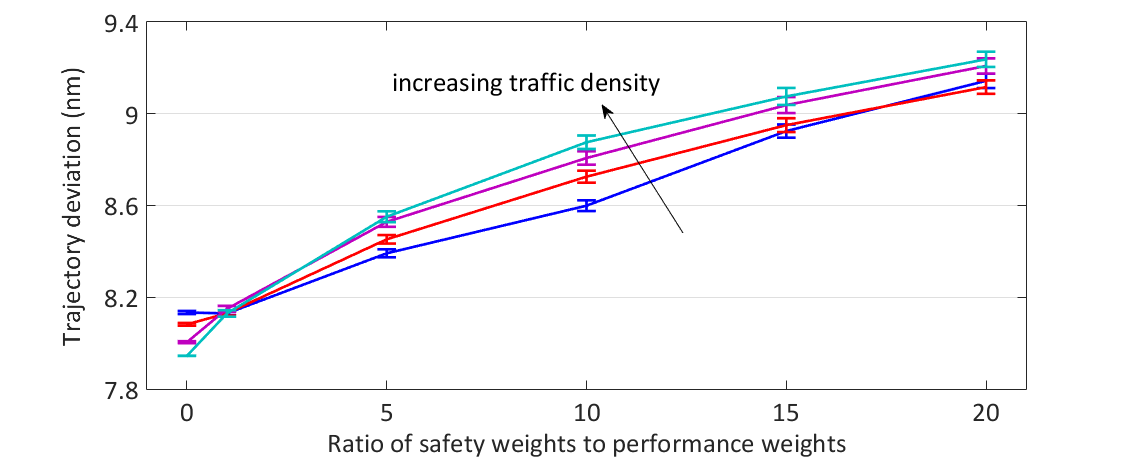}
		\caption{UAS trajectory deviation}\label{f:w2} 
	\end{subfigure}
	\quad 
	\begin{subfigure}[htb]{10cm}
			\centering
			\includegraphics[width=10cm,height=5cm,keepaspectratio]{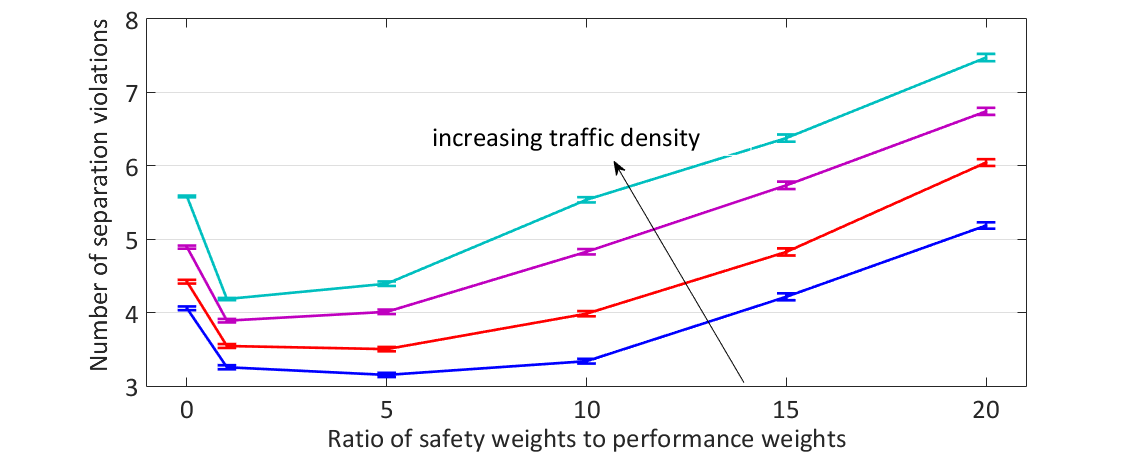}
			\caption{number of separation violations}\label{f:w3}
	\end{subfigure}
	\caption{Pilot model sensitivity analysis.}\label{f:w} 
\end{figure}

In this section, the sensitivity of the pilot model to its parameters, which are the weight vector components of the objective function in equation (\ref{eq:re}), is investigated. Specifically, the effect of the ratio of the sum of the weights of the safety components of the objective function over the sum of the weights of the performance components, $r=\frac{w1+w2+w3}{w4+w5+w6}$, is investigated for various traffic densities. The results of this analysis for various traffic densities in the HAS is depicted in Fig.~\ref{f:w}. It is seen that as r increases, the trajectory deviations of both the manned aircraft and the UAS increases, regardless of the traffic density. Cooperation of the manned aircraft and the UAS to resolve the conflict reduces the number of separation violations up to a certain value of $r$. However, the number of violations starts increasing with further increase in $r$. What this means is that, as pilots become more sensitive about their safety and start to overreact to probable conflicts with extreme deviations from their trajectories, the traffic is effected in a negative way.

\begin{figure}[t!]
	\centering	
	\includegraphics[width=10cm]{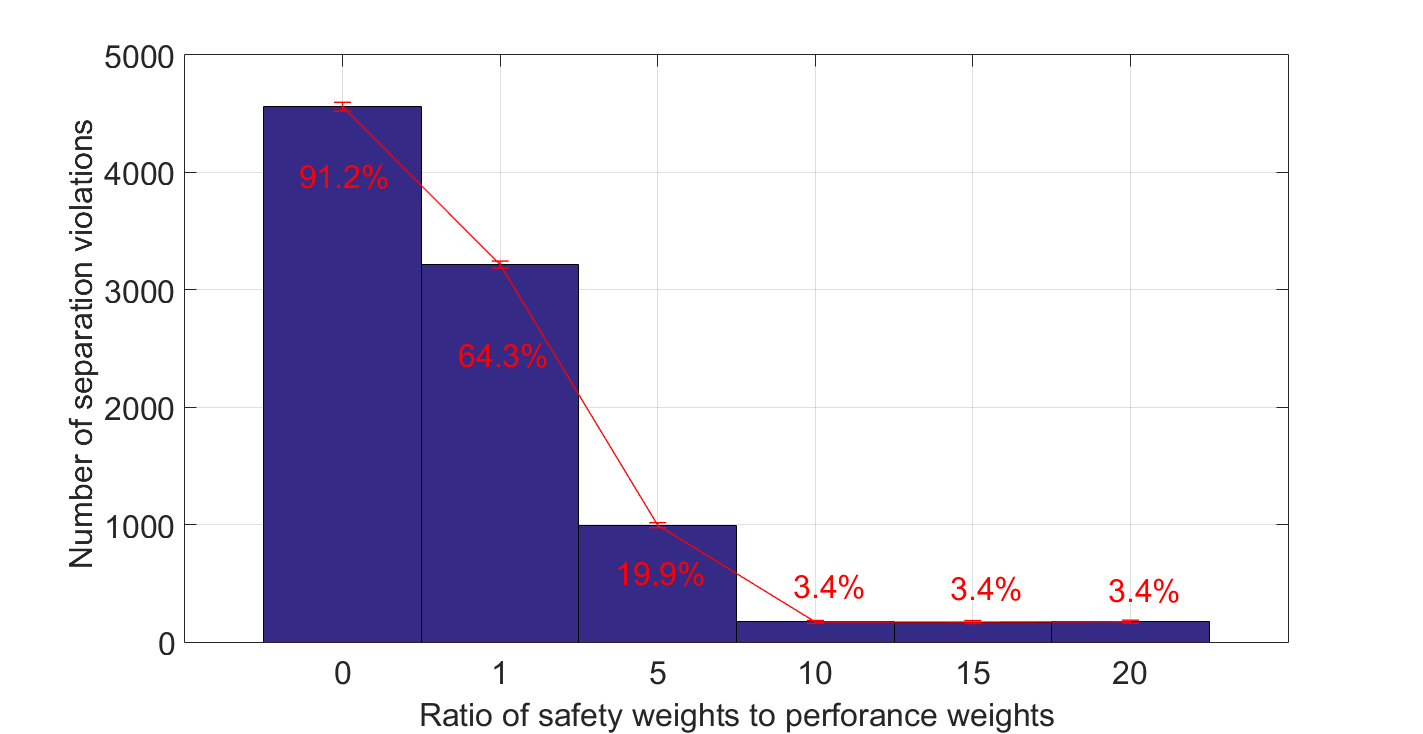}
	\caption{Pilot model sensitivity analysis for the single encounter scenario.}
	\label{f:encN}
\end{figure}

Fig.~\ref{f:encN} presents the effect of increasing the ratio $r$ in a single encounter scenario where surrounding traffic does not exist. The percentage values provided in the figure is obtained for 5000 encounters. For each $r$ value these 5000 encounter ``episodes'' are repeated 1000 times to obtain reliable statistics. As expected, in the absence of surrounding traffic, the increase in the ratio $r$ decreases the number of separation violations. 

\textbf{Remark:} In this study, it is emphasized that humans are not expected to behave optimally in complex situations in the presence of multiple decision makers and automation due to 1) limited observation space, 2) limited processing power and 3) limited information about other decision makers. The exploited modeling framework captures this suboptimal pilot response using several tools explained in earlier sections, including the RL algorithm which provides convergence guarantees to a local maximum. The pilot behavior we observe in Fig.~\ref{f:w} is an example of suboptimal behavior where, in the presence of surrounding traffic, increased safety parameter weights, after a certain point, can cause extreme trajectory deviations and increased separation violations. Fig.~\ref{f:encN}, on the other hand, presents the expected behavior of decreasing violations with increased safety weights, when the scenario is much simpler.

\subsubsection{The effect of distance and time horizons on performance and safety} \label{SAA Comparison}
\begin{figure}[t!]	
	\centering
	\begin{subfigure}[htb]{10cm}
			\centering
			\includegraphics[width=10cm,height=5cm,keepaspectratio]{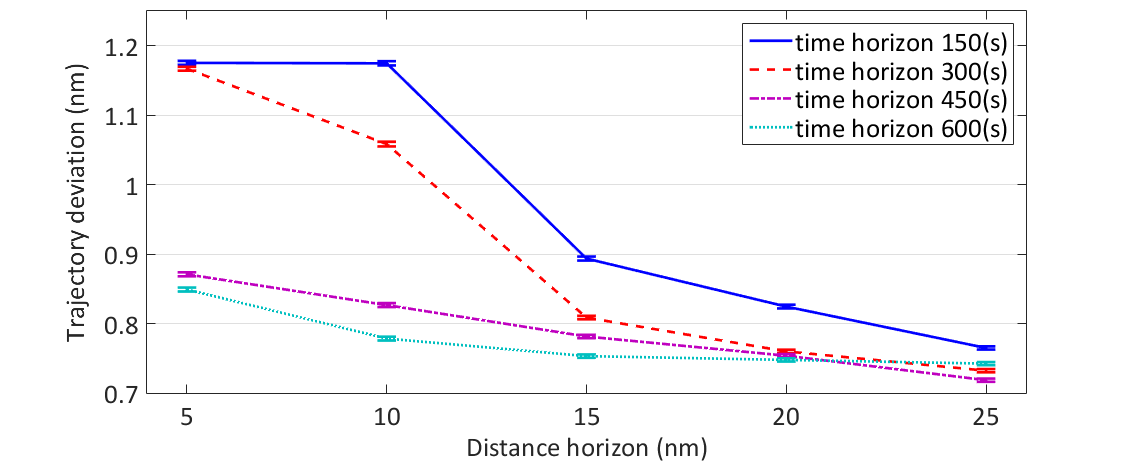}
		\caption{manned aircraft trajectory deviation}\label{f:SAA1_1}		
	\end{subfigure}
	\quad
	\begin{subfigure}[htb]{10cm}
			\centering
			\includegraphics[width=10cm,height=5cm,keepaspectratio]{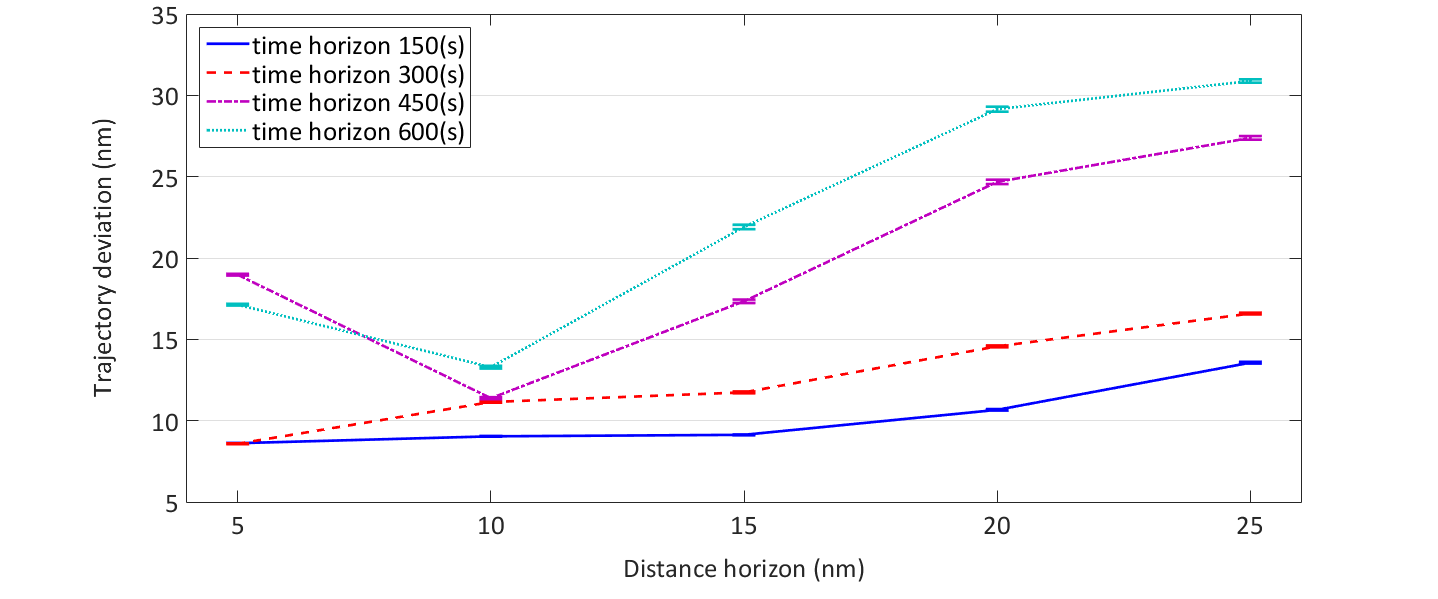}
		\caption{UAS trajectory deviation}\label{f:SAA1_2}
	\end{subfigure}
	\quad
	\begin{subfigure}[htb]{10cm}
			\centering
			\includegraphics[width=10cm,height=5cm,keepaspectratio]{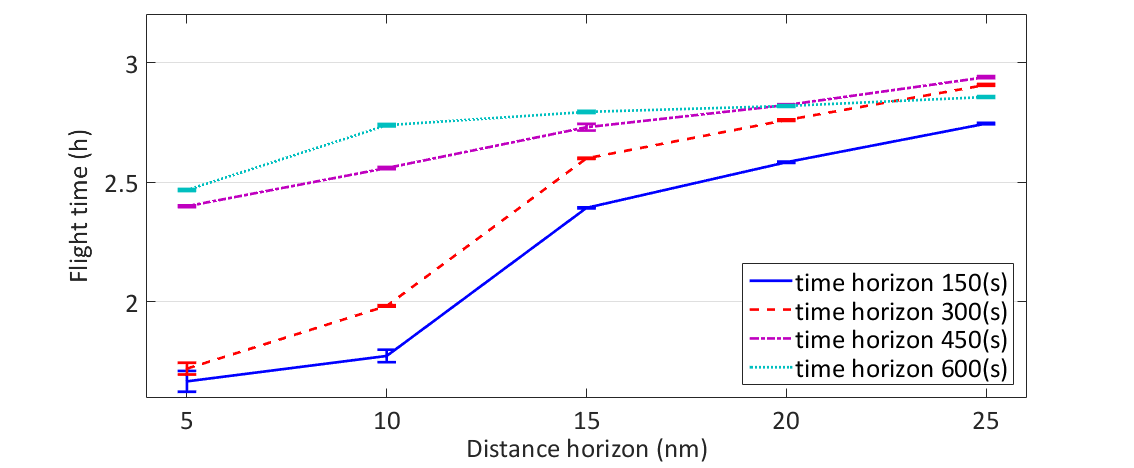}
			\caption{UAS flight time}\label{f:SAA1_3}
	\end{subfigure}
	\quad
	\begin{subfigure}[htb]{10cm}
			\centering
			\includegraphics[width=10cm,height=5cm,keepaspectratio]{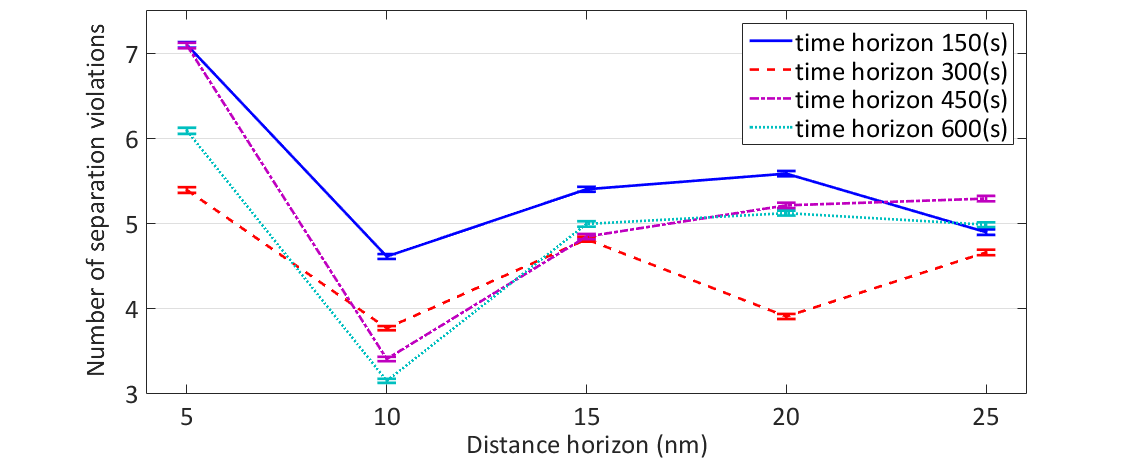}
			\caption{number of separation violations}\label{f:SAA1_4}
	\end{subfigure}
	\caption{Safety vs. performance in HAS, when SAA1 is employed.}\label{f:SAA1}
\end{figure}

\begin{figure}[t!]	
	\centering
	\begin{subfigure}[htb]{10cm}
			\centering
			\includegraphics[width=10cm,height=5cm,keepaspectratio]{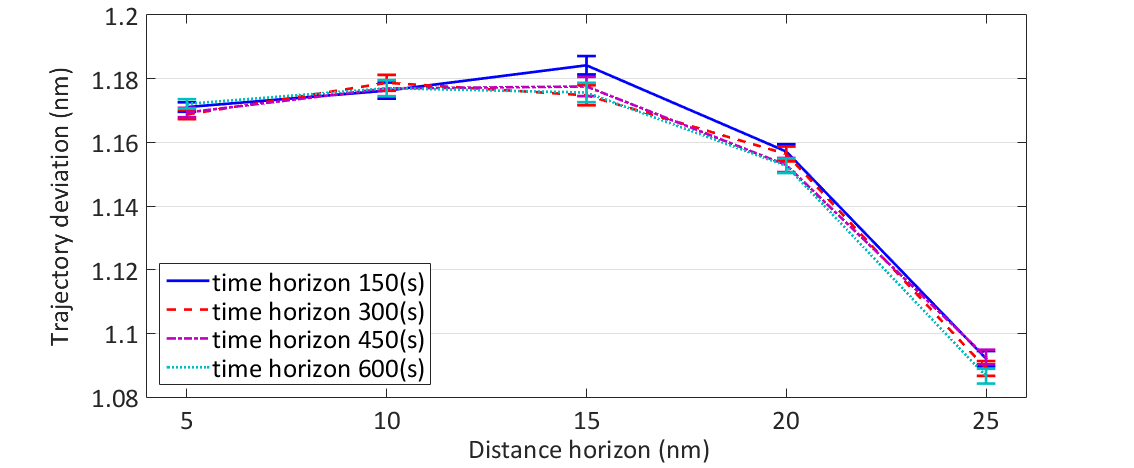}
		\caption{manned aircraft trajectory deviation}\label{f:SAA2_1}		
	\end{subfigure}
	\quad
	\begin{subfigure}[htb]{10cm}
			\centering
			\includegraphics[width=10cm,height=5cm,keepaspectratio]{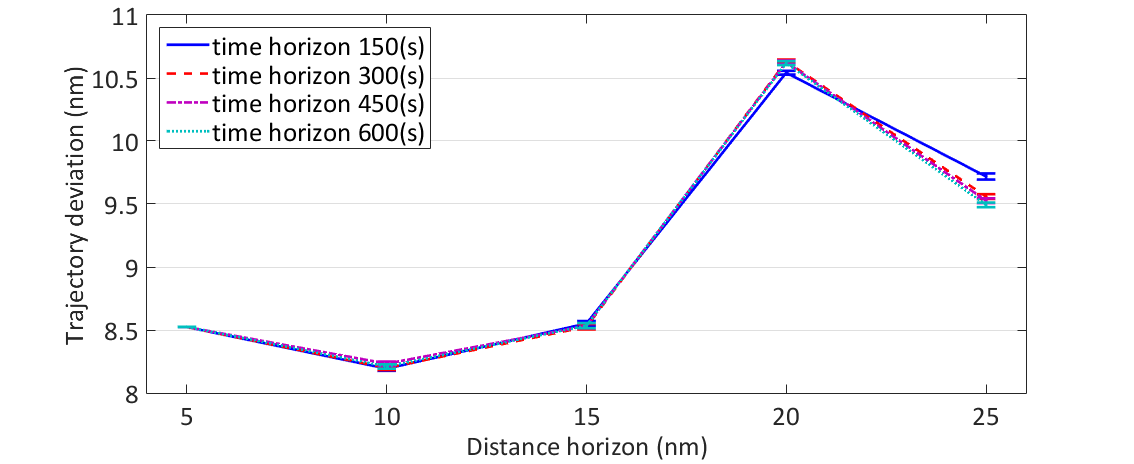}
		\caption{UAS trajectory deviation}\label{f:SAA2_2}
	\end{subfigure}
	\quad
	\begin{subfigure}[htb]{10cm}
			\centering
			\includegraphics[width=10cm,height=5cm,keepaspectratio]{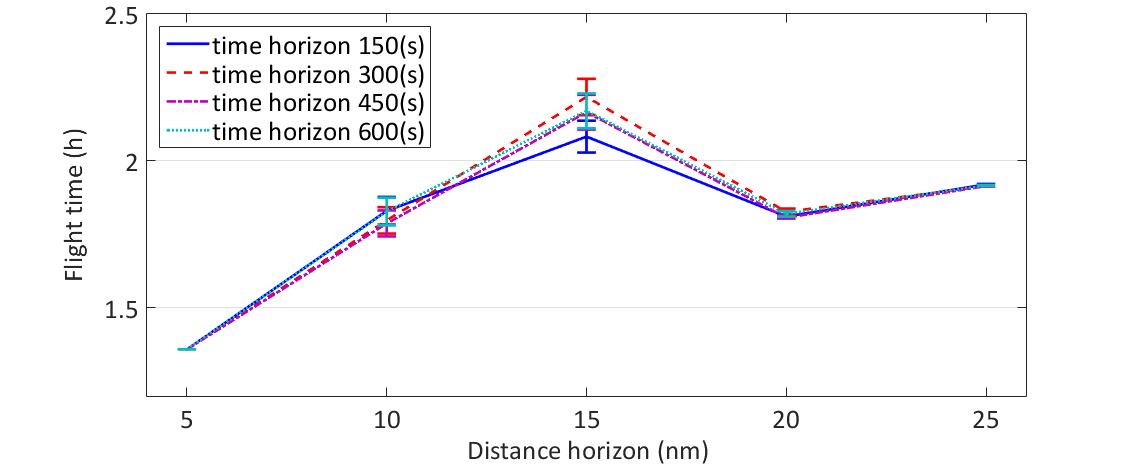}
			\caption{UAS flight time}\label{f:SAA2_3}
	\end{subfigure}
	\quad
	\begin{subfigure}[htb]{10cm}
			\centering
			\includegraphics[width=10cm,height=5cm,keepaspectratio]{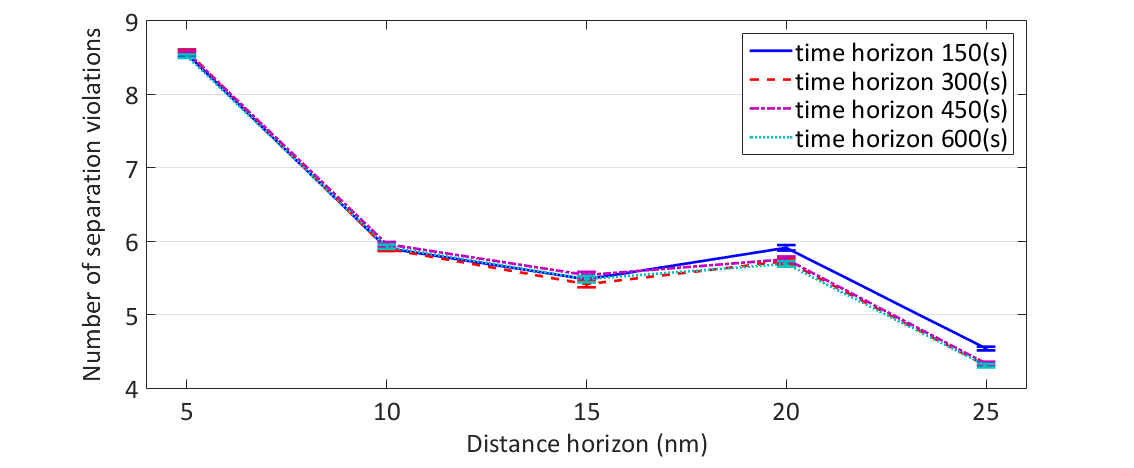}
			\caption{number of separation violations}\label{f:SAA2_4}
	\end{subfigure}
	\caption{Safety vs. performance in HAS, when SAA2 is employed.}\label{f:SAA2}
\end{figure}

Although standard separation distance for manned aviation is $3-5nm$ \cite{Perez:12}, UAS might require wider separation requirements compared to manned aircraft. In the following analysis, horizontal separation requirement for the UAS is called the \textit{distance horizon}, and the effect of it is reflected into the simulation by defining this value as the ``scan radius'' for the SAA algorithm: The SAA algorithm considers an intruding aircraft as a possible threat only if the aircraft is within the scan radius. Another variable whose effect is investigated is defined as the time to separation violation and is called the \textit{time horizon}. In the simulation, the time horizon is used as the time interval, within which the UAS predicts a probable conflict. Figure~\ref{f:SAA1} and Fig.~\ref{f:SAA2} show the effects of the time horizon and the distance horizon on the safety and performance of the system, when SAA1 and SAA2 algorithms are used, respectively. When SAA1 algorithm is employed, it is seen in Fig.~\ref{f:SAA1} that increasing the distance horizon of the UAS makes the SAA1 system detect probable conflicts from a larger distance, which in turn increases the UAS trajectory deviation. Increasing the time horizon makes a similar effect on trajectory deviation. High UAS trajectory deviation results in higher flight times for the UAS to complete its mission. In addition, higher distance and time horizons reduce the trajectory deviations of the manned aircraft since conflicts are resolved mostly by the UAS. When the UAS foresees the probable conflicts earlier, with increased time and distance horizons, the number of separation violations generally decreases. Increasing the distance and time horizons after a certain point does not improve the safety (number of separation violations), since the UAS starts to disturb the traffic unnecessarily due to the overreactions of the SAA system. 

The first observation that strikes the eye, in the case of SAA2 system utilization (see Fig.~\ref{f:SAA2}), is that the time horizon variations do not affect the results as much as the case of SAA1 system utilization. The second important difference of the SAA2 algorithm is that increasing the distance horizon consistently improves the safety (Fig.~\ref{f:SAA2_4}), unlike the case of SAA1 algorithm utilization where larger distance horizon values do not make a major effect on safety. The reason for this difference can be explained by comparing Figs.~\ref{f:SAA1_2} and \ref{f:SAA2_2}: SAA1 causes the UAS deviate from its trajectory significantly more than the SAA2 and thus separation violation numbers for the SAA1 do not improve further after a point due to a significant impact on the surrounding traffic. However, it should be noted that, in general, the violation numbers of the SAA1 is lower than that of SAA2 (see Figs.~\ref{f:SAA1_4}-\ref{f:SAA2_4}). After this quantitative analysis, it can be said that the SAA1 system results in a safer flight (less number of violations) whereas the SAA2 system provides a higher performance flight (lower deviations from the trajectory). 

\begin{figure}[t!]	
	\centering
	\begin{subfigure}[htb]{10cm}
			\centering
			\includegraphics[width=10cm,height=4.49cm,keepaspectratio]{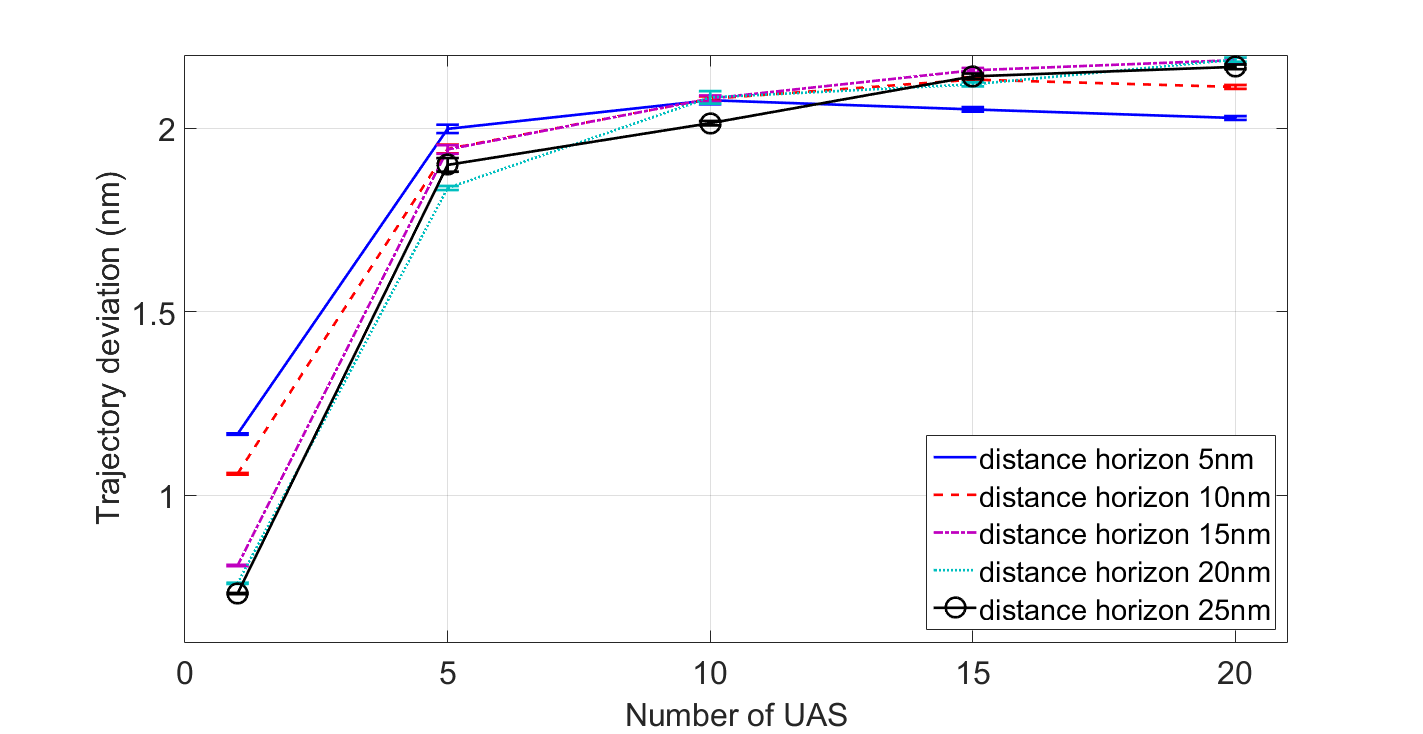}
		\caption{manned aircraft trajectory deviation}\label{f:exUAS1N}		
	\end{subfigure}
	\quad
	\begin{subfigure}[htb]{10cm}
			\centering
			\includegraphics[width=10cm,height=4.49cm,keepaspectratio]{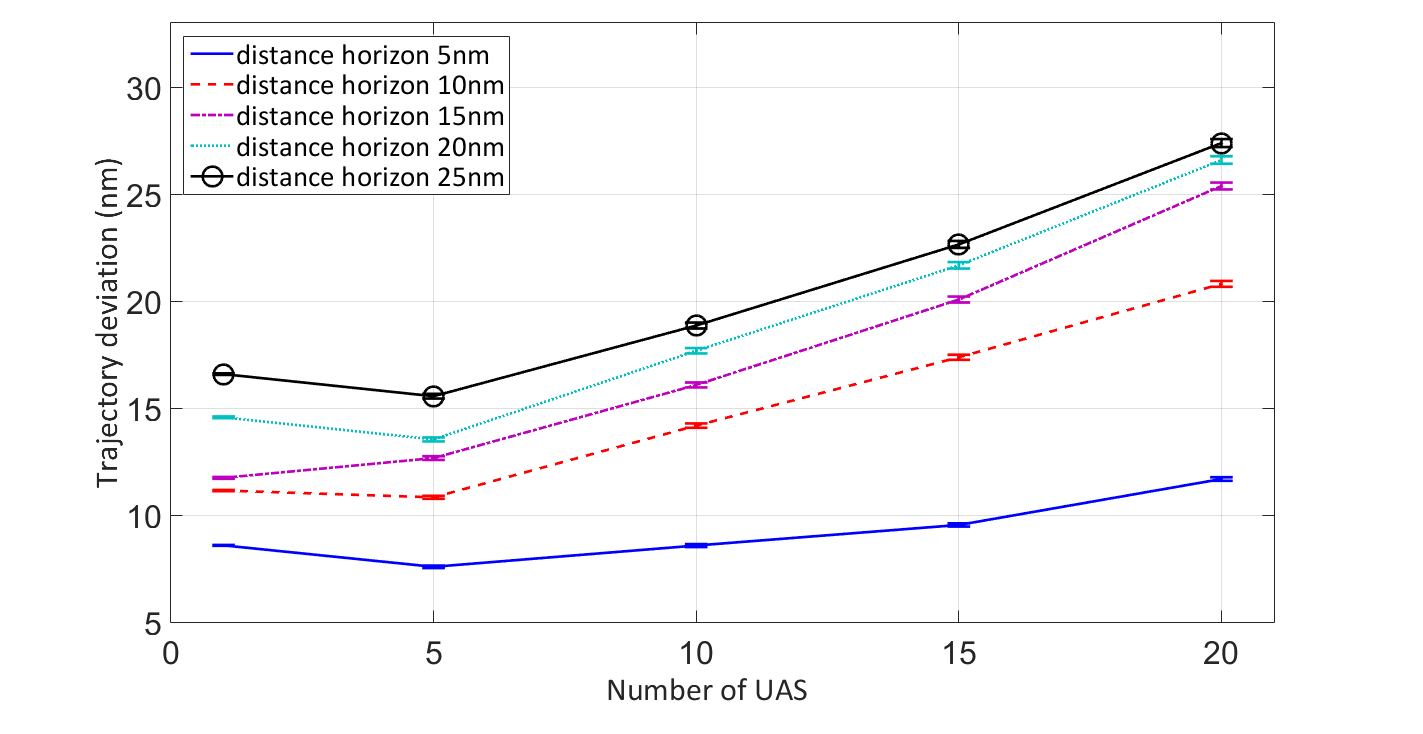}
		\caption{UAS trajectory deviation}\label{f:exUAS2N}
	\end{subfigure}
	\quad
	\begin{subfigure}[htb]{10cm}
			\centering
			\includegraphics[width=10cm,height=4.49cm,keepaspectratio]{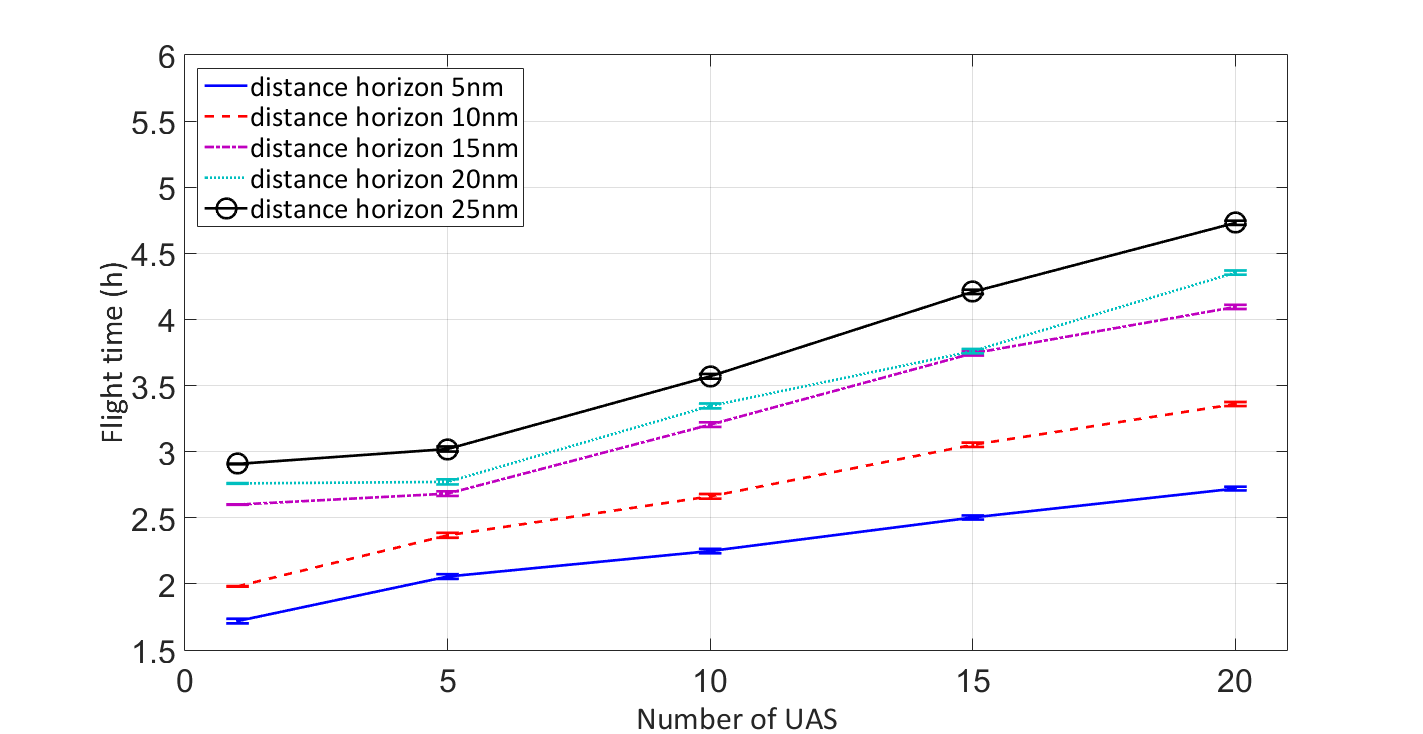}
			\caption{UAS flight time}\label{f:exUAS3N}
	\end{subfigure}
	\quad
	\begin{subfigure}[htb]{10cm}
			\centering
			\includegraphics[width=10cm,height=4.49cm,keepaspectratio]{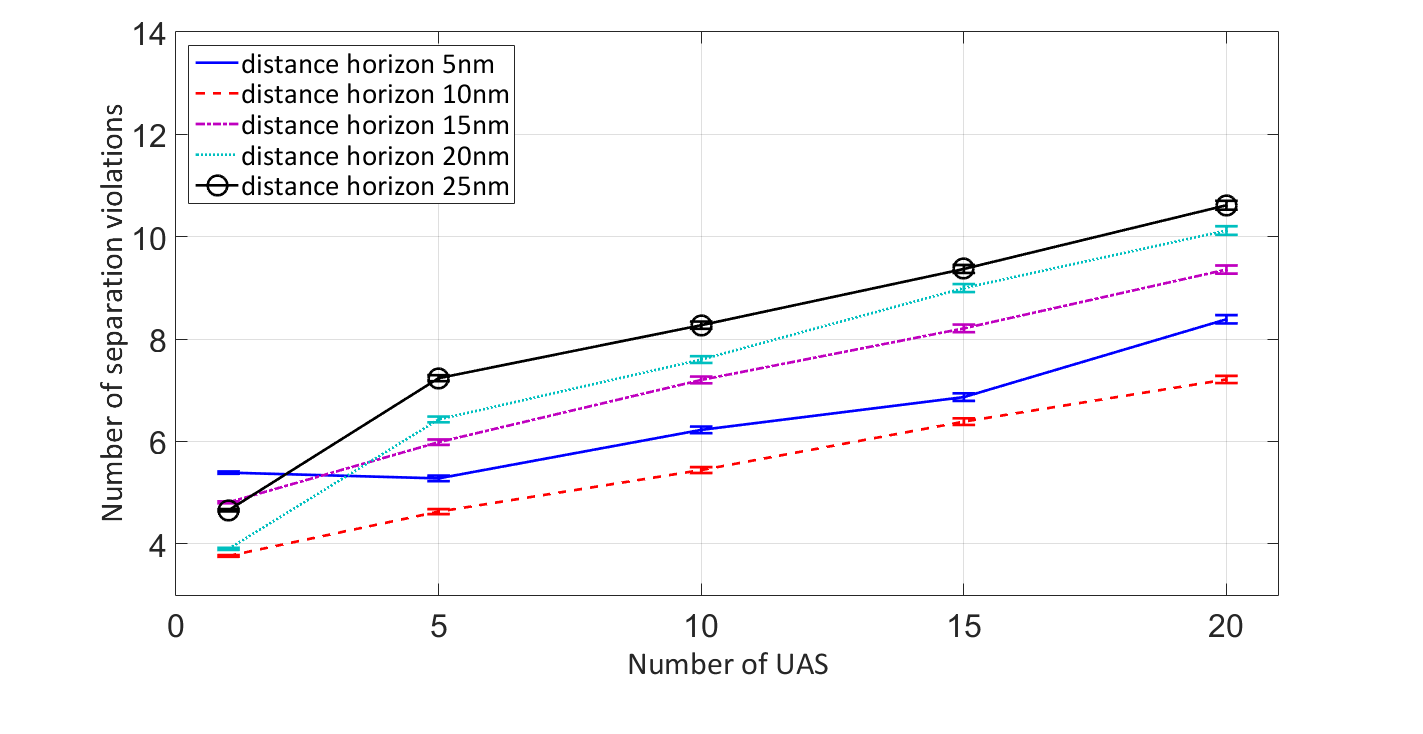}
			\caption{number of separation violations}\label{f:exUAS4N}
	\end{subfigure}
	\caption{Safety vs. performance in HAS, when SAA1 is employed, in a multi-UAS scenario.}\label{f:exUAS}
\end{figure}

It is important to note that when the technologies and procedures mature enough to enable full integration of UAS into the NAS, it would not be unrealistic to expect that the ratio of unmanned to manned aircraft will increase dramatically. Since HAS is a complex system where several intelligent agents move simultaneously, it is impossible to predict the effects of increased UAS presence. Therefore, it is important and useful to investigate the response of the overall system to increased number of UAS. Fig.~\ref{f:exUAS} shows the effect of increasing the number of UAS in HAS. It is seen that, as the number of UAS increases, trajectory deviations, flight times and separation violations increase. It is noted that no mode/phase changes are observed in the system.


\subsubsection{Separation responsibility analysis} \label{separation responsibility analysis}
\begin{figure} [t!]	
	\centering
	\begin{subfigure}[htb]{10cm}
		\centering
		\includegraphics[width=10cm,height=5cm,keepaspectratio]{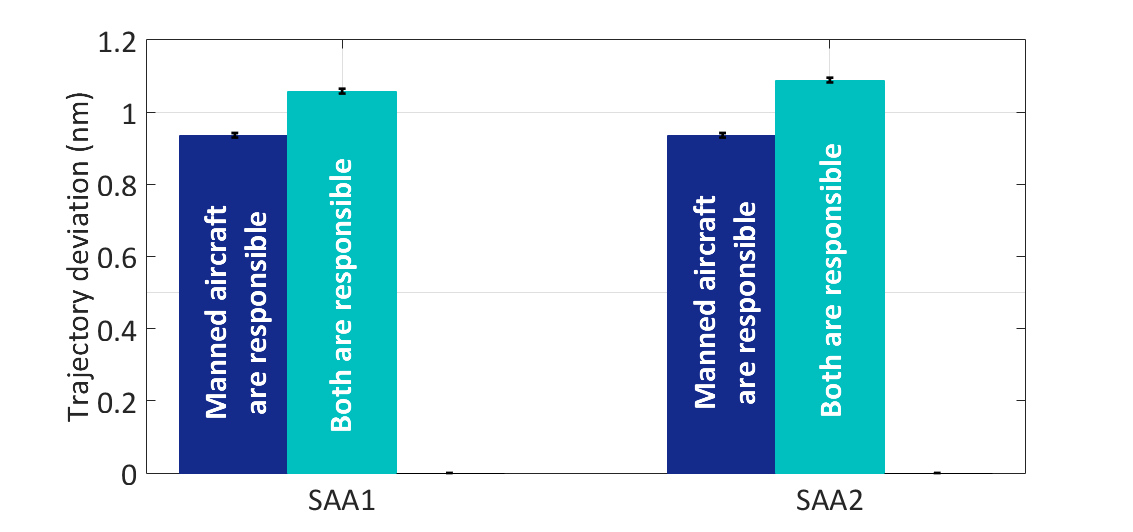}
		\caption{manned aircraft trajectory deviation}\label{f:R1}		
	\end{subfigure}
	\quad
	\begin{subfigure}[htb]{10cm}
		\centering
		\includegraphics[width=10cm,height=5cm,keepaspectratio]{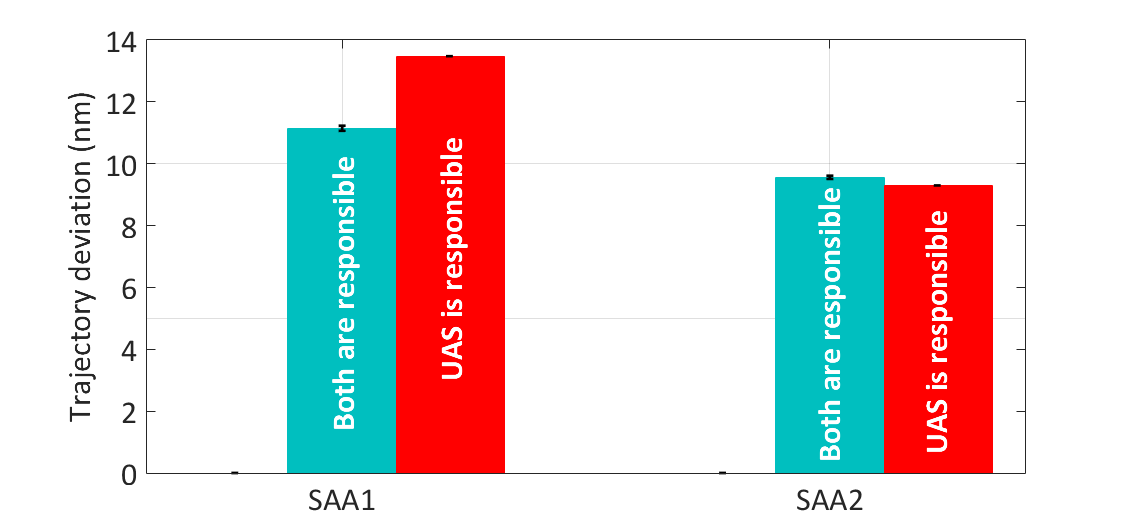}
		\caption{UAS trajectory deviation}\label{f:R2}
	\end{subfigure}
	\quad
	\begin{subfigure}[htb]{10cm}
			\centering
			\includegraphics[width=10cm,height=5cm,keepaspectratio]{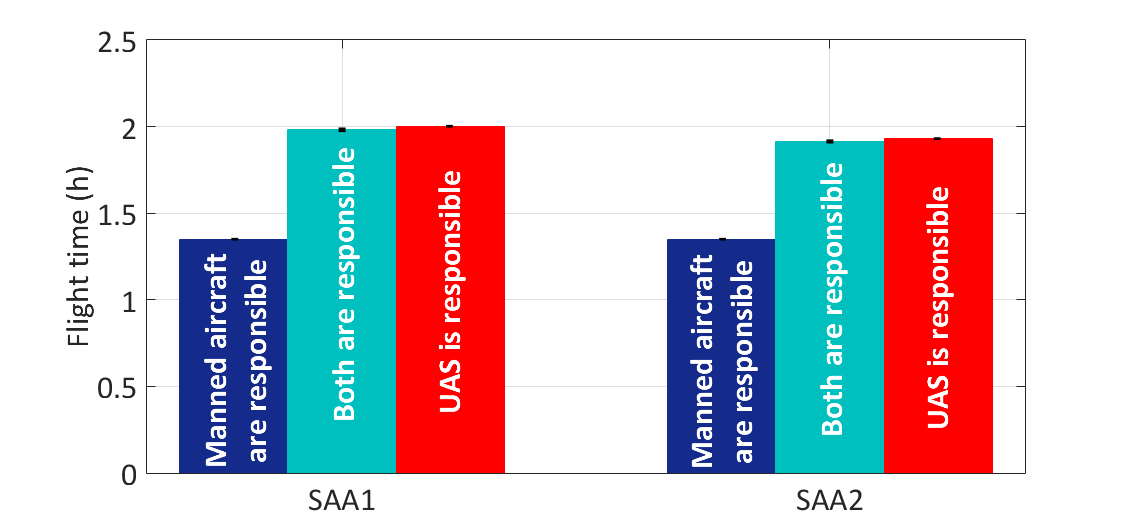}
			\caption{UAS flight time}\label{f:R3}
	\end{subfigure}
	\quad
	\begin{subfigure}[htb]{10cm}
			\centering
			\includegraphics[width=10cm,height=5cm,keepaspectratio]{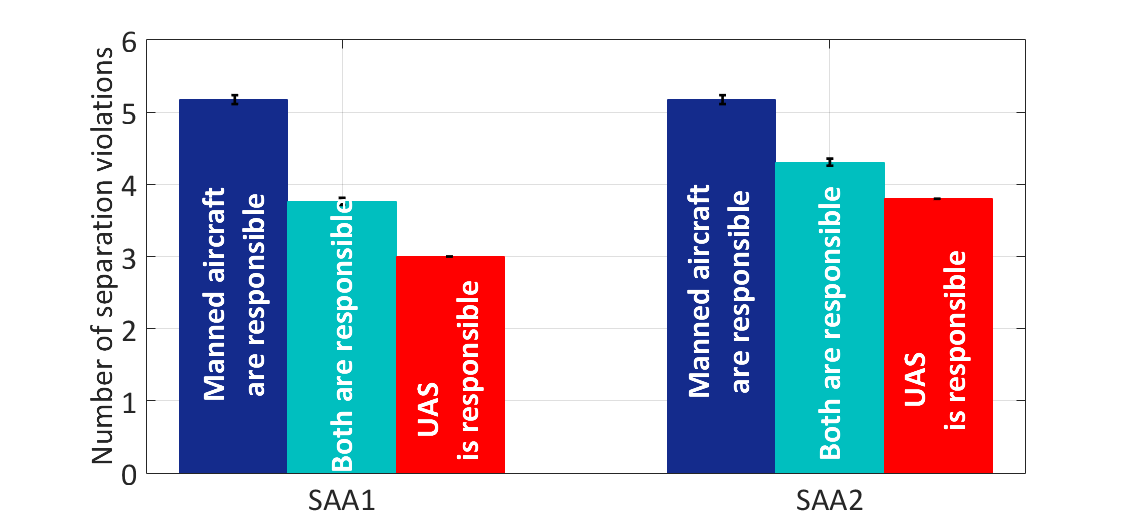}
			\caption{number of separation violations}\label{f:R4}
	\end{subfigure}
	\caption{SAA1, SAA2 and safety vs. performance in HAS.}\label{f:R}
\end{figure}

Another issue to be addressed that is important in studying the integration of UAS into NAS is the separation responsibility \cite{NASA:12}: it is crucial to determine which of the agents (manned aircraft or UAS) will take the responsibility of the conflict resolution. Fig.~\ref{f:R} depicts a comparison of different resolution responsibility cases: manned aircraft are responsible (blue), both manned aircraft and UAS are responsible (cyan) and only the UAS is responsible (red). In the case when only manned aircraft are responsible for conflict resolution, UAS is forced to continue its path without executing the SAA system and manned aircraft act as level-1 and level-2 DMs. In the case when the UAS is responsible for the conflict resolution, the manned aircraft are forced to continue their path without changing their heading and the UAS executes its SAA system. In the case when both the manned aircraft and the UAS are responsible for the conflict resolution, they both execute their evasive maneuvers. Figure~\ref{f:R1} shows that manned aircraft deviate more from their trajectory when both the UAS and the manned aircraft share resolution responsibility, compared to the case when only the manned aircraft are responsible. This is true for both the SAA1 (the results on the left) and the SAA2 (the results on the right) algorithms. The reason for increased trajectory deviation for the manned aircraft in the case of shared responsibility is that the pilots' assumptions about possible UAS actions are not always correct which forces the pilots to make additional adjustments in their trajectory, which in turn increases manned aircraft trajectory deviations. 

On the other hand, Fig.~\ref{f:R2} shows that the UAS deviates from its trajectory more when it is responsible for the resolution, compared to the case when the responsibility is shared, when SAA1 is utilized. For the case of SAA2 utilization, deviations are less and do not change much based on the responsibility assignments. Figure~\ref{f:R3} shows, as expected, that for both SAA1 and SAA2, the UAS flight times are the shortest when only the manned aircraft become responsible for the resolution. Perhaps the most important result is given in Fig.~\ref{f:R4}, where it is shown that for both SAA1 and SAA2 utilizations, the safest case is when the resolution responsibility is given to the UAS.

\chapter{3D Framework}

\section{UAS Integration Scenario}\label{UAS Integration Scenario}
\begin{figure}[htb]
	\centering	
	\includegraphics[width=13.5cm]{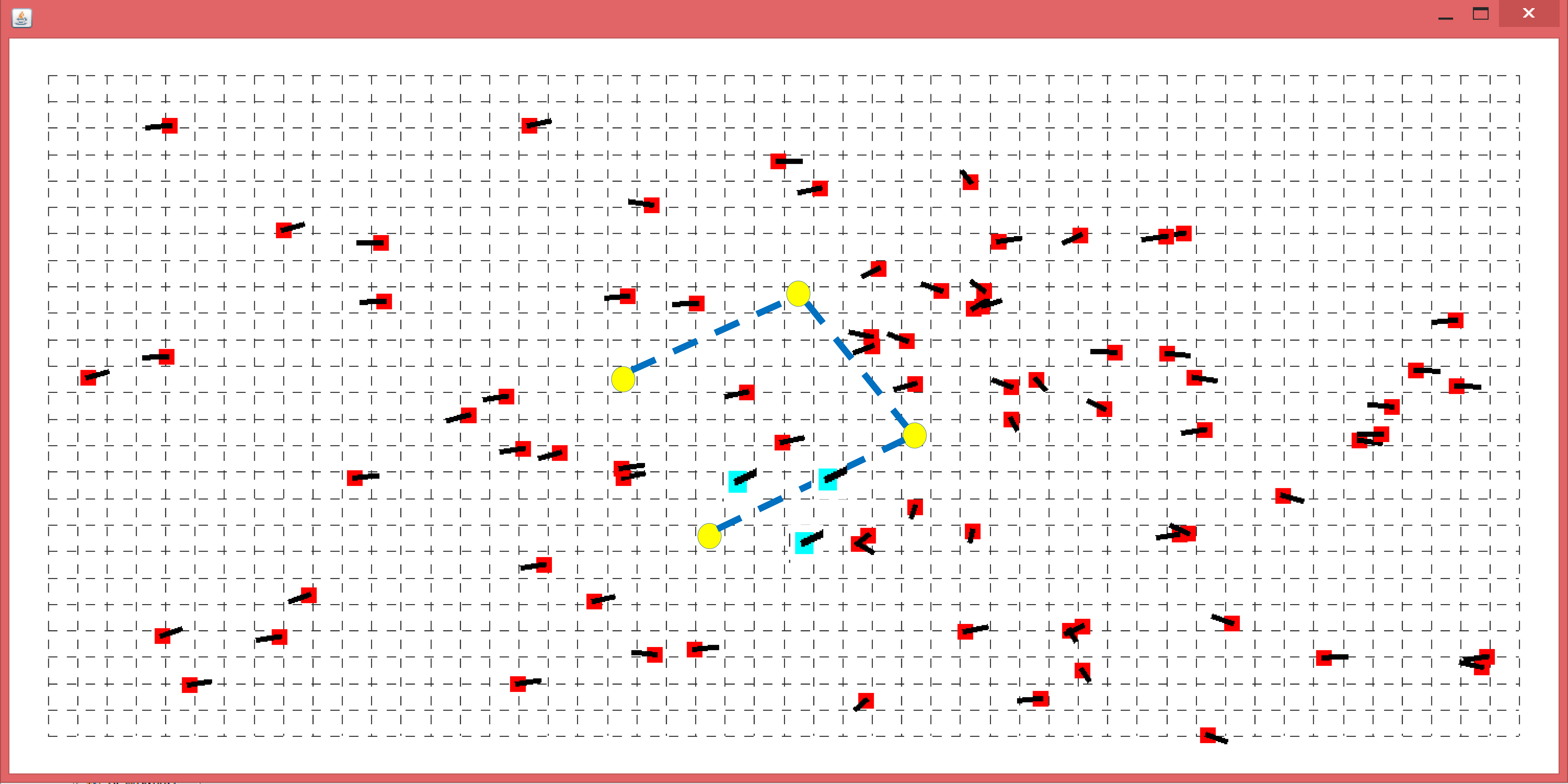}
	\caption{2D snapshot of the airspace scenario in the simulation platform.}
	\label{f:HAS}
\end{figure}

In order to evaluate the outcomes of integrating Unmanned Aircraft Systems (UAS) in to the National Air Space (NAS), a Hybrid Air Space (HAS) scenario, where manned and unmanned aircraft co-exist is designed and explained in this section. The scenario consists of 188 manned aircraft and 3 UAS. The size of the airspace is $600km\times300km\times45000ft$. The initial positions, velocities, headings and altitudes of the aircraft are obtained from Flightradar24 website which provides live air traffic data (\url{http://www.flightradar24.com}). The data is collected from air traffic volume on Colorado state, USA airspace on 11 March, 2015. The manned aircraft in the scenario execute maneuvers based on the pilot model obtained using a combination of reinforcement learning and level-k reasoning, the details of which is explained in the following Section \ref{Pilot Decision Model}. Multiple UAS can be randomly located in the airspace and move based on their pre-programmed flight plan from one waypoint to another. Figure~\ref{f:HAS} shows a snapshot of the scenario with multiple manned aircraft and three UAS moving through their multiple waipoints. All aircraft whether manned or unmanned are flying at different altitudes and this snapshot depicts a 2D projection of their configuration , on the horizontal plane. The red squares correspond to manned aircraft and the cyan squares correspond to UAS, which are flying at different altitudes. All aircraft, manned or unmanned have continuous dynamics, which are provided in the following sections. Yellow circles show the predetermined waypoints that the UAS with the highest altitude is required to pass. The waypoints of other two UAS are not shown in this snapshot. The dashed blue lines passing through the waypoints show the predetermined path of the UAS. It is noted that the UAS do not follow this path exactly since they need to deviate from their original trajectory to avoid possible conflicts using an on-board Sense and Avoid (SAA) algorithm, which is obtained from \cite{Fasano:08} and \cite{Mujumdar:11}.

In the scenario, it is assumed that each aircraft is able to receive the surrounding traffic information using Automatic Dependent Surveillance-Broadcast (ADS-B) technology. ADS-B technology can provide an own-ship aircraft the identification, position and velocity information of surrounding aircraft that are also equipped with ADS-B.

\subsection{UAS Conflict Detection And Avoidance Logic}\label{UAS Conflict Detection And Avoidance Logic}
UAS fly according to their pre-programmed flight plans marked by the yellow shown in the Fig.~\ref{f:HAS}. UAS are assumed to have the dynamics of RQ-4 Global Hawk with operation speeds of $340knots$, respectively \cite{Dalamagkidis:09}. UAS are also equipped with SAA systems which enable them to detect trajectory conflicts and to initiate evasive maneuvers, if necessary. If no conflict is detected, the UAS continues to follow its mission plan. Either receiving a conflict resolution command from the SAA system or flying based on their pre-defined flight plan, UAS always receives a velocity command during the flight. The velocity vector variation dynamics of the UAS is modeled by a first order dynamics with a time constant of $1s$ \cite{Mujumdar:11} which is represented as:
\begin{equation}
\dot{\vec{v}}= -(\vec{v}-\vec{v}_{d}),
\end{equation}
where $\vec{v}$ and $\vec{v}_{d}$ are the current and the desired/commanded velocity vectors, respectively. The two SAA logic that are utilized in this study are developed by Fasano \textit{et al.} \cite{Fasano:08}, which is referred as SAA1, and Mujumdar \textit{et al.} \cite{Mujumdar:11}, which is referred as SAA2. Both of the SAA logics are consist of two phases; a \textit{conflict detection} phase and a \textit{conflict resolution} phase. The conflict detection phase is the same for both SAA1 and SAA2. A conflict is detected if the minimum distance between the UAS and the intruder aircraft is calculated to be less than a minimum required distance, $R$, during a predefined time interval. The minimum distance is calculated by projecting the trajectories of the UAS and the intruder aircraft in time. Once the conflict is detected, SAA1 and SAA2 suggest their own velocity adjustment commands in order to resolve the conflict. The velocity adjustment command of the SAA1 and SAA2 logics, $\vec{v_{A}^{d1}}$ and  $\vec{v_{A}^{d2}}$ are given in the equations below
\begin{equation}
\label{e:function}
\vec{v}_{A}^{d1} =
\left[ \frac{v_{AB}\cos(\eta-\zeta)}{\sin(\zeta)}[\sin(\eta)\frac{\vec{v}_{AB}}{v_{AB}}-\sin(\eta-\zeta)\frac{\vec{r}}{\|\vec{r}\|}] \right]+\vec{v}_{B}
\end{equation}

\begin{equation}
\label{e:function}
\vec{v}_{A}^{d2} =
\frac{-\vec{v}_{A}(\frac{\vec{r}_{0}.\vec{v}_{AB}}{\|\vec{v}_{AB}\|})-(R-\|\vec{r}_{m}\|)\frac{\vec{r}_{m}}{\|\vec{r}_{m}\|}}{\|-\vec{v}_{A}(\frac{\vec{r}_{0}.\vec{v}_{AB}}{\|\vec{v}_{AB}\|})-(R-\|\vec{r}_{m}\|)\frac{\vec{r}_{m}}{\|\vec{r}_{m}\|}\|}
\end{equation}
where, $\vec{v_{A}}$ and $\vec{v_{B}}$ refer to the velocity vectors of the UAS and the intruder. $\vec{r}$ and $\vec{v_{AB}}$ denote the relative position and velocity between the UAS and the intruder, respectively. $\zeta$ is the angle between $\vec{r}$ and $\vec{v_{AB}}$ and $\eta$ is calculated as $\eta=\sin^{-1}\frac{R}{\|\vec{r}\|}$. $\vec{r}_{0}$ refers to the initial relative position vector between the UAS and the intruder. In the case of multiple conflict detection, the UAS will start an evasive maneuver to resolve the conflict that is predicted to happen earliest. The velocity adjustment suggested by the SAA1 logic guarantees minimum deviation from the trajectory, while in the case of the SAA2 logic, UAS moves to resolve the conflict until it retains the minimum safe distance with the intruder.  

\subsection{Manned Aircraft}\label{Manned Aircraft}
All manned aircraft are assumed to be in their en-route phase of travel with constant speed, $v$, in the range of $[150-550] knots$. Pilots may decide to change the heading angle for $\pm45^{\circ}$, or change the pitch angle for $\pm10^{\circ}$, or may decide to keep both the heading and pitch angles unchanged. Once the pilot gives a heading or pitch command, the aircraft moves to the desired heading and pitch, $\psi_{d}$ and $\theta_{d}$, in the constant speed mode where, the heading and pitch change is modeled by a first order dynamics with the standard rate turn: a turn in which an aircraft changes its heading at a rate of $3^{\circ}$ per second ($360^{\circ}$ in 2 minutes) \cite{pilot-handbook:08}. This is modeled as a first order dynamics with a time constant of $10s$ ($45 \times (1-1/e)/3 \approx 10$). Therefore, the aircraft heading and pitch angle dynamics can be given as
\begin{equation}
\dot{\psi}= -\frac{1}{10}\times(\psi-\psi_{d})
\end{equation}
\begin{equation}
\dot{\theta}= -\frac{1}{10}\times(\theta-\theta_{d})
\end{equation}
and the velocity, $\vec{v}=(v_{x},v_{y},v_{z})$, is then obtained as:
\begin{equation}
\\v_{x}= \|\vec{v}\|\sin\psi\cos\theta.
\end{equation}
\begin{equation}
\\v_{y}= \|\vec{v}\|\cos\psi\cos\theta.
\end{equation}
\begin{equation}
\\v_{z}= \|\vec{v}\|\sin\theta.
\end{equation}

\section{Pilot Decision Model}\label{Pilot Decision Model}
The proposed pilot decision making process model in this study is formed by combining of two methodology: the dynamic level-k reasoning and the neural fitted Q iteration (NFQ), which is an approximate reinforcement. A level-k-type model is trained by assigning level-k-1-type behavior to all of the agents (manned aircraft) except the one that is being trained. The trainee learns to react as best as he/she can in this environment using NFQ. Thus, the resulting behavior becomes a level-k type. This process starts with training a level-1 type behavior and continues until the highest desired level is reached. Once all of the desired levels are obtained, the training stage ends and, in the simulation stage, the obtained level-k reaction models are then are used in a airspace scenario explained in Section \ref{Pilot Decision Model} where both manned aircraft and UAS co-exist. In the simulation certain proportions of level-0, level-1 and level-2 behavior type are assigned to the manned aircraft. It is noted that each of level-1 and level-2 agents can change their level-k behavior type based on Dynamic level-k reasoning method after observing their intruder's behavior.

\subsection{Dynamic Level-k Reasoning}\label{Dynamic Level-k Reasoning}
Level-k reasoning is a game theoretical model where the main idea is that humans have various levels of reasoning in their decision-making process \cite{Chong:16}. It has been observed that reasoning levels are related to the cognitive abilities of humans \cite{Gill:16}. The level hierarchy is iteratively defined such that the level-k rule is a best response to the level-(k-1) rule.  A level-1 decision maker (DM), for example, assumes that the other agents in the scenario are level-0 and takes actions accordingly to provide the best response. A level-2 DM takes actions to give the best response to other DMs that have level-1 reasoning and so on. From a modeling standpoint, the level-0 rule represents an initial point from which more sophisticated rules can be obtained iteratively. A Level-0 here represents a ``nonstrategic" DM who does not take into account other DMs, possible moves when choosing his/her own actions. This behavior can also be considered as ``reflexive" because it only reacts to the immediate observations. In this study, a level-0 pilot flies an aircraft with constant heading and pitch angles starting from its initial position toward its destination.

\begin{figure}[htb]
	\centering	
	\includegraphics[width=8cm]{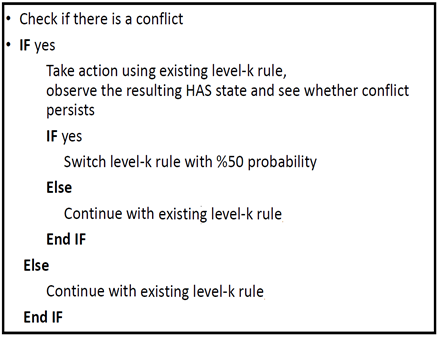}
	\caption{Dynamic level-k reasoning pseudo-code.}
	\label{f:PCN}
\end{figure}

In its conventional form, level-k reasoning help model the interactions between the DMs where a level-k DM assumes that the interacted DM has level-(k-1) reasoning. Although this approach proved to be successful in predicting  short term or one-shot interactions, it misses the point that agents, during their interactions, may update their assumptions about the other agents and therefore update their own behavior. To remedy this problem, we introduce a closed loop algorithm which allows the agents to dynamically update their reasoning levels if a conflict is detected. This algorithm is explained in Fig.~\ref{f:PCN}, where a pseudo-code is provided. It is noted that in the simulations, a simpler version of the algorithm provided in Fig.~\ref{f:PCN} is used, where one of the agents are randomly selected to switch their level, instead of giving each agent a $\%50$ chance for level switching.

\subsection{Neural Fitted Q Iteration}\label{Neural Fitted Q Iteration}
Reinforcement learning is a mathematical learning method based on reward and punishment \cite{Marco:12}. The agent interacts with an environment through its observations, actions and a scalar reward signal received from the environment. The agent's aim is to select actions that maximizes the cumulative future reward. Given a state, when an action increases (decreases) the value of an objective function (reward), which defines the goals of the agent, the probability of taking that action increases (decreases). Reinforcement learning algorithms involve estimating the state-action value function, or ``the Q value", which is a measure of how valuable, in terms of maximizing the total accumulated rewards, taking an action is, given the agent's state. This estimation is generally conducted in an iterative fashion by updating these Q values after each training step. In classical Q-learning reinforcement learning algorithm, for discrete state and action spaces, the update rule is given by \cite{Rie:05} and \cite{Rie:11}: 
\begin{equation}
Q(s,a)\rightarrow(1-\alpha)Q(s,a)+\alpha(r(s,a,s')+\gamma\max\limits_{a}Q(s',a))
\end{equation}
where, $s$, $a$, and $s'$ refer to the state, action, and the successor state, respectively. $\alpha$ is a learning rate $\gamma$ is a discounting factor. In the case of continuous state space and discrete action space, Neural Fitted Q Iteration (NFQ) method \cite{Rie:05} and \cite{Rie:11} approximates the Q value function using a neural network of type multi-layer perceptron. For a given state-action pair the neural network takes the state and action as its input and provides an approximate value of the corresponding Q value for the state-action pair. The method's aim is to minimize the following error function (see \cite{Rie:05}):

\begin{equation}
(Q(s,a)-(r(s,a,s')+\gamma\max\limits_{a}Q(s',a)))^{2}.
\end{equation}
where, $r(.)$ is the reward signal that agent receives from the environment after the transition from state, $s$ to state, $s'$ by taking action $a$. This error function measures the deviations between state-action Q values approximated by the multi-layer perceptron ($Q(s,a)$) and the target value ($r(s,a,s')+\gamma\max\limits_{a}Q(s',a))$). In the NFQ method, Q value functions are updated in a batches which means that the entire set of input patterns ($(s_{i},a_{i}), i=0,1,2,3...$) and target patterns ($r(s_{i},a_{i},s_{i}')+\gamma\max\limits_{a}Q(s_{i}',a_{i})), i=0,1,2,3...$) are collected and the update is performed at the end of a full episode, where an episode is defined as .... Concisely, the NFQ consists of two major steps: the generation of a training set and training a multi-layer perceptron using this set to obtain a Q-value function approximating the optimal state-action Q-values, at the end of each episode. The training is stopped whenever the received average reward per episode converges.

The goal of the reinforcement learning algorithm is to learn the optimal Q values by maximizing the agent's return, which is calculated via a reward/objective function. A reward function can be considered as a happiness function, goal function or utility function which represents, mathematically, the preferences of the pilot. In this paper, the pilot reward function is defined as
\begin{multline} 
reward = w1*(-C)+w2*(-S)+w3*(-A)+w4*(-P)
\end{multline}
In (11) $``C"$ is the number of aircraft within the collision region. Based on the definition provided by the Federal Aviation Administration (FAA), the radius of collision is taken as $500ft$ in the horizontal direction and $100ft$ in the vertical direction \cite{NextGen:07}. $``S"$ is the number of air vehicles within the separation region. The radius of the separation region is $5nm$ in the horizontal direction \cite{Perez:12} and $1000ft$ in the vertical direction based on the ``Reduced vertical separation minima" \cite{NextGen:07}. $``A"$ represents whether the aircraft is getting closer to the intruder or going away from the intruder in terms of their approach angle and takes the values of 1, for getting closer, or 0, for going away. $``P"$ represents whether the aircraft gets closer to or goes away from its trajectory vector in terms of angle and takes the values of 0, for getting closer, or 1, for going away. 

\begin{figure}[t!]
	\centering	
	\includegraphics[width=8cm]{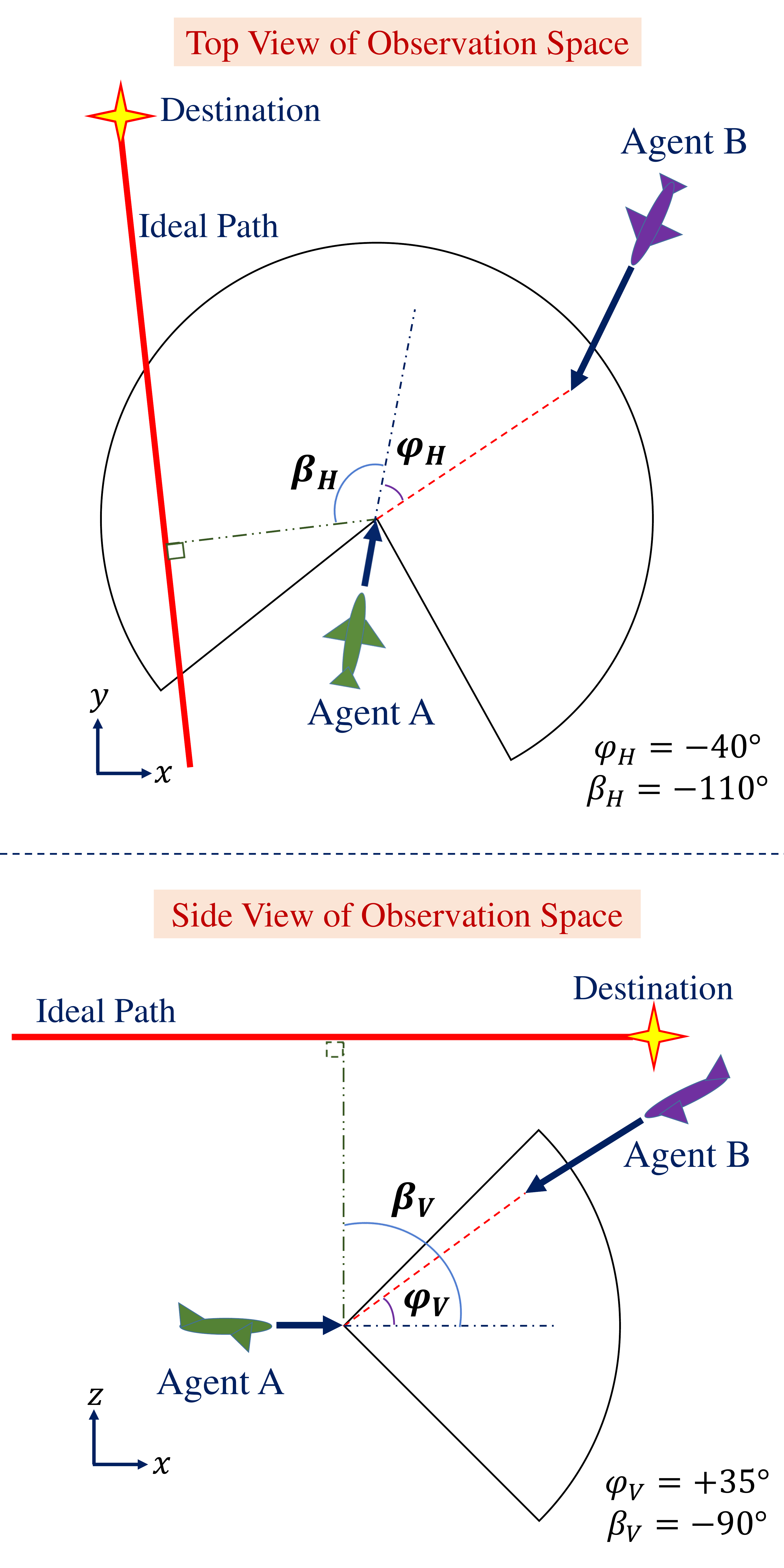}
	\caption{Pilot observation space.}
	\label{f:ObservationSpace}
\end{figure}

\begin{figure}[t!]
	\centering	
	\includegraphics[width=8cm]{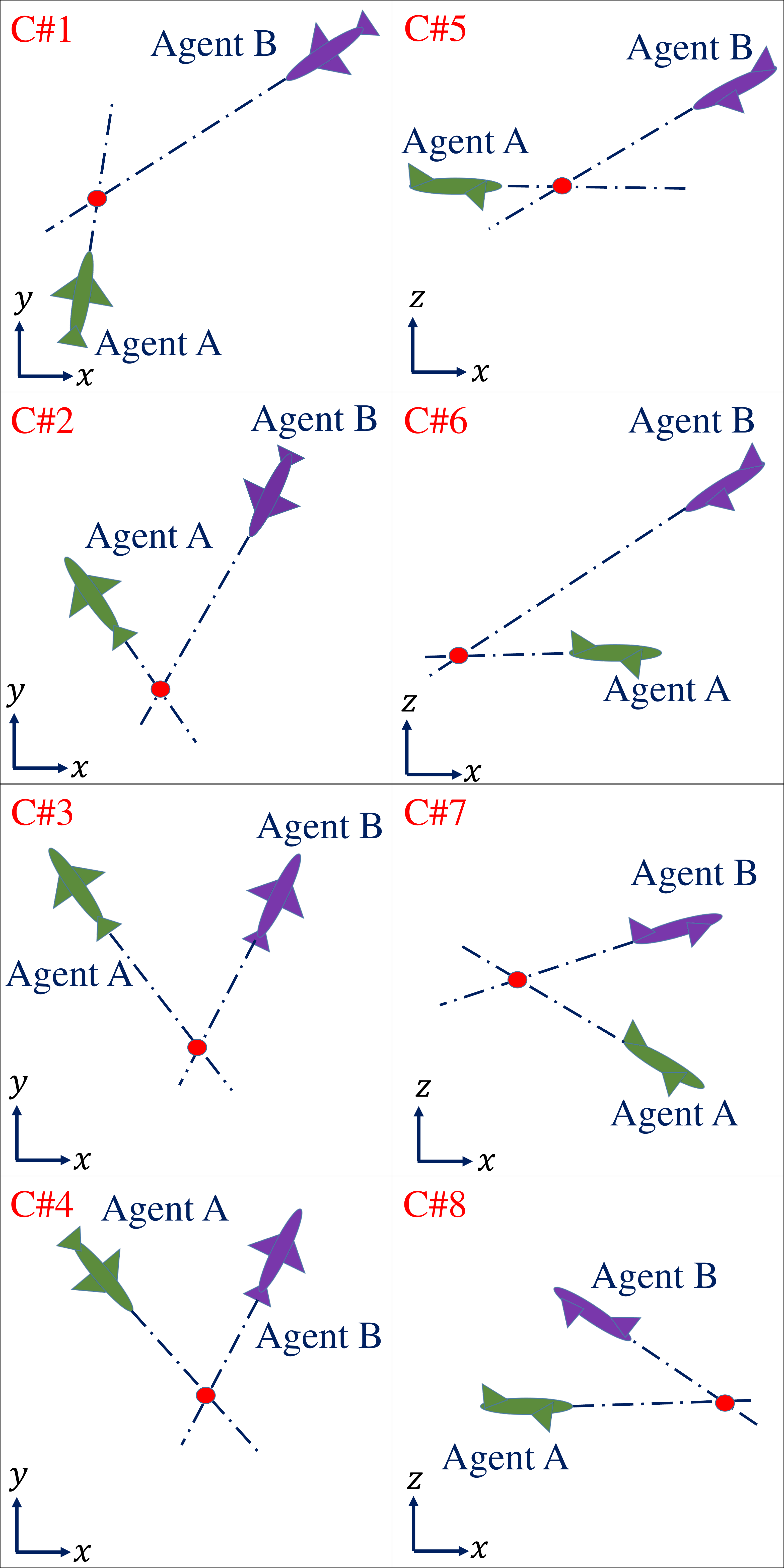}
	\caption{Encounter categories.}
	\label{f:EC}
\end{figure}

Although ADS-B provides the positions and the velocities of other aircraft, with his/her limited cognitive capabilities a pilot can not possibly process all this information during his/her decision making process. In this study, in order to model pilot limitations, including the limitations at visual acuity and perception depth, as well as the limited viewing range of an aircraft, it is assumed that the pilots can observe (or process) the information from a limited portion of the nearby airspace. This limited portion is called the ``observation space". Since the the aircraft are moving in a 3D region, the observation space is a 3D portion of the nearby airspace. This observation space is considered as a portion of a sphere centered at the location of the pilot. In order to illustrate the observation space, it is divided into horizontal and vertical parts and is schematically depicted in Fig.~\ref{f:ObservationSpace}. Since viewing range of a pilot may be different in horizontal and vertical directions, the observation space in these two directions are shown as different angular portions of a circle. Since the standard separation for manned aviation is $3-5nm$ \cite{Perez:12}, the  radius of the observation space is taken as a variable larger than the $5nm$. Whenever an intruder aircraft moves toward the observation space (see Fig.~\ref{f:ObservationSpace}, where Agent B is the intruder), the approach geometry is defined by two angles: $\phi_{H}$, in the horizontal plane, and $\phi_{V}$, in the vertical plane. The aircraft's angular orientation with respect to his/her ideal trajectory is also defined by two angles: $\beta_{H}$, in the horizontal plane, and $\beta_{V}$, in the vertical plane. Fig.~\ref{f:ObservationSpace} depicts a typical example, where the aircraft B is moving toward the observation space with $\phi_{H}=-40^{\circ}$, $\phi_{V}=-40^{\circ}$, $\beta_{H}=-110^{\circ}$ and $\beta_{V}-90^{\circ}$. Aircraft relative orientations are also coded as different ``encounter types". Figure~\ref{f:EC} depicts 8 types of encounter geometries projected in the horizontal plane (left column) and four types of encounter projected in the vertical plane (right column). These geometries are indicated as $C\#i, i=1, 2, ..., 8$ in the figure. Finally, the observation space includes the pilot's memory of what their actions were at the previous time step. Given an observation, the pilots can choose between five actions: turn $45^{\circ}$ left, go straight, turn $45^{\circ}$ right, pitch $10^{\circ}$ up, or pitch $10^{\circ}$ down. It is noted that these pilot commands are filtered through the aircraft dynamics provided in Section~\ref{Manned Aircraft}. 


The information of the observations and actions of the pilots fed to the neural network which is in charge of approximating the Q values. The input vector fed into the neural network is $[sign(\beta_{H}), sign(\beta_{V}), intruder status,$ $sign(\phi_{H}),$ $sign(\phi_{V}),$ $encounter type, previous action, action]^{T}$, where $sign(.)$ takes the values of $+1$, $-1$ or $0$ if its argument is a positive value, negative value or zero, respectively. The $intruder status$ is taken as $1$ whenever an intruder is detected by the pilot, and $0$, otherwise. The $encounter type$ (see Fig.~\ref{f:EC}) is fed to the neural network in the form of a vector with 2 elements indicating the encounter type in horizontal plane and vertical plane. The 1st element take the values of $-1$, $-0.5$, $0.5$ and $+1$ for encounter types of $C\#1$, $C\#2$, $C\#3$ and $C\#4$ in horizontal plane, and $0$, otherwise. Similarly, the 2nd element take the values of $-1$, $-0.5$, $0.5$ and $+1$ for encounter types of $C\#5$, $C\#6$, $C\#7$ and $C\#8$ in vertical plane, and $0$, otherwise. The $previous action$ and $action$ are fed to the neural network in the form of a vector with 4 elements: actions ``turn $45^{\circ}$ left", ``go straight", ``turn $45^{\circ}$ right", ``pitch $10^{\circ}$ up", or ``pitch $10^{\circ}$ down" are coded as $[0,0,0,1]$, $[0,0,1,0]$, $[0,1,0,0]$, $[1,0,0,0]$ and $[-1,-1,-1,-1]$, respectively.

\section{Simulation Results and Discussion}\label{Simulation Results and Discussion}
In this section, a quantitative analysis of multiple UAS integration in a crowded airspace is presented. Before presenting these results, single encounter scenarios, where two manned aircraft with different reasoning levels which are in a collision path, are investigated.

\subsection{Single Encounter Scenarios Between Manned Aircraft}\label{Pilot with Various Level-k Type Behavior in Single Encounter Scenarios}

\begin{figure}[t!]
	\centering	
	\includegraphics[width=12cm]{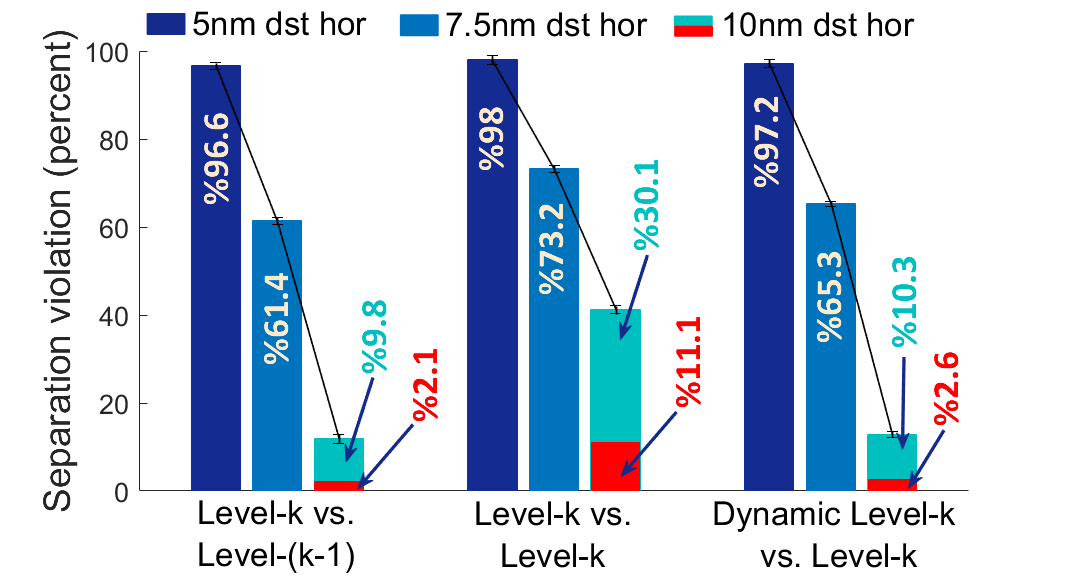}
	\caption{Separation violation rates. ``dst hor" refers to the distance horizon of the pilots. On the columns that shows the separation violation rates for a $10nm$ distance horizon, the cyan color shows the percentage of violations that occur when the encounters are ``difficult to resolve", meaning that the initial conditions of the encounters do not permit any type of pilot action to avoid a separation violation.}
	\label{f:pblevel}
\end{figure}

Figure ~\ref{f:pblevel} presents the separation violation rates of manned aircraft in 5000 random single encounters where pilots are modeled as level-1 and level-2 decision makers. Separation violation occurs when both the horizontal and the vertical distances between the two aircraft are less than the horizontal separation requirement, $5nm$ \cite{Perez:12} and the reduced vertical separation requirement, $1000ft$ \cite{Perez:12}, respectively. In the figure, separation violations are shown for 3 different ``distance horizon" values, which is the radius of the observation space depicted in Fig.~\ref{f:ObservationSpace}. Any distance less than the distance horizon is defined as miss separation. Pilots oversee a $20s$ time window prior to a probable miss separation distance with a $5second$ decision frequency for choosing their actions. Distance horizon takes three values: $5nm$ (equal to horizontal separation requirement), $7.5nm$ and $10nm$. Pilots are either level-0, level-1 or level-2 agents. There are 4 possible types of scenarios: 1) level-1 pilot vs. level-0 pilot, 2) level-2 pilot vs. level-1 pilot, 3) level-1 pilot vs. level-1 pilot, and 4) level-2 pilot vs. level-2 pilot. According to the Fig.~\ref{f:pblevel}, in the 1st type and 2nd type scenarios (level-k vs level-(k-1)), by increasing the distance horizon from $5nm$ to $10nm$ the separation violation rate decreases from $\%96.6$ to $\%11.9$. It is noted that $\%9.8$ of the $10nm$ distance horizon separation violations is occur during the scenarios where the encounters are difficult to be resolved. This is because for these type of encounters, no matter how pilots in the collision path maneuver to resolve the conflict the separation violation occur. In the 3rd type and fourth type of scenarios (level-k vs level-k) the separation violation rate deceases by increasing the pilots' distance horizon, however, separation violation rate is still high ($\%41.2$) even for the $10nm$ case. The reason for this high separation violation rate is that when a level-k agent encounters another agent with the same level, his/her assumption about the other becomes invalid. This problem and its implications were discussed in Section \ref{Dynamic Level-k Reasoning}, where a remedy were proposed (see Fig~\ref{f:PCN}), which resulted in a ``dynamic" level-k reasoning method. According to Fig.~\ref{f:pblevel}, the separation violation rate, for regular encounters (red), decreases from $\%11.1$ to $\%2.6$, when dynamic level-k reasoning is chosen as the interactive decision making model, for the case of $10nm$ distance horizon.

Two 3D single encounter scenarios are designed to schematically show the performance of pilots: 1) 1st scenario consists of two manned aircraft flying with constant speeds in different altitude levels where, their initial horizontal distance and vertical distances are $21.3nm$ and $1000ft$, respectively, 2) 2nd scenario consists of two manned aircraft where one of the manned aircraft is flying in a horizontal plane with constant speed and the other manned aircraft is leveling up with a constant vertical rate of $1968ft/min$ with initial horizontal distance and vertical distances of $21nm$ and $1000ft$, respectively. In these two scenarios, two manned aircraft are flying in an encounter path with each other and pilots can oversee conflicts in a $20s$ time window prior to miss separation. All pilot's distance horizon is considered to be $10nm$. Both of the aircraft can change the heading angle for $\pm45^{\circ}$, or change the pitch angle for $\pm10^{\circ}$, or may keep both the heading and pitch angles unchanged. In both of the scenarios 4 cases for pilots' level-k type behavior are considered: 1) a level-1 pilot vs. a level-0 pilot, 2) a level-2 pilot vs. a level-1 pilot, 3) a level-1 pilot vs. a level-1 pilot, and 4) a dynamic level-1 pilot vs. a level-1 pilot. 

\begin{figure}[t!]
	\centering	
	\includegraphics[width=8cm]{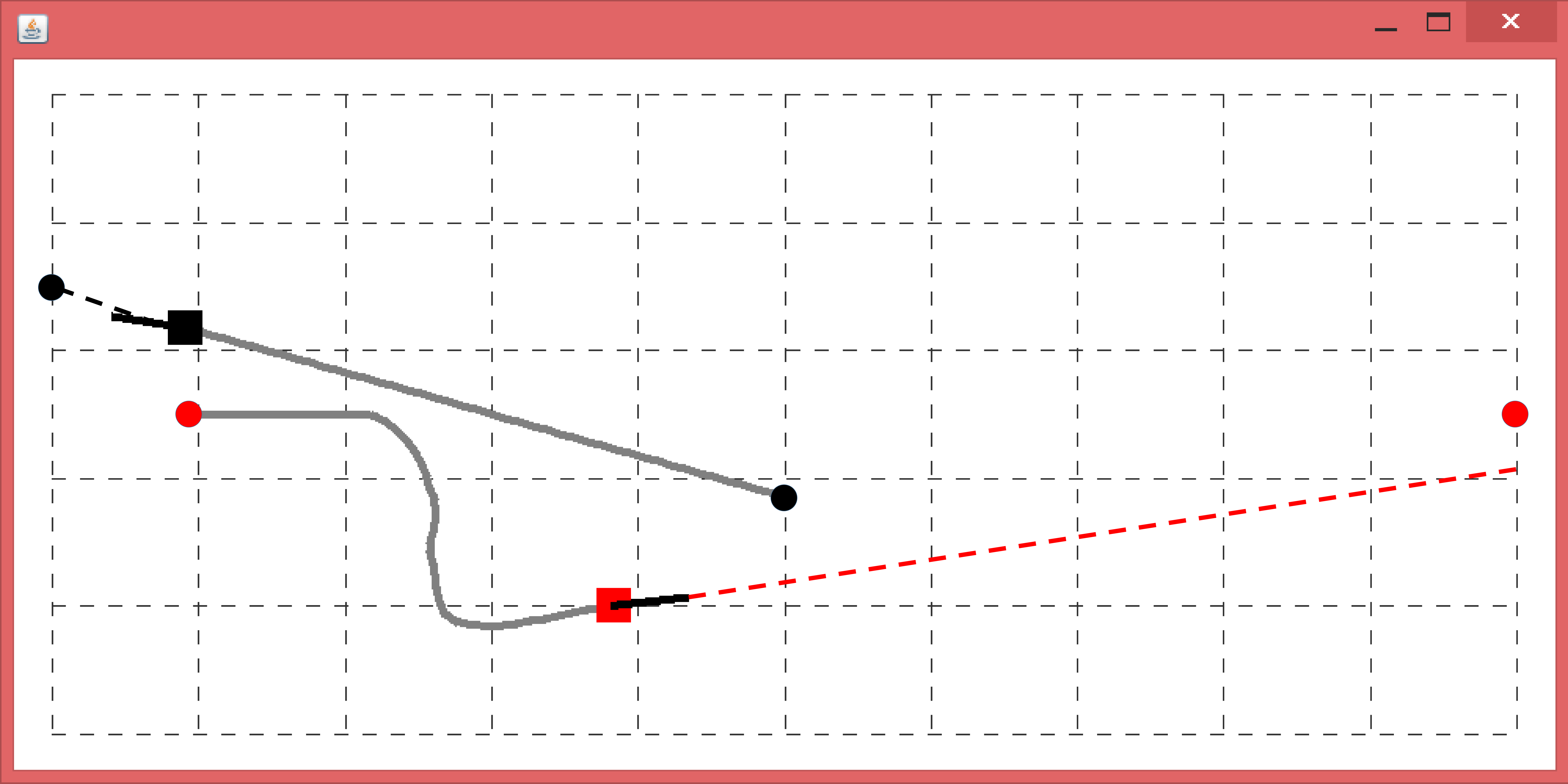}
	\caption{Sample encounter 1: level-1 pilot vs. level-0 pilot.}
	\label{f:S1}
\end{figure}

\begin{figure}[t!]
	\centering	
	\includegraphics[width=8cm]{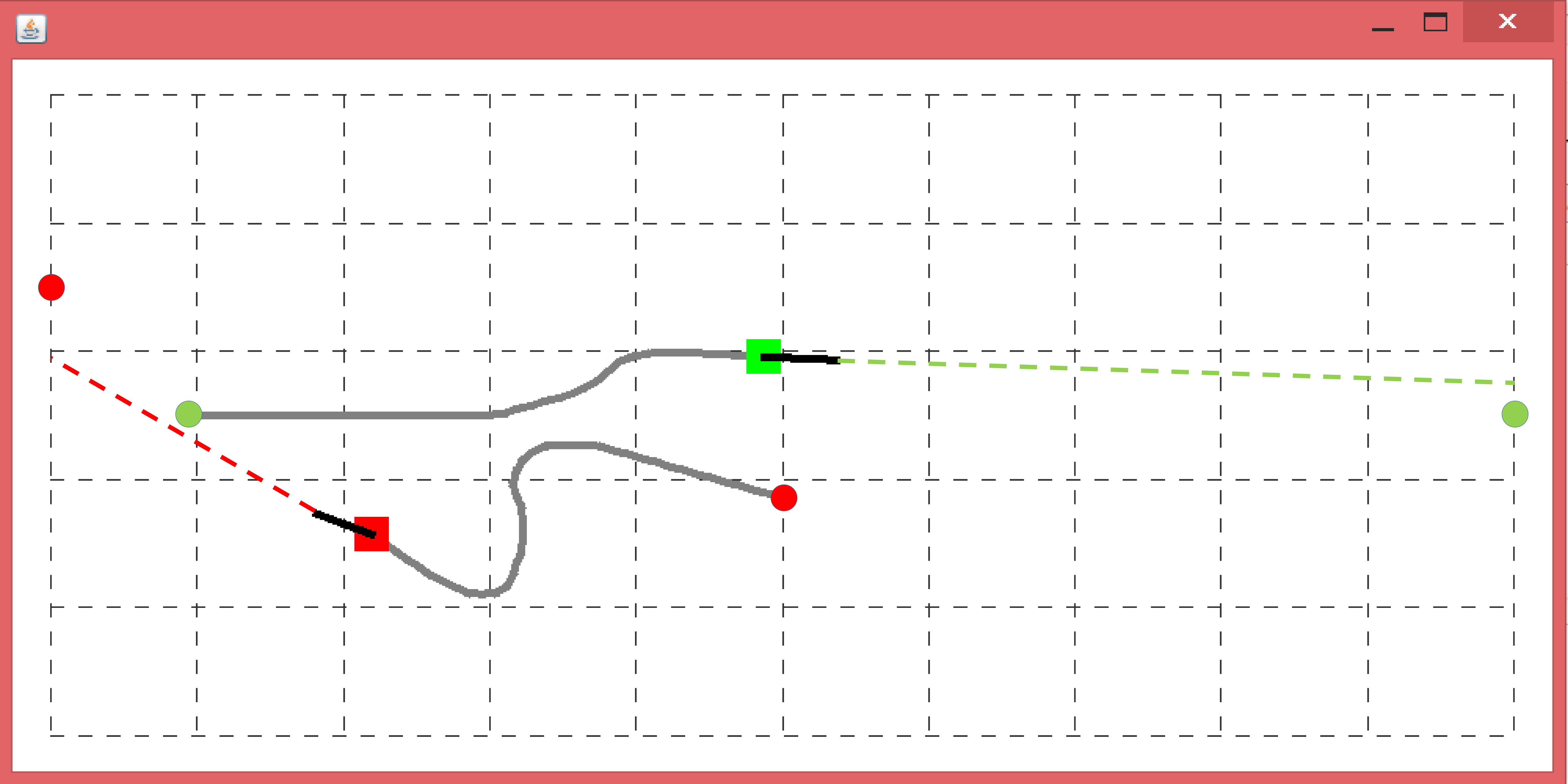}
	\caption{Sample encounter 1: level-2 pilot vs. level-1 pilot.}
	\label{f:S2}
\end{figure}

\begin{figure}[htb]
	\centering	
	\includegraphics[width=8cm]{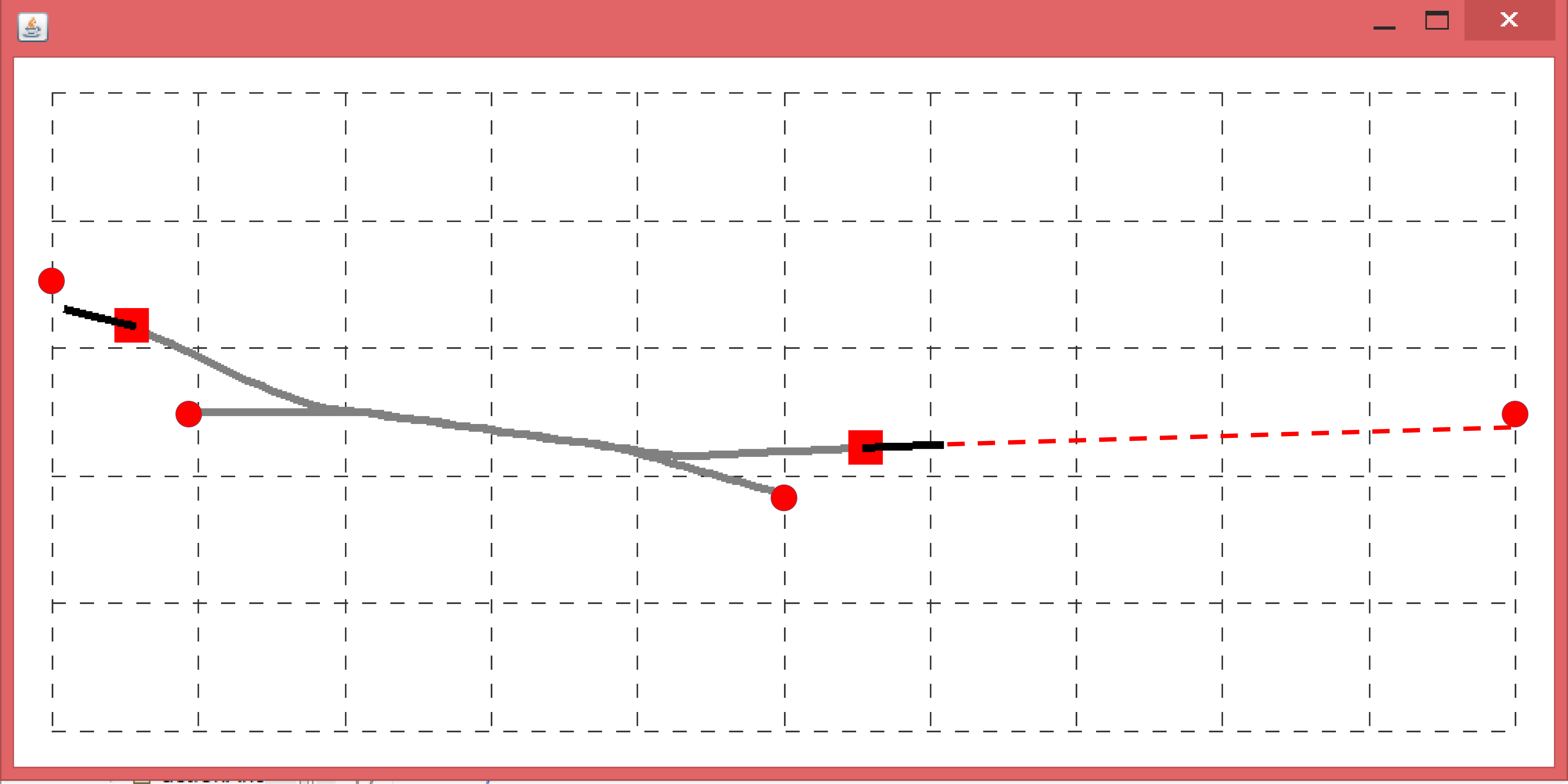}
	\caption{Sample encounter 1: level-1 pilot vs. level-1 pilot.}
	\label{f:S3}
\end{figure}

\begin{figure}[htb]
	\centering	
	\includegraphics[width=8cm]{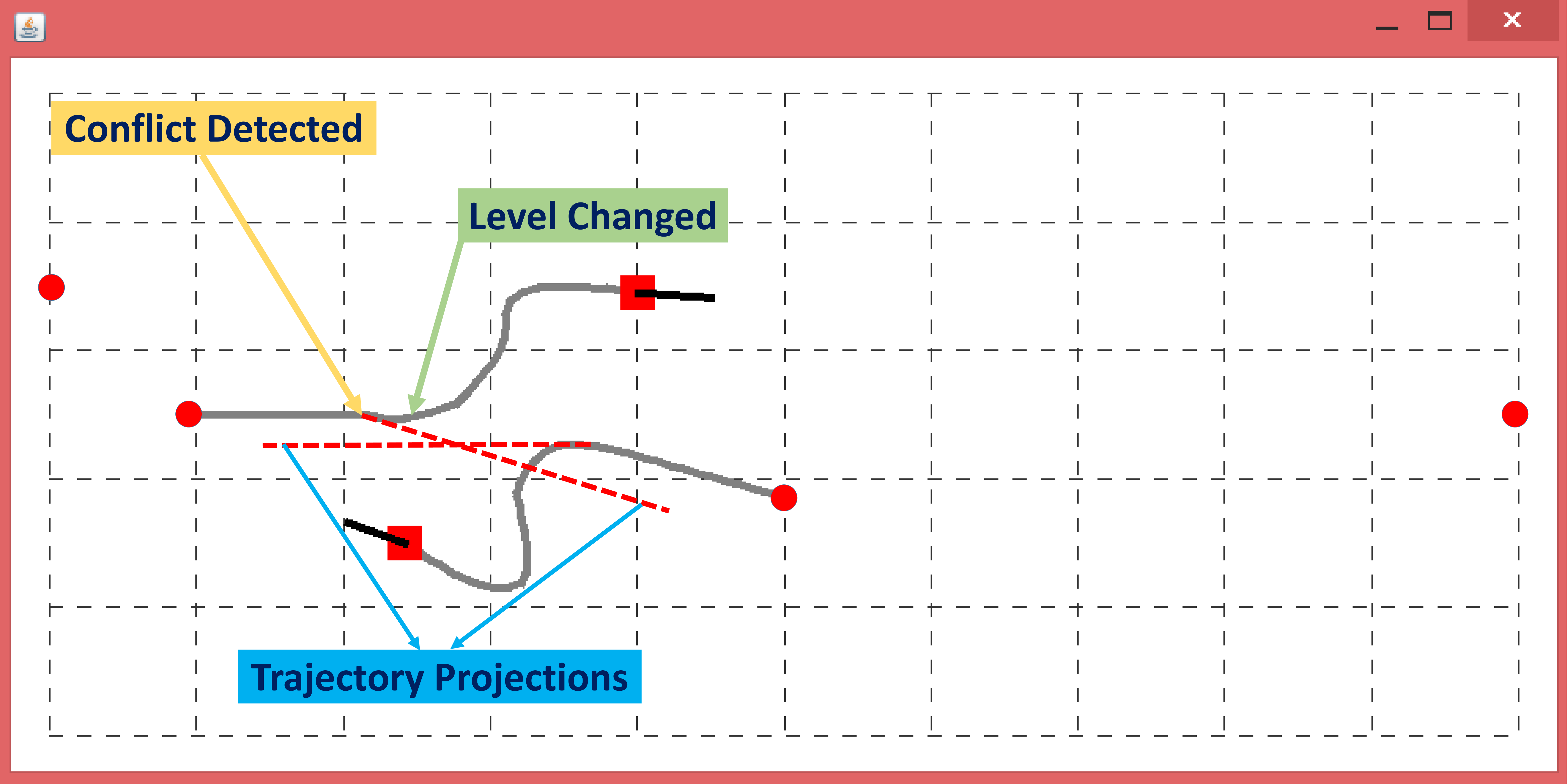}
	\caption{Sample encounter 1: dynamic level-1 pilot vs. level-1 pilot.}
	\label{f:S4}
\end{figure}

\begin{figure}[htb]
	\centering	
	\includegraphics[width=8cm]{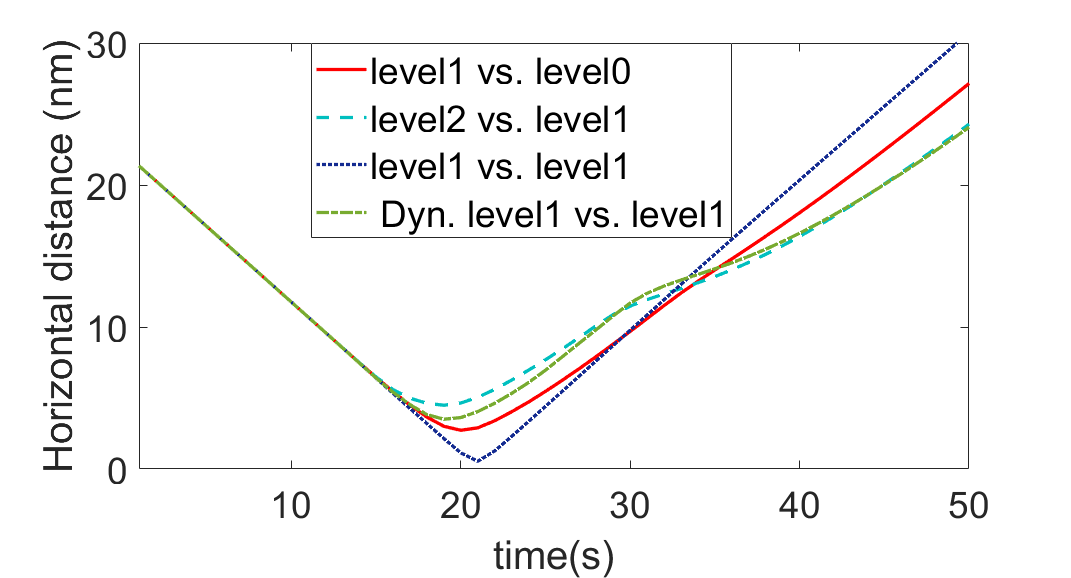}
	\caption{Sample encounter 1: horizontal distance.}
	\label{f:S5}
\end{figure}

\begin{figure}[htb]
	\centering	
	\includegraphics[width=8cm]{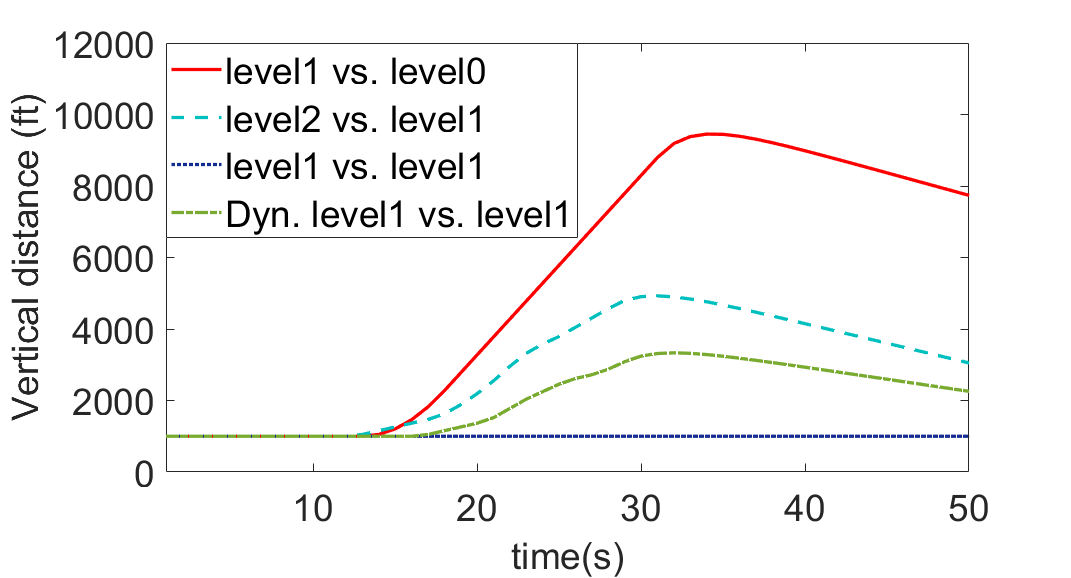}
	\caption{Sample encounter 1: vertical distance.}
	\label{f:S6}
\end{figure}

Figure~\ref{f:S1}, Fig.~\ref{f:S2}, Fig.~\ref{f:S3} and Fig.~\ref{f:S4} depict the 2D snapshots the four cases for the 1st scenario. The black squares, red squares and green squares correspond to manned aircraft with level-0 type behavior, level-1 type behavior and level-2 type behavior. The circles stand for the intila location and destination of the aircraft. The gray track line right behind the manned aircraft represent their traveled path from their initial positions to where they stand in the snapshot. Figure~\ref{f:S5} and Fig.~\ref{f:S6} depict the horizontal and vertical distances between the two aircraft for all the 4 cases in the 1st scenario. During the conflict resolution the minimum horizontal distance and vertical distance of the two aircraft are $2.7nm$ and $1825ft$ in the 1st case scenario, respectively and $4.5nm$ and $1471ft$ in 2nd case scenario. Therefore, pilots were able to resolve the conflict.

Taking a look at the 3rd case scenario where two pilots are level-1 type agents and assume that their intruders are level-1 type agents, pilots turn toward each other instead of resolving the encounter. During the encounter the minimum horizontal distance and minimum vertical distances are $0.5nm$ and $991ft$ and separation violation occurs. In the 4th case scenario, one of the pilots (the pilot starting its notion from left to right side of the snapshot) is able to observe its intruder's maneuver. This pilot first execute a maneuver such that its intruder is a level-1 type agent and turns toward its intruder. Then the pilot observes its intruder's maneuver within $10s$, project their trajectories and detects a critical conflict and changes its level-1 type rule to the level-2 type rule. In this case during the encounter the minimum horizontal distance and the minimum vertical distances are $3.5nm$ and $1051ft$, therefore, separation violation does not occur. It is noted that if pilots decided to go straight and do not participate in the conflict resolution the minimum horizontal distance and the minimum vertical distances are $0.54nm$ and $1000ft$.

\begin{figure}[htb]
	\centering	
	\includegraphics[width=8cm]{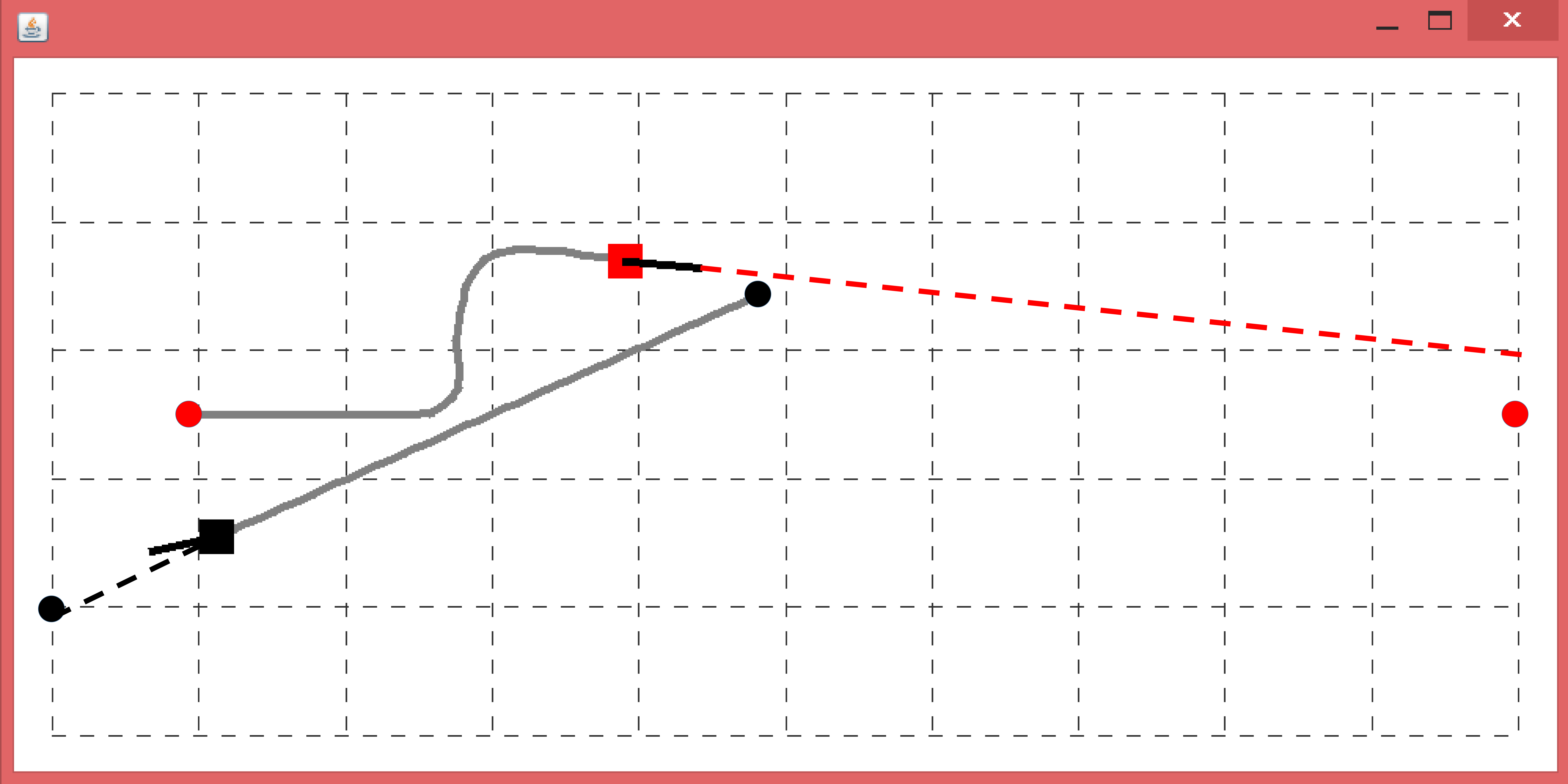}
	\caption{Sample encounter 2: level-1 pilot vs. level-0 pilot.}
	\label{f:S7}
\end{figure}

\begin{figure}[htb]
	\centering	
	\includegraphics[width=8cm]{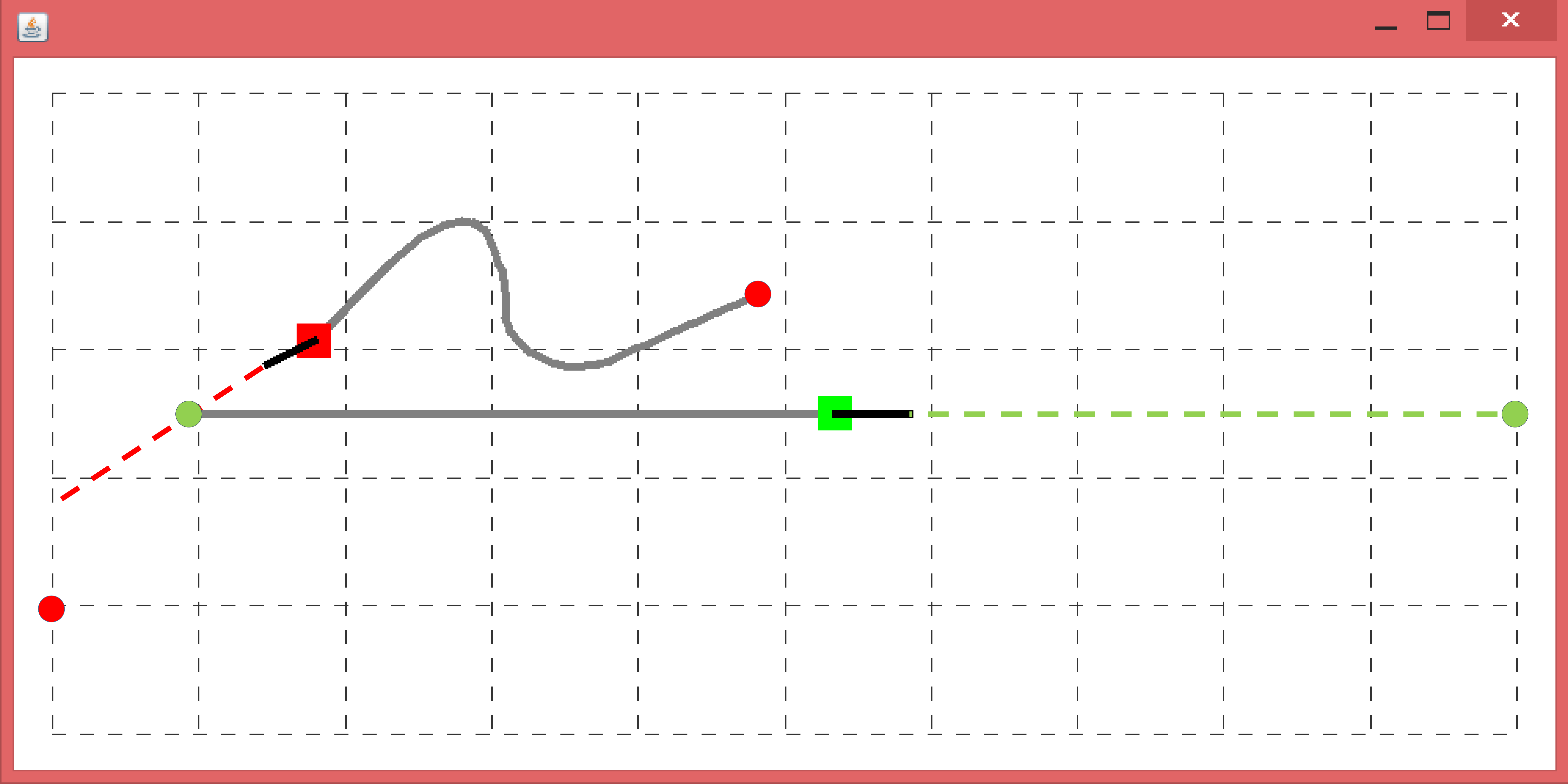}
	\caption{Sample encounter 2: level-2 pilot vs. level-1 pilot.}
	\label{f:S8}
\end{figure}

\begin{figure}[htb]
	\centering	
	\includegraphics[width=8cm]{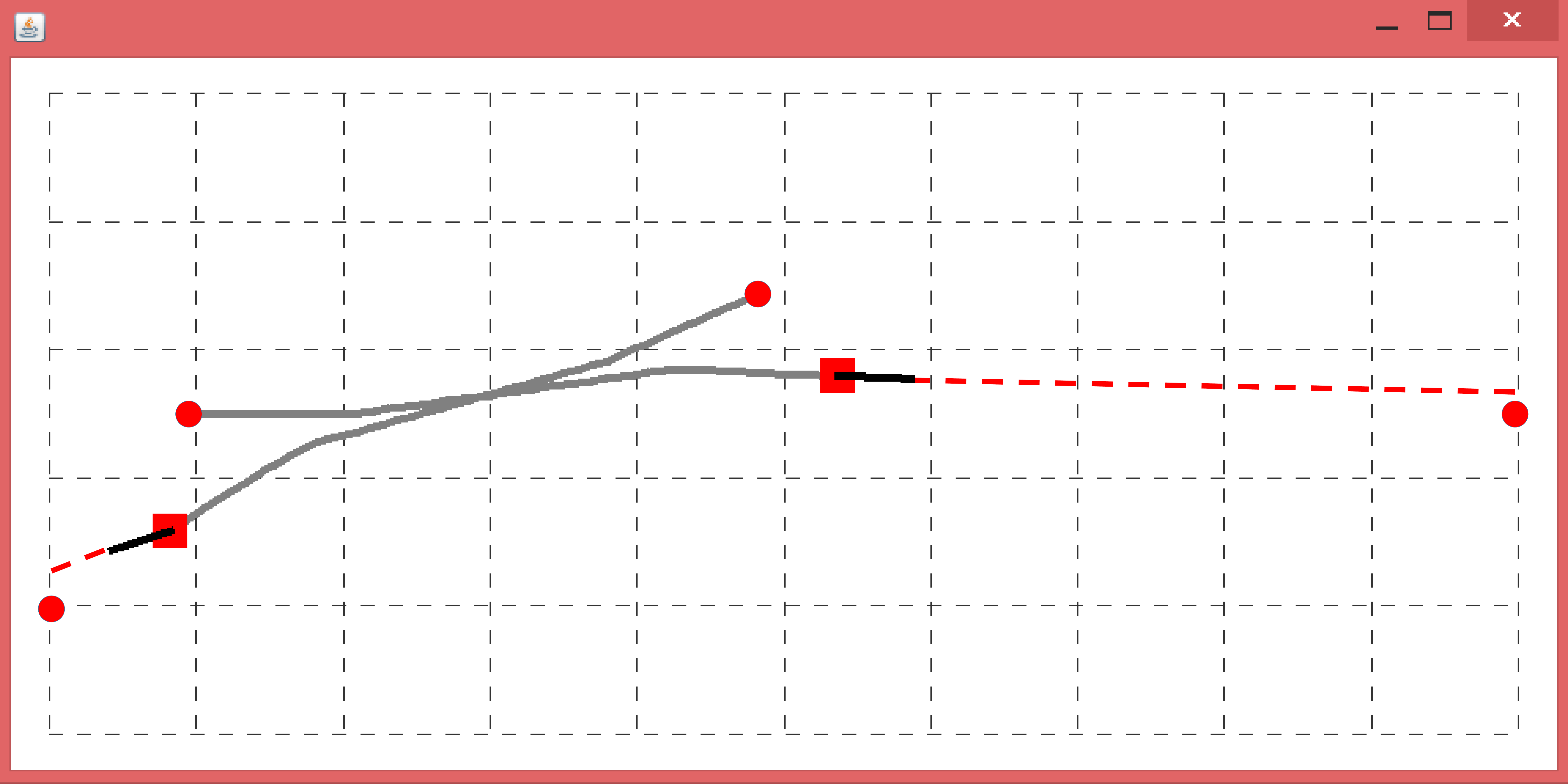}
	\caption{Sample encounter 2: level-1 pilot vs. level-1 pilot.}
	\label{f:S9}
\end{figure}

\begin{figure}[htb]
	\centering	
	\includegraphics[width=8cm]{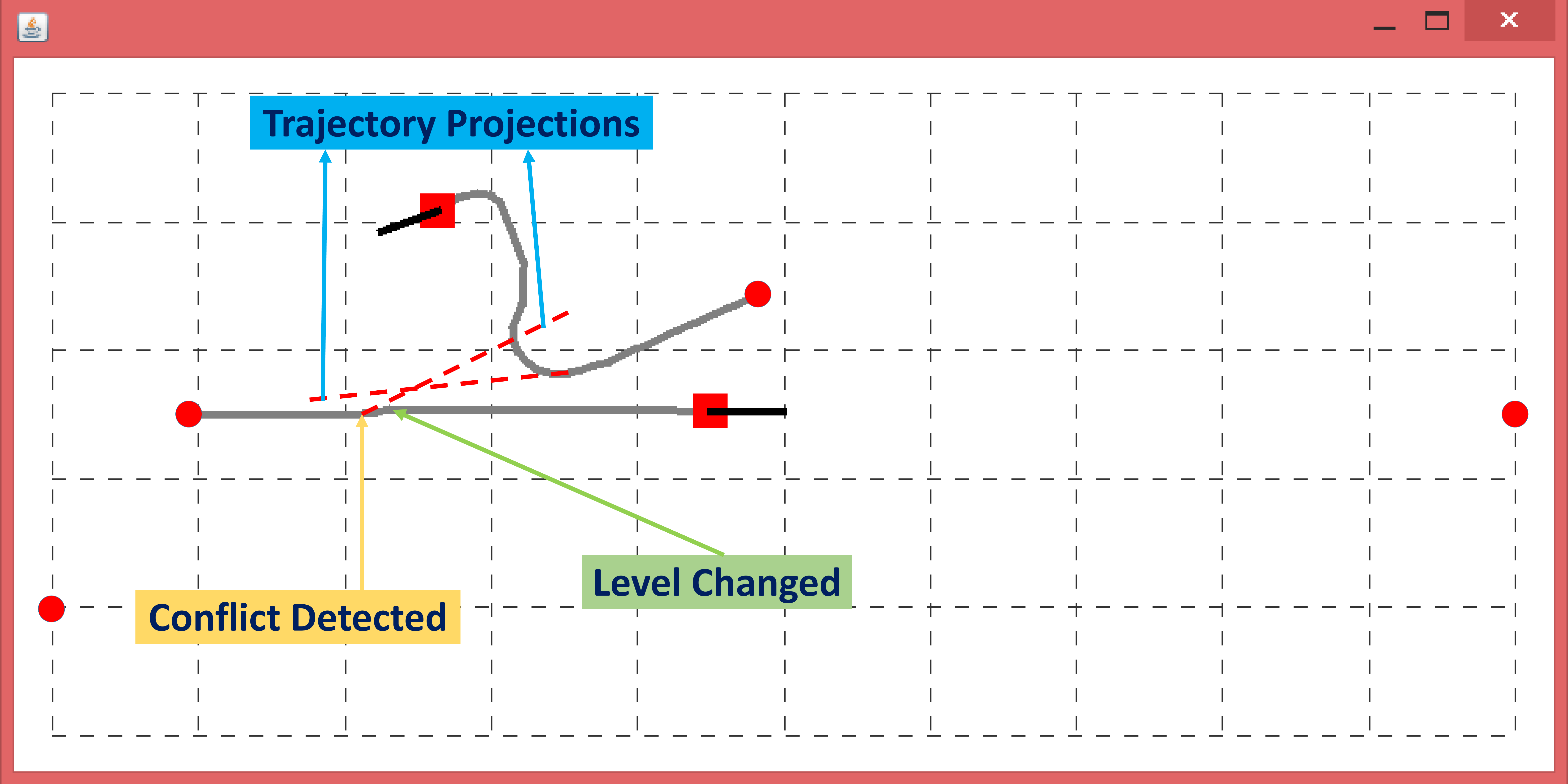}
	\caption{Sample encounter 2: dynamic level-1 pilot vs. level-1 pilot.}
	\label{f:S10}
\end{figure}

\begin{figure}[htb]
	\centering	
	\includegraphics[width=8cm]{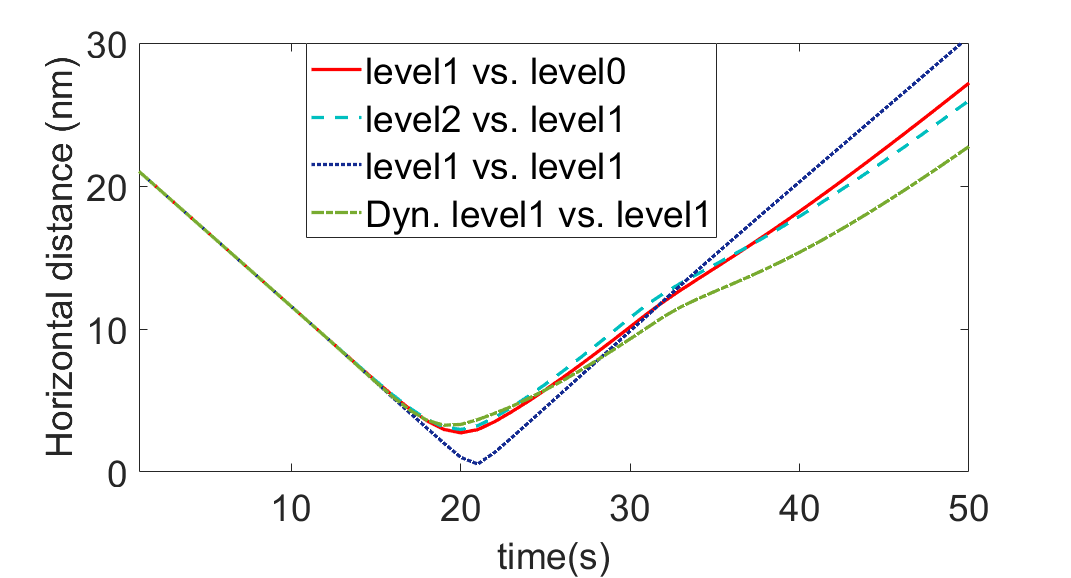}
	\caption{Sample encounter 2: horizontal distance.}
	\label{f:S11}
\end{figure}

\begin{figure}[htb]
	\centering	
	\includegraphics[width=8cm]{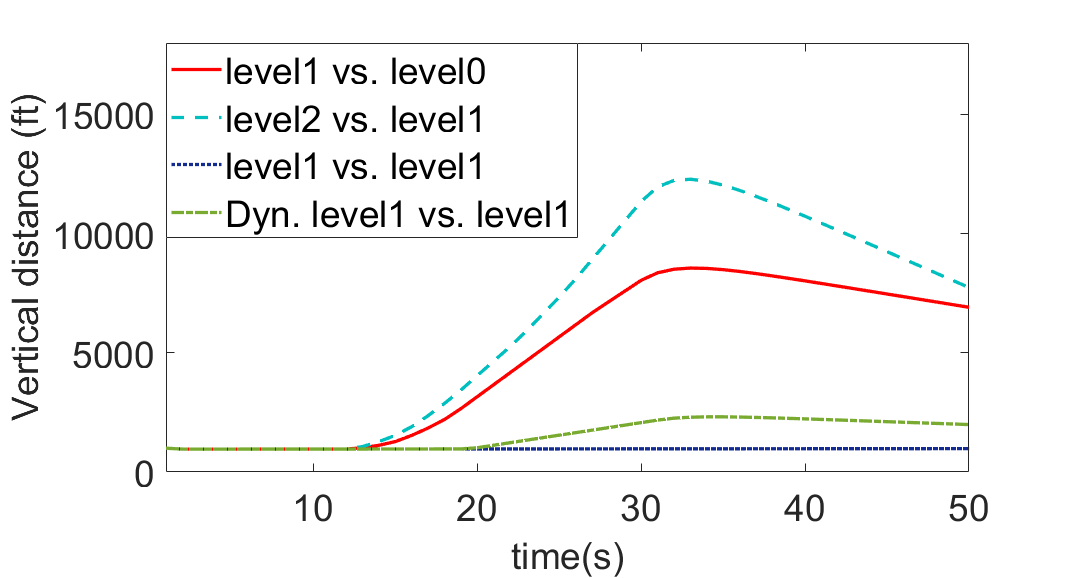}
	\caption{Sample encounter 2: vertical distance.}
	\label{f:S12}
\end{figure}

Figure~\ref{f:S7}, Fig.~\ref{f:S8}, Fig.~\ref{f:S9} and Fig.~\ref{f:S10} depict the 2D snapshots for the four cases of the 2nd scenario. Figure~\ref{f:S11} and Fig.~\ref{f:S12} represent the horizontal and vertical distances of the two manned aircraft. During conflict resolution in the 1st case scenario the minimum horizontal distance and the minimum vertical distance of the two aircraft are $2.8nm$ and $1847ft$ and in the 2nd case scenario, the minimum horizontal distance and the minimum vertical distance of the two aircraft are $3nm$ and $2367ft$. Similar to the 1st scenario, pilots resolve the conflict. During conflict resolution in the 3rd case scenario the minimum horizontal distance and the minimum vertical distance of the two aircraft are $0.56nm$ and $966ft$ and in the 4th case scenario, the minimum horizontal distance and the minimum vertical distance of the two aircraft are $3.3nm$ and $1001ft$. Similar to the 1st scenario, utilizing the dynamic level-k rule, pilots are able to resolve the conflict.

\subsection{Statistical Results For UAS Integration}\label{Statistical Results For UAS Integration}
In this section, the scenario explained in Section \ref{UAS Integration Scenario} is simulated to investigate the effect of \textit{responsibility assignment} for conflict resolution, on safety and performance for various number of UAS integrated in to the crowded airspace. It is noted that the separation responsibility is an important issue in addressing the integration of UAS into NAS \cite{NASA:12}: it is crucial to determine which of the agents (manned aircraft or UAS) will take the responsibility of the conflict resolution. Since the loss of separation is the most serious issue, the safety metric is taken as the total number of separation violations between all aircraft whether manned or unmanned. Performance metric, on the other hand, is taken as the averaged manned and unmanned aircraft trajectory deviations. In all of the simulations, level-0, level-1 and level-2 pilot policies are randomly distributed over the manned aircraft in such a way that 10\% of the pilots fly based on level-0 policies, 60\% of the pilots act based on level-1 policies and 30\% use level-2 policies. This distribution is obtained from human experimental studies discussed in \cite{Costa:09} and may not necessarily reflect the true distribution for pilots. It is noted, however, that level distribution can easily be adapted to other distributional data in this framework. Level-1 type and level-2 type pilots utilize the dynamic level-k reasoning method.

Figure~\ref{f:R1}, Fig.~\ref{f:R2}, Fig.~\ref{f:R3}, and Fig.~\ref{f:R4} depict a comparison of different resolution responsibility cases for the two SAA1 ans SAA2 logic: manned aircraft are responsible (dark blue), both manned aircraft and UAS are responsible (blue) and only the UAS is responsible (cyan). In the case when only manned aircraft are responsible for conflict resolution, UAS is forced to continue its path without executing the SAA system and manned aircraft act as dynamic level-1 and level-2 DMs. In the case when the UAS is responsible for the conflict resolution, the manned aircraft are forced to continue their path without changing their heading and the UAS executes its SAA system. In the case when both the manned aircraft and the UAS are responsible for the conflict resolution, they both execute their evasive maneuvers. Figure~\ref{f:R1} shows that manned aircraft deviate more from their trajectory when only the manned aircraft share resolution responsibility, compared to the case when both of the UAS and the manned aircraft are responsible. This is true for both the SAA1 (the results on the left) and the SAA2 (the results on the right) algorithms. 
On the other hand, Fig.~\ref{f:R2} shows that the UAS deviates from its trajectory more when it is responsible for the resolution, compared to the case when the responsibility is shared, when SAA1 is utilized. For the case of SAA2 utilization, the case is the same. Figure~\ref{f:R3} shows, as expected, that for both SAA1 and SAA2, the UAS flight times are the shortest when only the manned aircraft become responsible for the resolution. According to the Fig.~\ref{f:R4} for both SAA1 and SAA2 utilizations, the safest case is when the resolution responsibility is shared between the UAS and the manned aircraft..

\begin{figure}[htb]
	\centering	
	\includegraphics[width=8cm]{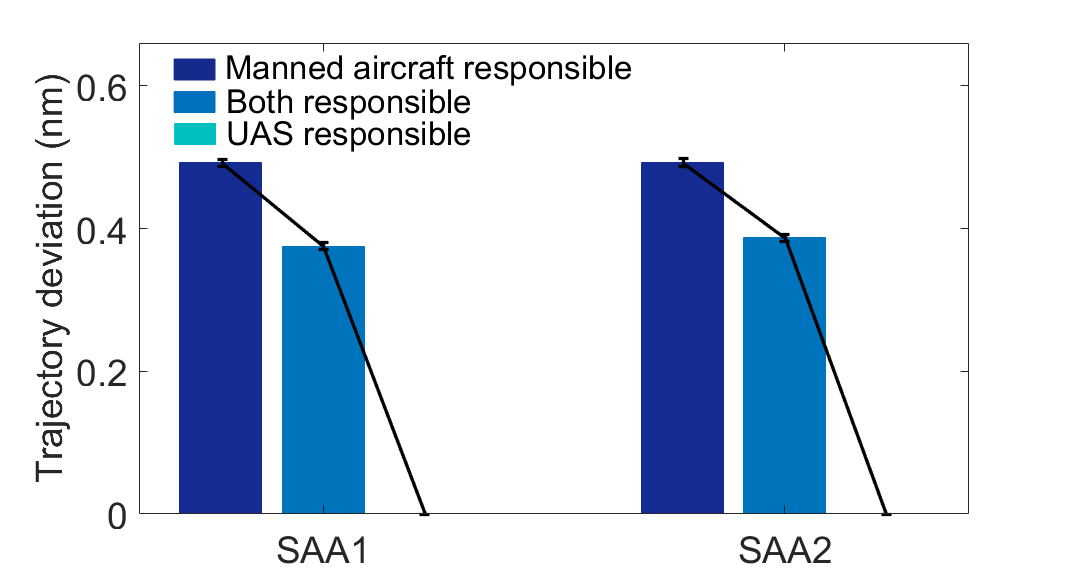}
	\caption{Average trajectory deviation of manned aircraft.}
	\label{f:R1}
\end{figure}

\begin{figure}[htb]
	\centering	
	\includegraphics[width=8cm]{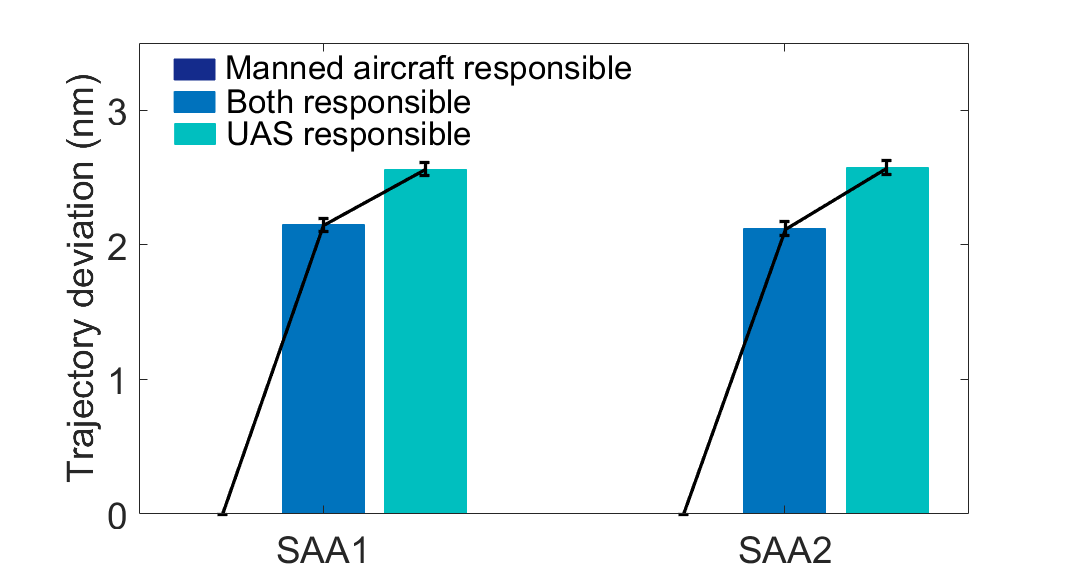}
	\caption{Average trajectory deviation of UAS.}
	\label{f:R3}
\end{figure}

\begin{figure}[htb]
	\centering	
	\includegraphics[width=8cm]{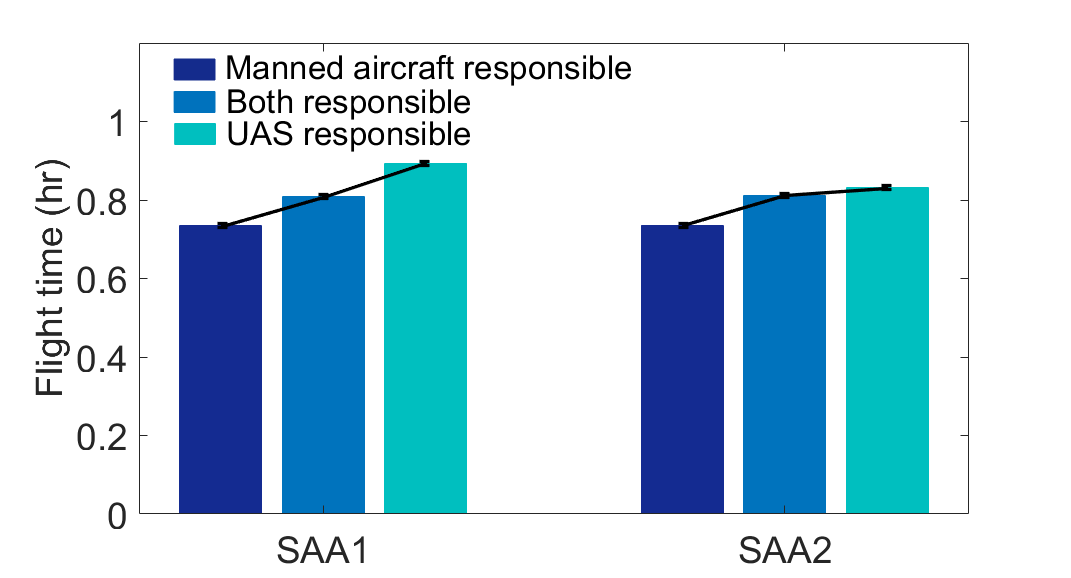}
	\caption{UAS flight time.}
	\label{f:R2}
\end{figure}

\begin{figure}[htb]
	\centering	
	\includegraphics[width=8cm]{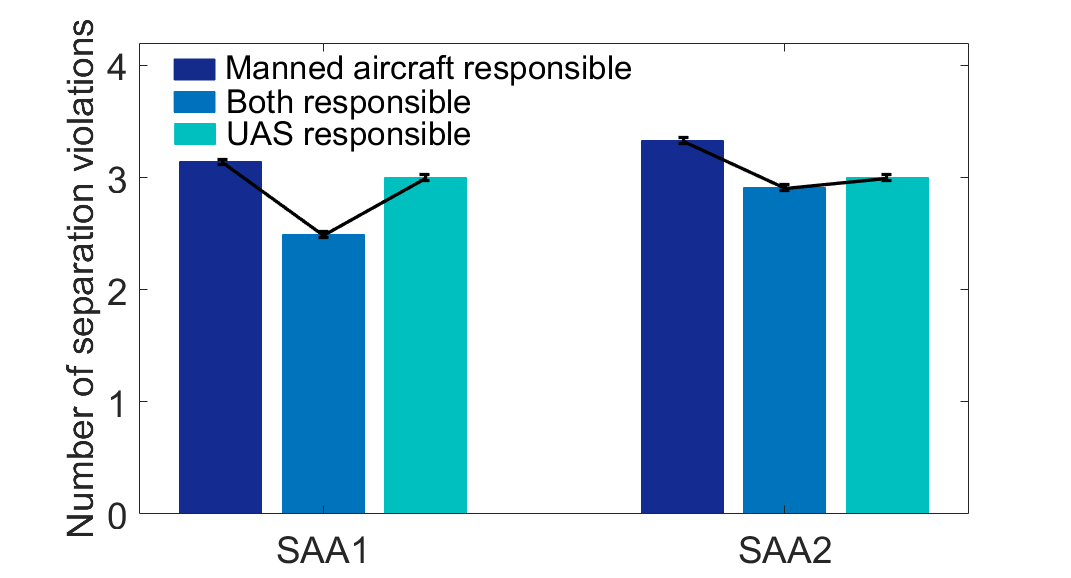}
	\caption{Separation violation between manned aircraft and UAS.}
	\label{f:R4}
\end{figure}

\chapter{Conclusion}

In this thesis, a 2D and a 3D game theoretical modeling framework are proposed for use in the integration of Unmanned Aircraft Systems (UAS) into the National Airspace System (NAS), as a means for concept evaluations. In both of these frameworks, the inclusion of the human decision making process in modeling the human pilot behavior is the main contribution, which fills an important gap in the literature of hybrid airspace systems (HAS) modeling, where in general, the pilot reactions are assumed to be known a priori. The method used in both of the approaches provides probabilistic outcomes of complex scenarios where both manned and unmanned aircraft co-exist. Thus, by being able to provide quantitative analyses, the proposed frameworks prove themselves to be useful in investigating the effect of various system variables, such as separation distances and the utilization of different Sense And Avoid (SAA) algorithms, on the safety and performance metrics of the airspace system. These frameworks can also be used to quantitatively analyze the effect of different responsibility assignments for conflict resolution and SAA design parameters such as UAS distance horizon and UAS time horizon, on safety and performance of the national airspace. These analyses can be helpful to evaluate technologies and concepts related to UAS integration into the NAS. It is noted that, the proposed frameworks are flexible so that any rules and procedures that the pilots are required to follow, for example Traffic Control Alert Systems (TCAS) advisories, can be incorporated into the model. Different manned aircraft and UAS dynamic models can be incorporated to the framework as well.

%


%
\newpage
\appendix

\backmatter

%
\bibliographystyle{IEEE_ECE}
\bibliography{whole}


\end{document}